\documentclass[11pt]{article}

\usepackage{fullpage}
\usepackage{setspace}
\usepackage{parskip}
\usepackage{titlesec}
\usepackage[section]{placeins}
\usepackage{xcolor,colortbl} 
\usepackage{breakcites}
\usepackage{lineno}

\usepackage[utf8]{inputenc}
\usepackage[english]{babel}

\usepackage{subcaption}
\usepackage{float}

\usepackage{graphicx}
\usepackage{multirow}
\usepackage{amsmath,amssymb,amsfonts}
\usepackage{amsthm}
\usepackage{mathrsfs}
\usepackage[title]{appendix}
\usepackage{xcolor}
\usepackage{textcomp}
\usepackage{manyfoot}
\usepackage{booktabs}
\usepackage{algorithm}
\usepackage{algpseudocode}
\usepackage{listings}

\usepackage[space]{grffile}
\usepackage{latexsym}
\usepackage{longtable}
\usepackage{tabulary}
\usepackage{array}
\usepackage{tabularx}

\usepackage{float}

\usepackage{times}

\PassOptionsToPackage{hyphens}{url}
\usepackage[colorlinks = true,
            linkcolor = blue,
            urlcolor  = blue,
            citecolor = blue,
            anchorcolor = blue]{hyperref}
\usepackage{etoolbox}

\renewenvironment{abstract}
  {{\bfseries\noindent{\abstractname}\par\nobreak}\footnotesize}
  {\bigskip}

  {\begin{tabular}{|p{13cm}}}
  {\end{tabular}}
  
\titlespacing{\section}{0pt}{*3}{*1}
\titlespacing{\subsection}{0pt}{*2}{*0.5}
\titlespacing{\subsubsection}{0pt}{*1.5}{0pt}

\usepackage{authblk}

\providecommand\citet{\cite}
\providecommand\citep{\cite}

\newif\iflatexml\latexmlfalse

\begin{document}

\title{Synthetic Time Series Generation via Complex Networks\footnote{Accepted authors manuscript (AAM) published in International Journal of Data Science and Analytics, 10.1007/s41060-026-01271-x. Final source webpage: \url{https://doi.org/10.1007/s41060-026-01271-x}. CONTACT Vanessa Freitas Silva. Email: \href{mailto:vanessa.silva@fc.up.pt}{vanessa.silva@fc.up.pt} }}

\author[1]{Jaime Vale}%
\author[1,2]{Vanessa Freitas Silva}%
\author[1,2]{Maria Eduarda Silva}%
\author[3,4]{Fernando Silva}%

\affil[1]{Faculdade de Ciências, Universidade do Porto}%
\affil[2]{INESC TEC-CRACS}%
\affil[3]{Faculdade de Economia, Universidade do Porto}%
\affil[4]{INESC TEC-LIAAD}%

\vspace{-1em}

\date{}

\begingroup
\let\center\flushleft
\let\endcenter\endflushleft
\maketitle
\endgroup

\selectlanguage{english}
\begin{abstract}
Time series data are essential for a wide range of applications, yet access to high-quality datasets is often constrained by privacy concerns, acquisition costs, and labelling challenges. Synthetic time series generation has emerged as a promising approach to address these limitations. In this work, we investigate the use of complex network mappings for synthetic time series generation, focusing on the Quantile Graph (QG) representation and its inverse. While the inverse QG mapping has been previously proposed, its potential as a general-purpose data generator has not been systematically evaluated.
We address this gap through a comprehensive empirical study assessing both the fidelity and utility of synthetic time series generated by the Inverse Quantile Graph (InvQG) framework. The evaluation combines statistical feature analysis, network-based topological characteristics, and performance in downstream clustering and classification tasks, using simulated and real-world datasets. The results show that InvQG effectively preserves marginal distributions and short-term temporal dependencies across a wide range of models, while exhibiting predictable limitations in capturing long-range or higher-order dynamics.

\end{abstract}%

\section{Introduction}\label{sec1}

Time series data are central to a wide range of applications, yet access to high‑quality datasets is often constrained by privacy, cost, and collection limitations. Synthetic time series generation has therefore emerged as a key strategy to enable data‑driven modelling while mitigating these constraints~\cite{Lin2020}. A central challenge, however, is to generate synthetic data that simultaneously preserve essential statistical and temporal properties of the original series while remaining useful for downstream analytical tasks such as clustering, classification or forecasting. 

Deep generative models, particularly Generative Adversarial Networks (GANs)~\cite{Lin2020, Brophy2023, Yoon2019}, have become a dominant approach for synthetic time series generation due to their flexibility in modelling complex data distributions. However, these methods typically require extensive training, careful tuning, and substantial computational resources, and may suffer from instability and limited interpretability~\cite{iglesias2023data}. These challenges motivate the exploration of alternative approaches that prioritise transparency, robustness, and efficiency, while still achieving competitive fidelity and utility.

In this work, we investigate the use of complex network mappings for synthetic time series generation, focusing on the Quantile Graph (QG) representation and its inverse mapping. While inverse QG mappings have previously been proposed for time series reconstruction~\cite{Campanharo2011}, their potential as a general‑purpose synthetic data generator has not been systematically evaluated. We address this gap by introducing the \textit{Inverse Quantile Graph} (InvQG) framework as a practical and interpretable method for generating synthetic time series based on transition dynamics encoded in quantile space.

Through a comprehensive empirical study on both simulated and real‑world datasets, we assess the method along three complementary dimensions: statistical fidelity, structural similarity via network representations, and utility in downstream clustering and classification tasks. The results show that InvQG effectively preserves marginal distributions and short‑term temporal dependencies across a broad range of time series models, while exhibiting predictable limitations in capturing long‑range or higher‑order dynamics. Despite these limitations, the generated data remain informative for downstream analysis and closely match the empirical distribution of the original data.

Overall, this work positions InvQG as a simple yet effective framework for synthetic time series generation, offering a transparent alternative to deep generative models and highlighting the trade‑offs between modelling flexibility, fidelity, and interpretability.

\paragraph{Contributions} 
This paper systematically evaluates the Inverse Quantile Graph (InvQG) mapping as a framework for synthetic time series generation, with the following main contributions:

\begin{itemize}
    \item We investigate the InvQG mapping as a general-purpose framework for synthetic time series generation, a use case that has not been systematically studied.
    \item We propose a comprehensive evaluation framework combining statistical features, network-based representations, and downstream mining tasks performance.
    \item We provide a detailed empirical analysis of the strengths and limitations of the method, showing its effectiveness in preserving short-term dependencies and its limitations in capturing long-range temporal structure.
    \item We benchmark InvQG against state-of-the-art GAN-based models, highlighting trade-offs in fidelity, interpretability, and computational cost.
\end{itemize}

Our analysis emphasises not only performance gains but also the conditions under which the method is expected to succeed or fail. Collectively, these contributions position InvQG as a competitive alternative for synthetic time series generation.

The remainder of the paper is structured as follows. We begin by introducing the relevant terminology and background, including related work on synthetic time series generation and time series mappings. We then present the proposed methodology, and conclude with a comprehensive empirical evaluation and benchmarking against
state-of-the-art GAN-based methods.

\section{Background and Related Work}\label{sec2}

In this section, we introduce the fundamental concepts and terminology for the remainder of this document.  We start by  reviewing related work in time series synthesizing methods. Then, we provide a detailed description of the mapping methods employed in this work, namely the \textit{Quantile Graph} (QG) mapping and its inverse~\cite{Campanharo2011}.

\subsection{Time Series Data Synthesizing and Augmenting Methods}

Time series data generation is not a novel concept but has gained traction in theory and practice as a promising solution to data shortages and limitations that hinder processing and extracting meaningful insights from the data~\cite{Asghar2017, Shi2011, Benhamouda2016} and training deep learning models effectively~\cite{iglesias2023data, iwana2021empirical, wen2020time}.

Data Augmentation is the process of modifying or creating new data by adding noise, performing permutations and by generating new synthetic data, all while preserving original data characteristics. In augmentation-related transformations, methods include manipulation techniques based on deformation, shortening, enlargement, and data modification~\cite{iglesias2023data}. For example, \textit{time slicing} involves cutting portions of data (windows) to create new samples, \textit{jittering} adds noise to simulate inherent data variability, \textit{scaling} modifies the magnitude of observations, either uniformly (homogeneous scaling) or variably at specific points (magnitude warping), \textit{rotation} transforms time series data by applying a rotation matrix (converted from original time series) to represent it in a new orientation, \textit{permutation} rearranges fixed or variable-length window segments of a time series to create new combinations, and \textit{channel permutation}, applied to multivariate time series, swaps the positions of features (channels) to expand the dataset~\cite{iwana2021empirical,iglesias2023data}. Traditional methods rely on imputation and statistical approaches. These typically involve applying mathematical expressions or perform a series of transformations to modify the data or simulate new samples in order to complete or augment the dataset~\cite{klopries2024itf}. 
Early approaches make use of simple interpolation techniques (such as linear and spline interpolation), regression models, and machine learning-based methods like K-Nearest Neighbors (KNN). 

Simulations or modelling are also often employed to replicate real-world scenarios when data is scarce. Examples include  autoregressive models, such as mixture autoregressive (MAR)~\cite{kang2020gratis}, Markov chains~\cite{negra2008model}, and seasonal-trend decomposition (STL)-based techniques~\cite{kegel2018feature}. 
Although these methods are useful, they often lack diversity and quality and fail to model or simulate complex temporal dynamics and feature interactions~\cite{klopries2024itf}. An emerging solution that is gaining increasing attention is the generation of synthetic time series that replicate the statistical and dynamic properties of the original datasets. Recent techniques based on deep-learning methods have been proposed for this purpose, including  \textit{Generative Adversarial Networks} (GANs)~\cite{Lin2020, Brophy2023, Yoon2019}  and \textit{Variational Autoencoders} (VAEs)~\cite{iglesias2023data, hu2020time}.
VAE-based methods~\cite{kingma2013auto} employ probabilistic latent variable models to learn the underlying distribution of time series data, enabling the generation of new data by sampling from the learned latent space~\cite{hu2020time}. Although VAEs are effective for synthetic data generation, their use has become less prevalent with the advent of newer neural network models, such as GANs, which can produce larger and more diverse datasets. Nevertheless, VAEs remain valuable in scenarios that require precise control over the variability of generated data, maintaining their  relevance in specific domains~\cite{iglesias2023data}. GANs are powerful tools for synthesizing high-quality, diversified, and privacy-preserving time series data. GANs not only generate synthetic datasets but also support tasks such as data augmentation, denoising, and imputing missing values~\cite{Brophy2023, Gao2022}. 

A standard GAN consists of two neural networks: the generator that creates synthetic data instances, aiming to mimic the original data; and the discriminator that evaluates whether a given data instance is real or synthetic. 
These networks are trained jointly using a zero-sum adversarial framework, where the generator improves by learning to "fool" the discriminator, and the discriminator evolves to better distinguish real from synthetic data. Despite their success, traditional GANs often struggle to preserve the temporal dependencies inherent to time series data~\cite{Brophy2023, Gao2022, Yoon2019}.

To address these limitations, specialized GAN architectures have been developed. 
Some more recent examples include: 
\textit{Recurrent Conditional GAN} (RCGAN) that focuses on recurrent structures to better capture sequential relationships~\cite{esteban2017real}; 
\textit{TimeGAN} that introduces a supervised loss function and an embedding network to better capture temporal dynamics and conditional distributions in time series data~\cite{Yoon2019}; 
and \textit{DoppelGANger} that enhances scalability and utility for complex time series by introducing conditional inputs and auxiliary variables~\cite{lin2019}.
Despite their strengths, GANs face several challenges. Training instability is a common issue due to the adversarial nature of the model, often leading to mode collapse, where the generator produces limited variations and repetitive outputs of data. Furthermore, achieving convergence is difficult, as the generator and discriminator must reach a balance, which may not always occur in practice. Additionally, GANs are computationally expensive, requiring significant resources for effective training. Applying GANs to large-scale datasets or high-resolution time series often amplifies these challenges, requiring architectural innovations or alternative frameworks~\cite{Brophy2023, iglesias2023data}.
These limitations hinder the widespread application of GANs and imply careful tuning and domain-specific adjustments for a successful implementation.

\subsection{Mapping Time Series into Quantile Graphs}\label{sec_qg}

Time series analysis using network science involves mapping time series data ($\boldsymbol{Y}$) into complex networks (graphs, $G$), composed of nodes ($V$) and edges ($E$). Nodes represent specific time points or temporal patterns, while edges capture temporal dependencies or relationships between pairs of nodes. Common mapping approaches rely on principles such as visibility, probability transition, proximity, and statistical principles~\cite{silva2021time}. 
In this work, we focus on the probability transition approach.

Transition probability mappings capture temporal dynamics from the original data and encode them into an adjacency matrix that defines the resulting graph. 
In particular, the \textit{Quantile Graph} (QG)~\cite{Campanharo2011} 
applies this approach by representing a time series $\boldsymbol{Y} = \{Y_t\}_{t=1}^T$ 
as a directed and weighted graph $G$ obtained from the following steps
 (see Algorithm~\ref{alg:qg}):
\begin{description}
    \item[\textbf{Step 1:}] split the support of the time series into $\mathcal{Q}$ sample quantiles, $\{q_1, q_2, \ldots, q_{\mathcal{Q}}\}$.
    \item[\textbf{Step 2:}] map each quantile, $q_i$, into a node $v_i \in V$ in the graph $G$.
    \item[\textbf{Step 3:}] link two nodes, $v_i$ and $v_j$, with a directed and weighted edge, $(v_i,v_j, w_{i,j}) \in E$, 
    based on how often the time series transitions between their corresponding quantiles 
        $q_i$ and $q_j$, in the series. Weight $w_{i,j}$ measures the frequency a data point $Y_t$ in quantile $q_i$ is followed by $Y_{t+1}$ in quantile $q_j$.
    \item[\textbf{Step 4:}] normalize the adjacency matrix of $G$ into a Markov transition matrix, $\boldsymbol{W}$, where $\sum_{j=1}^{\mathcal{Q}} w_{i,j} = 1$, for each $i=1,\ldots,\mathcal{Q}$, where the weight $w_{i,j}$ represents the transition probability between the corresponding quantile samples.
\end{description}

\begin{figure}[H]
    \centering
        \includegraphics[width=\textwidth]{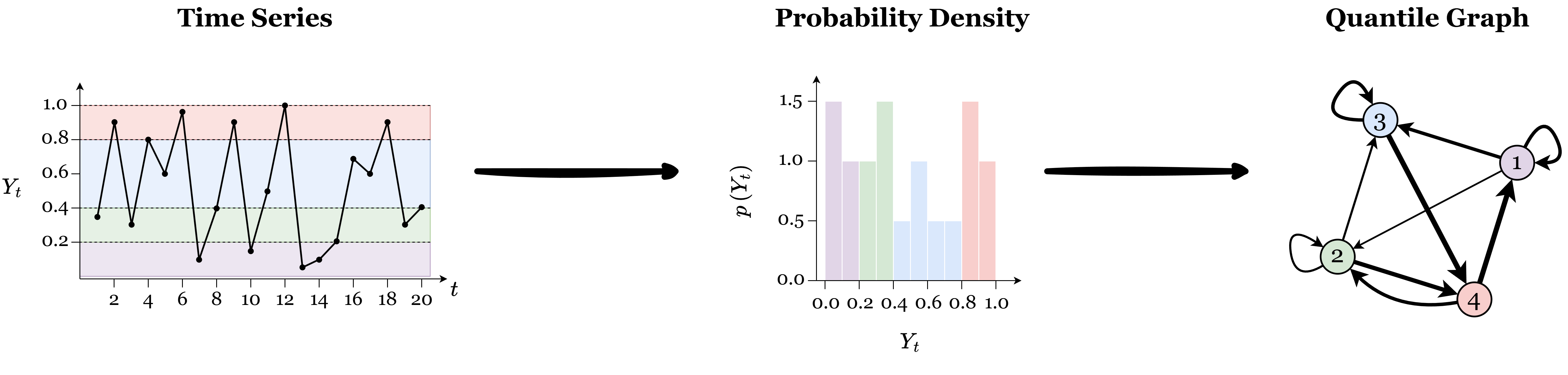}
    \caption[Illustrative example of the Quantile Graph (QG) mapping]{Illustration of the Quantile Graph (QG) mapping applied to a toy time series, showing the transition from time series data to a directed weighted graph representation. \textit{Source:} Adapted from \cite{Campanharo2011, silva2021time}.}
    \label{fig:qg_method_process}
\end{figure}

Figure~\ref{fig:qg_method_process} illustrates the QG mapping method.

\newpage 

\begin{algorithm}[H]
\caption{Quantile Graph (\textit{Source:} Adapted from~\cite{silva2024multilayer})}\label{alg:qg}

    \hspace*{\algorithmicindent} \textbf{Input:} {time series $\boldsymbol{Y}$ and number of quantiles $\mathcal{Q}$} \\
    \hspace*{\algorithmicindent} \textbf{Output:} {Markov transition matrix $\boldsymbol{W}$ and the sample quantiles $quantiles$} \\  
    \hspace*{\algorithmicindent} {\textbf{quantile} calculates the range of quantiles from time series support} \\
    \hspace*{\algorithmicindent} {\textbf{index\_of} finds the quantile of a given value}
    
    \begin{algorithmic}[1]
        \Procedure{QG}{$\boldsymbol{Y}, \mathcal{Q}$}

            \State $\boldsymbol{W} \gets \text{Array}(\mathcal{Q}, \mathcal{Q})$
            \State $quantiles \gets \text{Array}(\mathcal{Q})$
            \State $q \gets \text{Array}(\mathcal{Q})$

            \For{$i \gets 1 \text{ to } \mathcal{Q}$}
                \State $quantiles[i] \gets i / \mathcal{Q}$
            \EndFor
            \State $q \gets \text{quantile}(\boldsymbol{Y}, quantiles)$ 

            \For {$i \gets 1 \text{ to } \text{size}(\boldsymbol{Y}) - 1$}
                \State $vi \gets \text{index\_of}(\boldsymbol{Y}[i], q)$
                \State $vj \gets \text{index\_of}(\boldsymbol{Y}[i+1], q)$
                \State $\boldsymbol{W}[vi][vj] \gets \boldsymbol{W}[vi][vj] + 1$
            \EndFor

            \For{$i \gets 1 \text{ to } \mathcal{Q}$}
                \State $sum \gets 0$
                \For{$j \gets 1 \text{ to } \mathcal{Q}$}
                    \State $sum \gets sum + {W}[i][j]$
                \EndFor
                \For{$j \gets 1 \text{ to } \mathcal{Q}$}
                        \State $\boldsymbol{W}[i][j] \gets \boldsymbol{W}[i][j] / sum$
                \EndFor
            \EndFor

            \State \Return $\boldsymbol{W}, quantiles$
            
        \EndProcedure
    \end{algorithmic}
\end{algorithm}

\newpage

Campanharo et al. in~\cite{Campanharo2011} introduced an inverse mapping function for the QG described above enabling the generation of a new time series $\boldsymbol{Y}^*$ from an existing quantile graph $G$ that represents the original series $\boldsymbol{Y}$. 
This mapping leverages the Markov transition matrix $\boldsymbol{W}$, associated to $G$'s adjacency matrix, to reconstruct $\boldsymbol{Y}$. 

Although the inverse QG mapping enables the reconstruction of time series from their network representation, its potential as a mechanism for synthetic time series generation remains largely unexplored. In particular, there is a lack of systematic studies assessing its ability to preserve statistical properties, temporal dynamics, and performance in downstream tasks.

In this work, we address this gap by proposing the inverse QG mapping as a practical and interpretable framework for synthetic time series generation. We provide a systematic empirical evaluation of its effectiveness, focusing on both fidelity to the original data and utility in downstream applications.

\section{Synthetic Time Series via Quantile Graphs }\label{sec3}

Let $\boldsymbol{Y} = \{Y_1, Y_2, \ldots, Y_T\}$ represent a time series of $T$ sequential  observations. The goal is to generate  synthetic time series $\boldsymbol{Y}^* = \{Y_1^*, Y_2^*, \ldots, Y_T^*\}$, such that $\boldsymbol{Y}^*$ retains the key statistical, structural and dynamic properties of $\boldsymbol{Y},$ i.e. fidelity preservation and the analysis of  $\boldsymbol{Y}^*$ is  meaningful in addressing particular tasks related to  $\boldsymbol{Y},$ such as classification or forecasting, i.e., utility preservation.

We introduce a 2-step approach:
\begin{enumerate}
    \item \textbf{Network‐Domain Mapping:} Introduce a forward mapping 
    \( \mathcal{M} : \mathcal{T} \;\longrightarrow\; \mathcal{G},\)
    which uses transition probabilities to transform a time series 
    \(\boldsymbol{Y} \in \mathcal{T}\) into a graph 
    \(G = (V, E, \boldsymbol{W}) \in \mathcal{G}\). 
    
    \item \textbf{Time‐Series Reverse Mapping (Reconstruction):} Define an inverse mapping 
    \(\mathcal{M}^{-1} : \mathcal{G} \;\longrightarrow\; \mathcal{T}\),
    to reconstruct a synthetic time series 
    \(\boldsymbol{Y}^* \in \mathcal{T}\) 
    from the Markov transition matrix \(\boldsymbol{W}\) representing graph $G$.
\end{enumerate}

\begin{figure}[H]
    \centering
        \includegraphics[width=\linewidth]{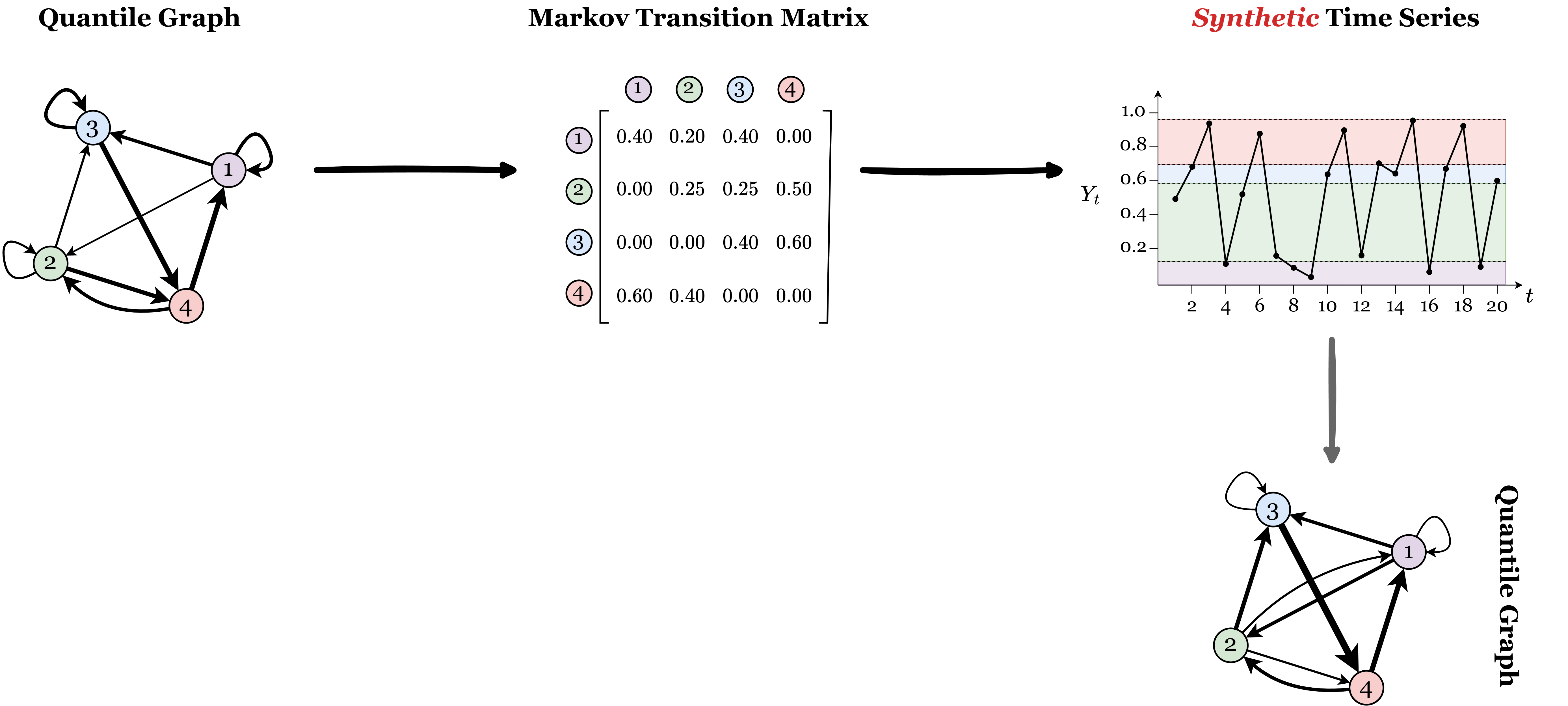}
    \caption[Illustrative example of the Inverse Quantile Graph mapping]{Illustration of the Inverse Quantile Graph (InvQG) mapping, showing the reconstruction of a synthetic time series from the quantile graph representation. \textit{Source:} Adapted from \cite{Campanharo2011}.}
    \label{fig:invqg_method_process}
\end{figure}

Focusing  on quantile-based mappings, our approach,  \textit{Inverse Quantile Graph} (InvQG) henceforward, is illustrated in Figure~\ref{fig:invqg_method_process} (see also~\cite{vale2024}).

InvQG mapping involves the following steps (see Algorithm~\ref{alg:re_construct_series}):

\begin{description}
    \item[\textbf{Step 1:}] Extract the adjacency (Markov transition) matrix \(\mathbf{W}\) of the quantile graph produced by QG mapping (line~\ref{alg2_line1}). 
    \item[\textbf{Step 2:}] Retrieve the sample quantiles of the original time series (line~\ref{alg2_line1}). 
    \item[\textbf{Step 3:}] Select an initial quantile \(q_k\) at random among those with non-zero outgoing probability (line~\ref{alg2_line5}). 
    \item[\textbf{Step 4:}] For \(t=1,\dots,T\):
        \begin{description}
            \item[\hspace{0.5cm}\textbf{Step 4.1:}] Determine the value range of the current quantile \(q_k\) (lines~\ref{alg2_line7}–\ref{alg2_line8}). 
            \item[\hspace{0.5cm}\textbf{Step 4.2:}] 
            Sample a value uniformly from that range (line~\ref{alg2_line9}). 
            \item[\hspace{0.5cm}\textbf{Step 4.3:}] 
            Transition to the next quantile \(q_{k'}\) by sampling according to the non-zero probabilities in row \(k\) of \(\mathbf{W}\) (line~\ref{alg2_line10}). 
        \end{description}
    \item[\textbf{Step 5:}] Return a synthetic time series data (line~\ref{alg2_line12}).
\end{description}

\newpage

\begin{algorithm}[H]
	\caption{Inverse Quantile Graph}   
    \label{alg:re_construct_series}

    \hspace*{\algorithmicindent} \textbf{Input:} {time series $\boldsymbol{Y}$ and number of quantiles $\mathcal{Q}$} \\
    \hspace*{\algorithmicindent} \textbf{Output:} {synthetic time series $\boldsymbol{Y}^*$} \\ 
    \hspace*{\algorithmicindent} {\textbf{init\_sample} select randomly sample initial quantile with non-zero probability} \\
    \hspace*{\algorithmicindent} {\textbf{runif} assign a random value following a uniform distribution in the quantile range} \\
    \hspace*{\algorithmicindent} {\textbf{select\_next\_quantil} select next node based on transition probabilities}

    \begin{algorithmic}[1]
    \Procedure{InvQG}{$\boldsymbol{Y}, \mathcal{Q}$}

        \State $\boldsymbol{W}, quantiles \gets \text{QG}(\boldsymbol{Y}, \mathcal{Q})$ \label{alg2_line1}
        \State $T \gets \text{size}(\boldsymbol{Y})$
        \State $\boldsymbol{Y}^* \gets \text{Array}(T)$
        \State $ni \gets \text{init\_sample}(\boldsymbol{W})$ \label{alg2_line5}
        
        \For{$t \gets 1$ \textbf{to} $T$} \label{alg2_line6}            
            \State $min\_range \gets quantiles[ni-1]$ \label{alg2_line7}
            \State $max\_range \gets quantiles[ni]$ \label{alg2_line8}
            \State $\boldsymbol{X}[t] \gets \text{runif}(min\_range, max\_range)$ \label{alg2_line9}
            \State $ni \gets \text{select\_next\_quantil}(\boldsymbol{W}[ni])$ \label{alg2_line10}
        \EndFor
        
        \State \Return $\boldsymbol{Y}^*$ \label{alg2_line12}
	\EndProcedure
    \end{algorithmic}
\end{algorithm}

\newpage

An important advantage of this synthetic time series generation method is its ability to produce multiple distinct realizations that preserve the original data’s dynamic and statistical characteristics. This diversity arises from the probabilistic elements in Algorithm~\ref{alg:re_construct_series}: the Markov transition matrix governs the selection of each quantile \(q_i\), and each value \(Y_t\) is sampled uniformly between the lower and upper bounds of the chosen quantile. By generating a richer ensemble of series, we can augment the dataset and thereby enhance the robustness and accuracy of downstream analytical tasks.

Our source code for both the QG and InvQG algorithms are available on GitHub: \url{https://github.com/vanessa-silva/InvQG}.

\paragraph{Computational Complexity}

The computational complexity of the InvQG method is dominated by the construction of the quantile graph and the generation of the synthetic time series. The QG construction (see Algorithm~\ref{alg:qg}) requires a single pass through the time series, assigning each timestamp to its corresponding quantile, resulting in $\mathcal{O}(T)$ complexity. The transition matrix has size $Q \times Q$, leading to an additional $\mathcal{O}(Q^2)$ cost. The synthetic series generation is also linear in $T$. 
Overall, the InvQG method has a computational complexity of $\mathcal{O}(Q T + Q^2)$, which simplifies to $\mathcal{O}(Q T)$ in practice, since $Q \ll T$.

In contrast, GAN-based approaches such as TimeGAN and DoppelGANger require iterative training procedures involving multiple epochs of neural network optimization, which significantly increases computational cost. Their training process involves repeated forward and backward passes over the data across several epochs, with cost scaling with the number of training sequences, the length of the time series, and the model capacity. While these factors depend on the dataset size and the complexity of the neural architecture, the factor $Q$ in InvQG is user-defined and typically small. Therefore, InvQG provides a more computationally efficient alternative, particularly in scenarios where fast data generation is required. In particular, GAN-based methods require a full training phase, whereas InvQG operates without iterative optimization. Moreover, the absence of a training phase makes the computational cost of InvQG predictable, depending only on the length of the input time series and the number of quantile levels, and not on the number of training sequences, the number of training epochs, or the model capacity.

\section{Empirical Evaluation}\label{sec4}

This section evaluates the fidelity and utility of synthetic time series generated by the Inverse Quantile Graph (InvQG) framework. We assess how well synthetic series reproduce the statistical, temporal, and structural properties of the original data, and whether they remain useful for downstream analytical tasks. The evaluation is conducted on simulated time series --- where ground‑truth properties are known -- and real-world smart‑meter data. We combine three complementary perspectives: (i) statistical feature preservation, (ii) network‑based topological similarity, and (iii) comparative visual and performance against state-of-the-art GAN-based generators.

\subsection{Methodology}

The empirical evaluation follows a structured and paired design to assess both the fidelity and utility of synthetic time series generated by the proposed InvQG framework, as well as to benchmark its performance against alternative generative approaches.
For each original time series $Y_i$ in a given dataset, a corresponding synthetic series $Y_i^*$ is generated using the QG and InvQG mappings. All analyses are performed on paired \textit{original}$–$\textit{synthetic} samples to enable a direct and controlled comparison. The evaluation comprises four complementary components:
\begin{itemize}
\item \textbf{Statistical Fidelity:} assessed via a set of interpretable time‑series features extracted with the \texttt{tsfeatures} R package~\cite{tsfeatures}, capturing distributional characteristics, such as trend, linearity, entropy, and short‑ and medium‑term autocorrelation structure.
\item \textbf{Utility via Network Representations:} evaluated using topological features derived from time‑series networks computed with the \textit{NetF} framework~\cite{silva2022novel}, as well as clustering and classification performance based on these features.
\item \textbf{Benchmarking against Generative Models:} the performance of InvQG is compared with two state‑of‑the‑art GAN‑based methods, \textit{TimeGAN} and \textit{DoppelGANger}, using qualitative and distributional similarity analyses. The execution time required for synthetic data generation across methods was also evaluated. 
\item \textbf{Robustness:} by analyzing the sensitivity of InvQG mapping to different numbers of quantiles, the single parameter $Q$.
\end{itemize}

\begin{figure*}[ht]
    \centering
        \includegraphics[width=\linewidth]{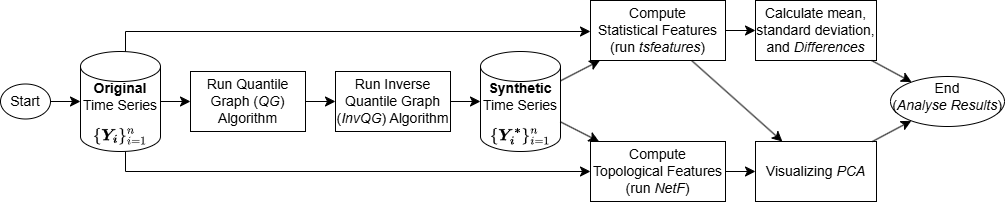}
    \caption{Overview of the evaluation methodology adopted to assess the fidelity, utility, and computational efficiency of synthetic time series generated with InvQG.}
    \label{fig:diag_meth}
\end{figure*}

This unified methodology ensures a fair, reproducible, and multi‑perspective assessment of InvQG, allowing us to identify both its strengths and its limitations across different data regimes. 
 The evaluation workflow is summarized in Figure~\ref{fig:diag_meth}.

For reproducibility purposes, the main source code and the datasets are made available in \url{https://github.com/vanessa-silva/InvQG}.

\subsection{Datasets}\label{sec:datasets}

Two distinct time series datasets are considered:
1) the \textit{artificial dataset}, consisting of simulated time series models designed to provide a controlled environment for methodological evaluation; and 
2) the \textit{real dataset}, consisting of smart meter readings that reflect real usage patterns and inherent variability. 

\subsubsection{Artificial Dataset}\label{sec:simul_data}

The artificial time‑series models were deliberately selected to represent a broad spectrum of temporal behaviors, including short‑ and long‑range dependence, periodicity, non‑stationarity, and regime‑switching dynamics. This diversity enables not only an assessment of the overall performance of the proposed method, but also a structured evaluation of its ability to capture specific structural characteristics and to reveal its limitations.
To control for relevant time‑series properties, we generate the original series using eleven well‑known statistical time‑series models, summarized  in Table~\ref{tab_tsmodels} (see also~\cite{silva2022novel} for further details). For each model, we simulate $n=100$ time series of length $T=10,000$ resulting in a total of  $1,100$ series. We refer to this collection as the \textit{artificial dataset}.

\begin{table*}[h]

\centering
\caption[Statistical time series models]{Statistical time series models for the artificial dataset.}

\begin{tabular}{lllc}
    \hline
    \textbf{Model} & \textbf{Parameters} & \textbf{Main Property} & \textbf{Notation} \\
    \hline
    
    \rule{0pt}{12pt}White Noise & $\epsilon_{t} \sim N(0,1)$ & Noise effect &  \textbf{WN} \\
    
    \rule{0pt}{12pt}AR$(1)$ & $\phi_{1} = -0.5$ & Smooth oscillation & \textbf{AR1 -0.5} \\
     & $\phi_{1} = 0.5$ & Moderate persistence & \textbf{AR1 0.5} \\
     & $\phi_{1} = 0.9$ & High persistence & \textbf{AR1 0.9} \\
    
    \rule{0pt}{12pt}AR$(2)$ & $\phi_{1} = 1.5$, $\phi_{2} = -0.75$ & Pseudo-periodic & \textbf{AR2} \\
    
    \rule{0pt}{12pt}ARIMA$(1,1,0)$ & $\phi_{1} = 0.7$ & Stochastic trend & \textbf{ARIMA} \\
    
    \rule{0pt}{12pt}ARFIMA$(1,0.4,0)$ & $\phi_{1} = 0.9$ & Long memory effect & \textbf{ARFIMA} \\
    
    \rule{0pt}{12pt}GARCH$(1,1)$ & $\omega = 10^{-6}$, $\alpha_1 = 0.1$, & Persistent periods of high & \multirow{2}{*}{\textbf{GARCH}} \\
     & $\beta_1 = 0.8$ & or low volatility & \\
    
    \rule{0pt}{12pt}SETAR$(1)$ & $\alpha = 0.5$, $\beta = -1.8$, $\gamma = 2$, & Regime-dependent  & \multirow{2}{*}{\textbf{SETAR}} \\
     & $r = -1$ & autocorrelation & \\
    
    \rule{0pt}{12pt}Poisson-HMM & $N = 2$, {$\bigl[ \begin{smallmatrix} 0.9 & 0.1\\ 0.1 & 0.9 \end{smallmatrix} \bigr]$} $\lambda \in \{10, 15\}$ & State transitions & \textbf{HMM} \\
     
    \rule{0pt}{12pt}INAR$(1)$ & $\alpha = 0.5$, $\epsilon_{t} \sim Po(1)$ & Correlated counts & \textbf{INAR} \\
    
    \hline
    \end{tabular}
    
\label{tab_tsmodels}
\end{table*} 

\subsubsection{Real Dataset}\label{sec:real_data}

We also employ an energy consumption dataset containing domestic electricity usage  from the East Midlands, UK~\cite{richardson2010electricity}~\footnote{Available at~\url{https://beta.ukdataservice.ac.uk/datacatalogue/studies/study?id=6583}}. The dataset consists of  smart meter readings from twenty-two different households, recorded at one-minute intervals over a two-year period (2008 and 2009). Each household exhibits a distinct consumption profile, providing substantial behavioral  diversity that poses a meaningful challenge for synthetic data generation and enables robust comparative evaluation. We will refer to this dataset as the \textit{real dataset}.

The preprocessing steps applied to the real dataset are summarized  below:
\begin{itemize}
    \item \textbf{Year Selection:} Although the dataset spans two years, we restrict the analysis to observations from 2008. Two of the twenty-two time series contain no measurements for 2009, and the remaining series exhibit a higher proportion of missing values in that year. Limiting the study to 2008 therefore yields a more stable and complete dataset without affecting the objectives of the analysis.

    \item \textbf{Common Time Window:} The household time series do not share identical start and end timestamps. To ensure a temporal comparability across units, we retain only the time window common to all twenty-two series, spanning from 2008-01-07 to 2008-07-02.

    \item \textbf{Time Aggregation:} Given the high temporal resolution of the original data (one-minute intervals), the time series are aggregated to  hourly frequency by summing the sixty-one-minute readings within each hour. 
    This results in a homogeneous collection of twenty-two hourly series, each of length $T = 4,262$. 
     While this aggregation reduces temporal granularity, it substantially improves computational efficiency and facilitates comparison with GAN-based generative methods, which are  computationally demanding. 
     
\newpage
    \item \textbf{Missing Data Handling:} After aggregation, the  hourly series still contain missing values, corresponding to hours with incomplete minute-level observations. 
    The proportion of  missing values varies across households, ranging from $0.02\%$ to $15.61\%$. 
    These missing data are imputed using a standard method based on seasonal decomposition and moving-average interpolation, implemented via the \texttt{na\_seasplit} function from the \texttt{imputeTS} package in R~\cite{imputeTS}. 
    This approach decomposes the series into seasonal components (intra-daily) and performs imputation within each component, making it well suited for  energy consumption data characterized by strong intra-daily seasonality.
 
\end{itemize}

\subsection{Fidelity and Utility Analysis}

In this section, we empirically analyze the fidelity and utility of the InvQG framework using both statistical and topological network features.

\subsubsection{Fidelity Analysis via Statistical Features}
\label{sec_fidelity}

We begin by assessing the statistical fidelity of synthetic time series generated with InvQG using the artificial dataset, which provides a controlled environment with known generative properties. Statistical features are extracted from both original and synthetic series using the \texttt{tsfeatures} R package~\cite{tsfeatures}, and all analyses are performed on paired differences (\textit{synthetic} $-$ \textit{original}). 

\begin{figure}[H]
  \includegraphics[width=\linewidth]{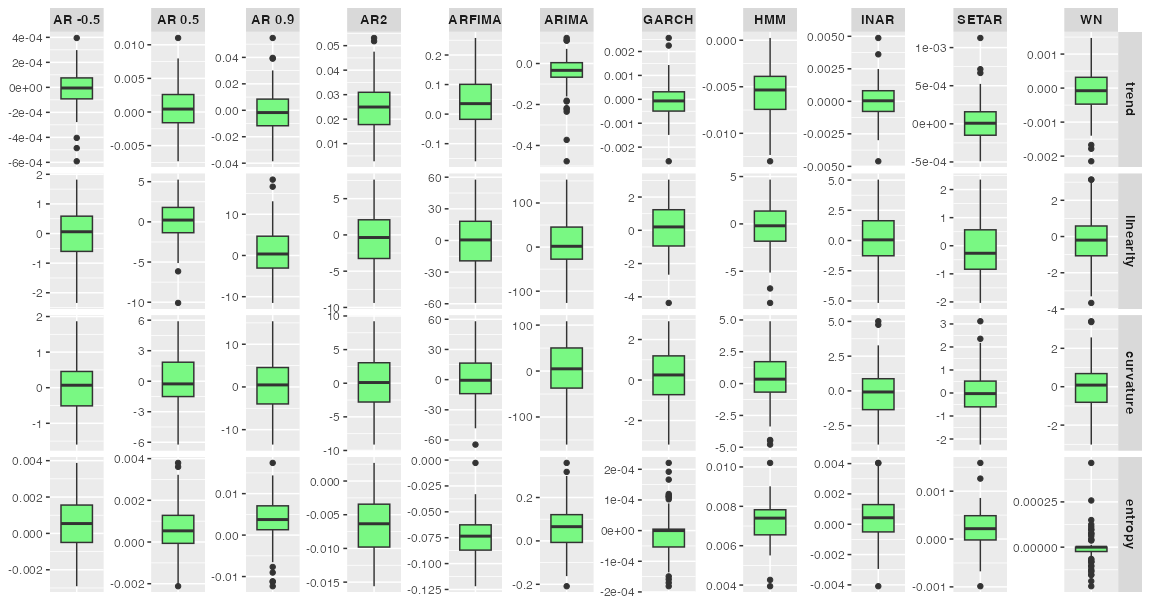}
  \caption{Boxplots of paired differences (synthetic - original) for \textit{trend}, \textit{linearity}, \textit{curvature} and \textit{entropy}.}
  \label{fig:boxplot_diff_hynmand_1}
\end{figure}

\begin{figure}[H] 
\includegraphics[width=\linewidth]{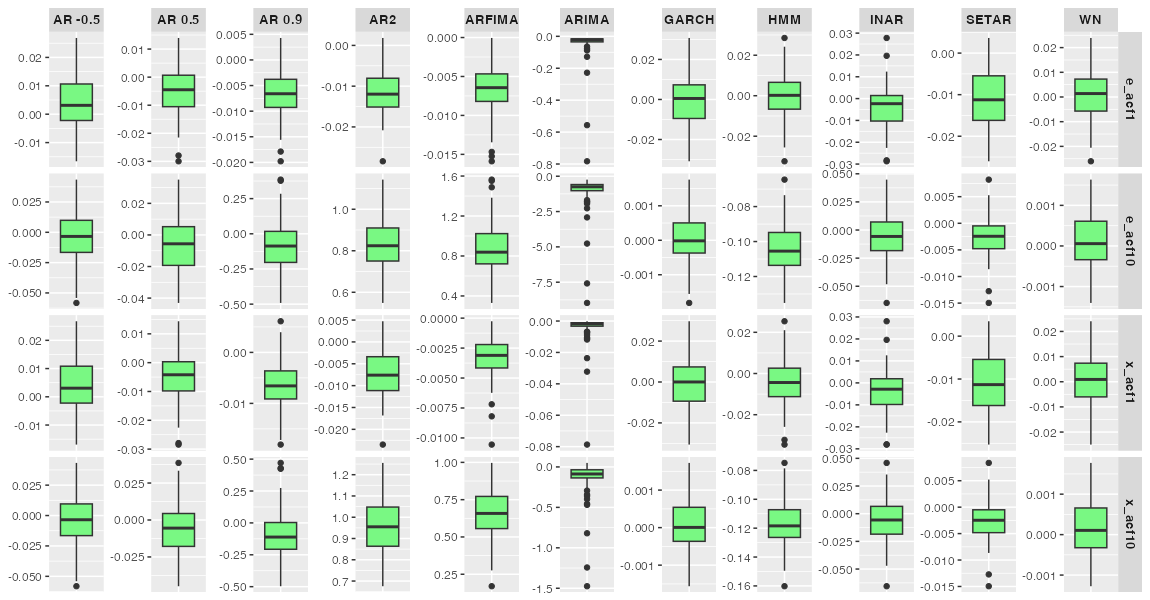}
  \caption{Boxplots of paired differences (synthetic - original) for \textit{e\_acf1}, \textit{e\_acf10}, \textit{x\_acf1} and \textit{x\_acf10}.}
  \label{fig:boxplot_diff_hynmand_2}
\end{figure}

Figures~\ref{fig:boxplot_diff_hynmand_1} and~\ref{fig:boxplot_diff_hynmand_2} present boxplots of the paired differences for statistical features, while Table~\ref{tab:mean_sd_diffs} reports their means and standard deviations.

Across most models and features, the distributions of the paired differences are centered around zero, indicating that the InvQG preserves the principal statistical characteristics of the original series. This is particularly evident for trend‑related measures, entropy, and short‑term autocorrelation (lag 1), demonstrating that the method effectively captures marginal distributions and immediate temporal dependencies across a broad range of time series models.

\begin{figure}[H]
    \centering
    \begin{subfigure}{0.5\textwidth}
        \centering
        \includegraphics[width=\linewidth]{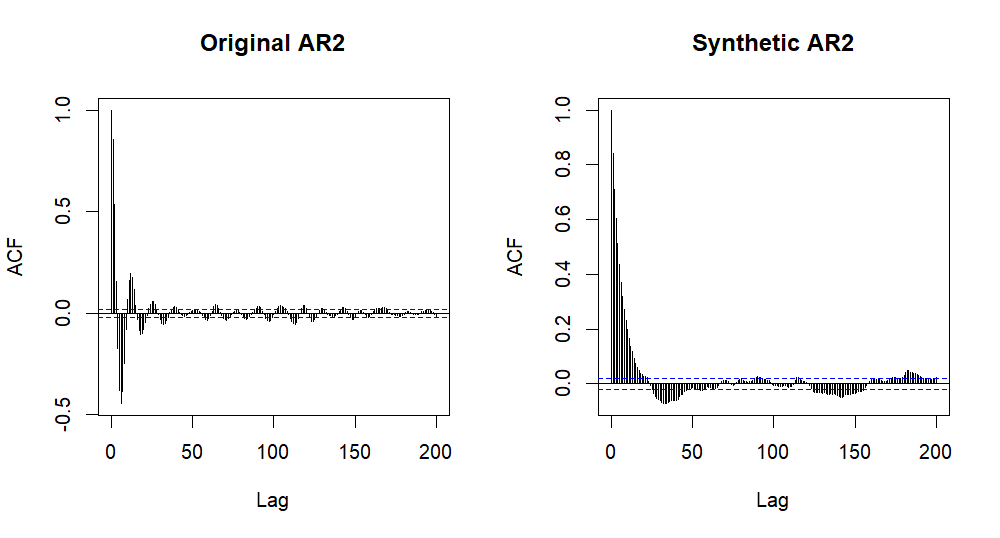}
        \caption{ACF}
        \label{fig:acf_plot_ar2}
    \end{subfigure}
    \hfill
    \begin{subfigure}{0.45\textwidth}
        \centering
        \includegraphics[width=\linewidth]{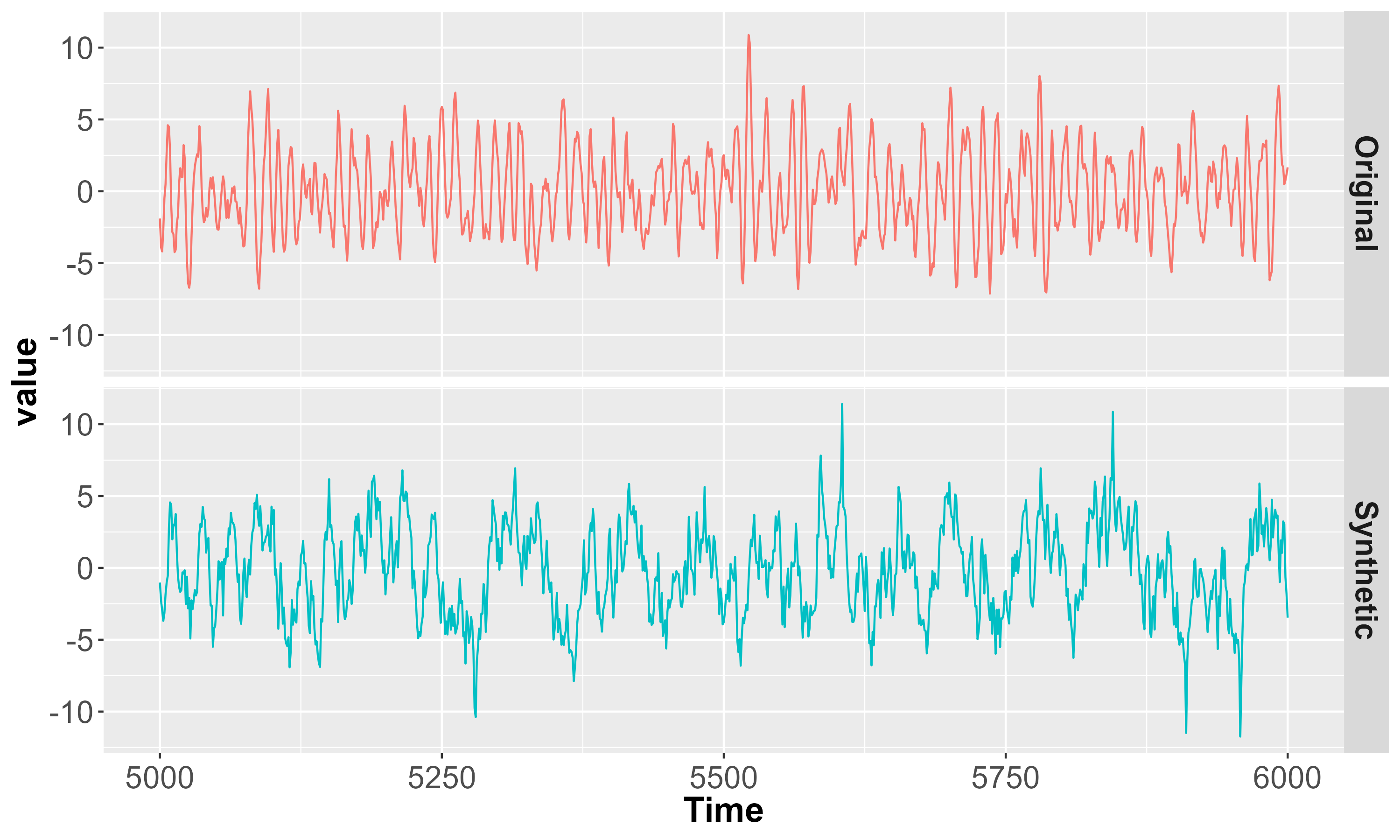}
        \caption{Time Series}
        \label{fig:ts_plot_ar2}
    \end{subfigure}
    \caption{Comparison between an original and synthetic AR2 time series. (a) Autocorrelation function (ACF). (b) Time series (first 1000 observations).}
    \label{fig:acf_ts_plot_ar2}
\end{figure}

\begin{figure}[H]
    \centering
    \begin{subfigure}{0.5\textwidth}
        \centering
        \includegraphics[width=\linewidth]{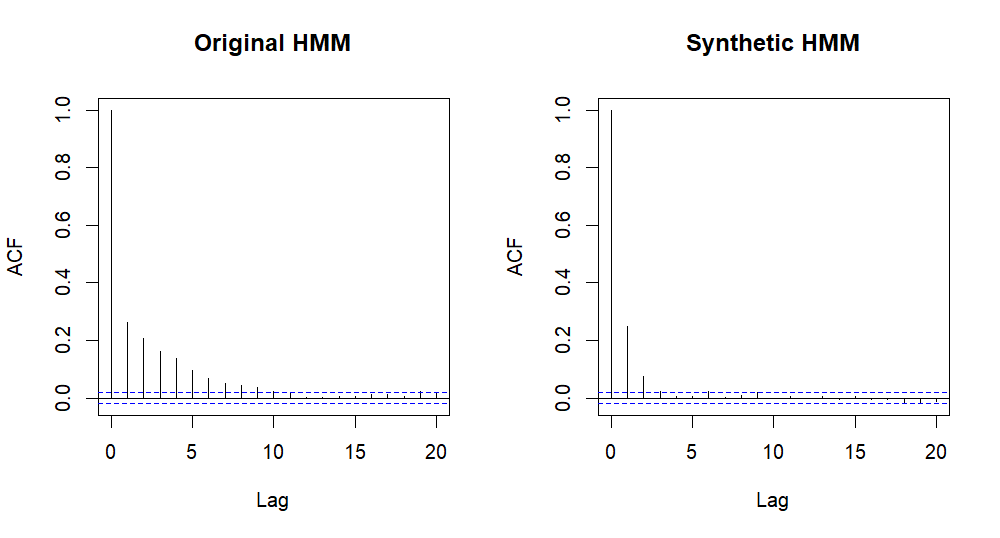}
        \caption{ACF}
        \label{fig:acf_plot_hmm}
    \end{subfigure}
    \hfill
    \begin{subfigure}{0.45\textwidth}
        \centering
        \includegraphics[width=\linewidth]{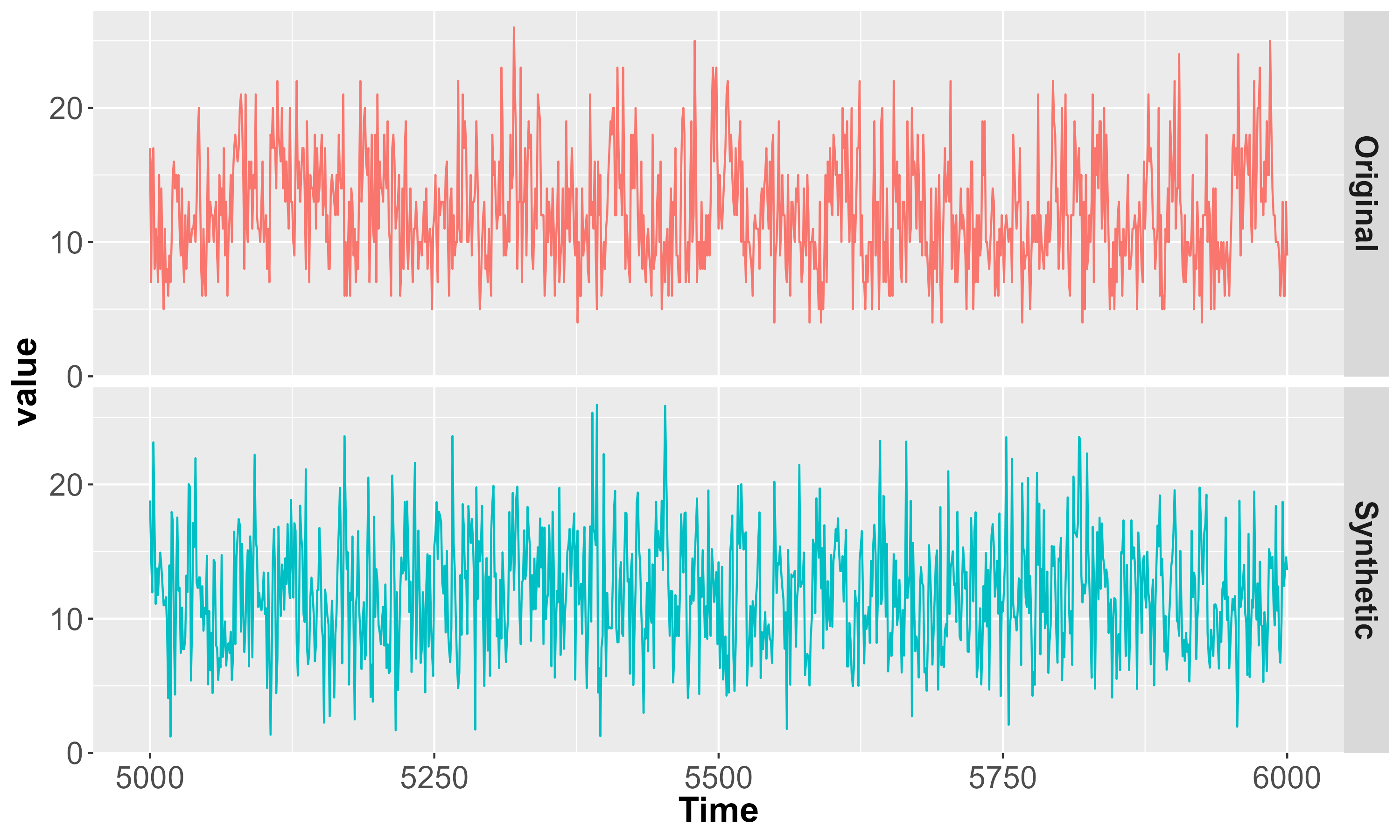}
        \caption{Time Series}
        \label{fig:ts_plot_hmm}
    \end{subfigure}
    \caption{Comparison between an original and synthetic HMM time series. (a) Autocorrelation function (ACF). (b) Time series (first 1000 observations).}
    \label{fig:acf_ts_plot_hmm}
\end{figure}

\begin{figure}[H]
    \centering
    \begin{subfigure}{0.45\textwidth}
        \centering
        \includegraphics[width=\linewidth]{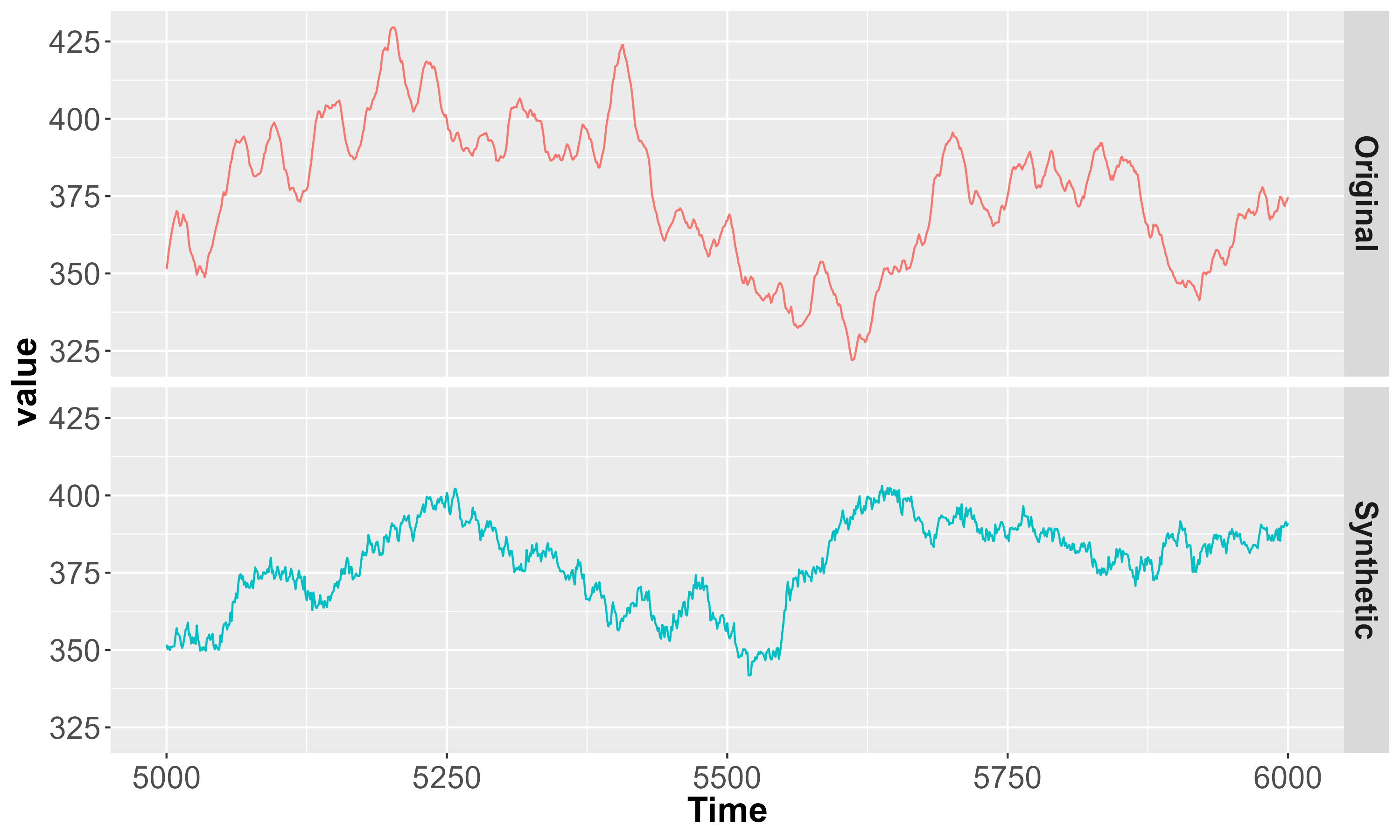}
        \caption{\textbf{ARIMA}}
    \end{subfigure}
    \hfill
    \begin{subfigure}{0.45\textwidth}
        \centering
        \includegraphics[width=\linewidth]{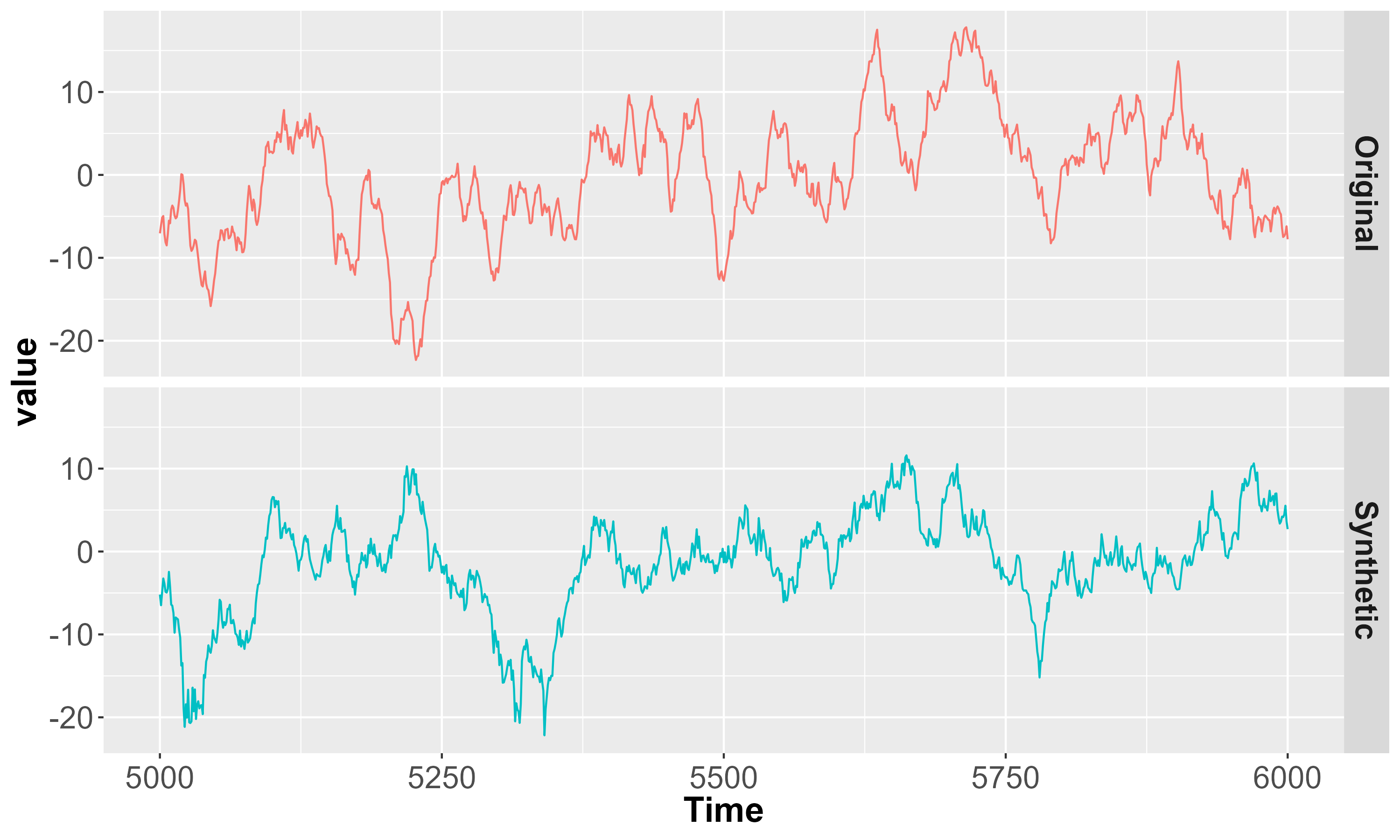}
        \caption{\textbf{ARFIMA}}
    \end{subfigure}
    \caption{Comparison between original and synthetic time series for (a) ARIMA and (b) ARFIMA models (first 1000 observations).}
    \label{fig:orig_vs_synth_arima_arfima}
\end{figure}

Systematic deviations emerge primarily for models characterized  by strong periodicity or long‑range dependence, notably AR2, ARFIMA, ARIMA, and HMM processes. These differences are most pronounced in autocorrelation‑based features at higher lags (e.g., lag 10), as shown in Figure~\ref{fig:boxplot_diff_hynmand_2} and quantified in Table~\ref{tab:mean_sd_diffs}. The underlying cause is structural: InvQG relies on a first‑order Markov transition model and therefore encodes dependencies only between consecutive observations. As a result, higher‑order cyclic behavior  (AR2) and slow memory decay (ARFIMA) are partially smoothed in the synthetic reconstructions.
These effects are illustrated by the representative examples in Figures~\ref{fig:acf_plot_ar2} - ~\ref{fig:orig_vs_synth_arima_arfima}. For AR2 series, the synthetic counterparts reproduce persistence but fail to fully recover the alternating cyclic structure present in the original autocorrelation function. Similarly, ARFIMA synthetic series exhibit a smoother, more regular decay of autocorrelation compared to the fractional decay observed in the originals. In the case of HMM‑generated data, regime‑switching behavior  is preserved, but state persistence is slightly reduced, leading to attenuated higher‑lag autocorrelations.

\newpage

\newpage

\newpage

To formally assess these differences, we applied paired Wilcoxon signed-rank tests to the feature differences for each model with false discovery rate (FDR) correction for multiple comparisons. 
The results (see Tables~\ref{tab:summ_stat_tests} and~\ref{app_tab:stat_tests}) confirm that statistically significant differences are largely confined to autocorrelation‑related features in models with strong temporal dependence. In contrast, for weakly dependent or memoryless processes --- such as WN, GARCH, and INAR --- most statistical features show no significant differences between original and synthetic series. These findings are consistent with the descriptive analysis and reinforce that InvQG preserves short‑term temporal structure while exhibiting predictable limitations for long‑range dependence.

\begin{table*}[!h]
\centering
{
\caption{Wilcoxon signed-rank test results per model and feature. Each cell shows the median difference (\textit{synthetic - original}) and the FDR-corrected $p$-value, in brackets. Asterisks (*) indicate statistically significant differences ($p_{\text{FDR}} < 0.05$).}
\centering
\scriptsize
\begin{tabular}[t]{lcccccccc}

\toprule

& \multicolumn{8}{c}{Median} \\
& \multicolumn{8}{c}{($p_{\text{FDR}} < 0.05$)} \\

\cmidrule(lr){2-9}
\multicolumn{1}{c}{\textbf{Model}} & \multicolumn{1}{c}{\textbf{trend}} & \multicolumn{1}{c}{\textbf{linea.}} & \multicolumn{1}{c}{\textbf{curvat.}} & \multicolumn{1}{c}{\textbf{entro.}} & \multicolumn{1}{c}{\textbf{e\_acf1}} & \multicolumn{1}{c}{\textbf{e\_acf10}} & \multicolumn{1}{c}{\textbf{x\_acf1}} & \multicolumn{1}{c}{\textbf{x\_acf10}} \\
\midrule
\multirow{2}{*}{\textbf{AR1 -0.5}} & -0.000 & 0.060 & 0.070 & 0.001 & 0.003 & -0.003 & 0.003 & -0.003\\
 & \footnotesize(0.872) & \footnotesize(0.687) & \footnotesize(0.609) & \footnotesize(8.05e-04)* & \footnotesize(8.96e-05)* & \footnotesize(0.089) & \footnotesize(8.96e-05)* & \footnotesize(0.089)\\
\midrule
\multirow{2}{*}{\textbf{AR1 0.5}} & 0.000 & 0.224 & -0.263 & 0.001 & -0.004 & -0.006 & -0.004 & -0.005\\
 & \footnotesize(0.192) & \footnotesize(0.629) & (0.974) & \footnotesize(7.71e-06)* & \footnotesize(8.90e-07)* & \footnotesize(9.43e-04)* & \footnotesize(1.40e-06)* & \footnotesize(0.002)*\\
\midrule
\multirow{2}{*}{\textbf{AR1 0.9}} & -0.002 & 0.360 & 0.475 & 0.004 & -0.007 & -0.087 & -0.007 & -0.111\\
 & \footnotesize(0.482) & \footnotesize(0.408) & \footnotesize(0.518) & \footnotesize(2.40e-09)* & \footnotesize(5.06e-16)* & \footnotesize(2.29e-05)* & \footnotesize(5.57e-16)* & \footnotesize(1.06e-05)*\\
\midrule
\multirow{2}{*}{\textbf{AR2}} & 0.025 & -0.358 & 0.104 & -0.006 & -0.012 & 0.824 & -0.008 & 0.955\\
 & \footnotesize(2.39e-17)* & \footnotesize(0.319) & \footnotesize(0.629) & \footnotesize(5.36e-17)* & \footnotesize(3.52e-17)* & \footnotesize(2.39e-17)* & \footnotesize(2.24e-15)* & \footnotesize(2.39e-17)*\\
\midrule
\multirow{2}{*}{\textbf{ARFIMA}} & 0.035 & 0.516 & -0.642 & -0.073 & -0.006 & 0.838 & -0.003 & 0.658\\
 & \footnotesize(1.03e-04)* & \footnotesize(0.988) & \footnotesize(0.893) & \footnotesize(2.39e-17)* & \footnotesize(2.39e-17)* & \footnotesize(2.39e-17)* & \footnotesize(2.39e-17)* & \footnotesize(2.39e-17)*\\
\midrule
\multirow{2}{*}{\textbf{ARIMA}} & -0.033 & 1.808 & 4.828 & 0.066 & -0.021 & -0.750 & -0.002 & -0.087\\
 & \footnotesize(3.86e-07)* & \footnotesize(0.556) & \footnotesize(0.401) & \footnotesize(2.84e-07)* & \footnotesize(2.39e-17)* & \footnotesize(2.39e-17)* & \footnotesize(2.39e-17)* & \footnotesize(1.26e-15)*\\
\midrule
\multirow{2}{*}{\textbf{GARCH}} & -0.000 & 0.203 & 0.251 & 0.000 & 0.000 & -0.000 & 0.000 & 0.000\\
 & \footnotesize(0.470) & \footnotesize(0.475) & \footnotesize(0.289) & \footnotesize(0.192) & \footnotesize(0.637) & \footnotesize(0.496) & \footnotesize(0.622) & \footnotesize(0.482)\\
\midrule
\multirow{2}{*}{\textbf{HMM}} & -0.005 & -0.199 & 0.350 & 0.007 & 0.000 & -0.105 & -0.004 & -0.118\\
 & \footnotesize(2.39e-17)* & \footnotesize(0.408) & \footnotesize(0.163) & \footnotesize(2.39e-17)* & \footnotesize(0.974) & \footnotesize(2.39e-17)* & \footnotesize(3.94e-04)* & \footnotesize(2.39e-17)*\\
\midrule
\multirow{2}{*}{\textbf{INAR}} & 0.000 & 0.055 & -0.078 & 0.000 & -0.002 & -0.006 & -0.003 & -0.006\\
 & \footnotesize(0.836) & \footnotesize(0.579) & \footnotesize(0.605) & \footnotesize(0.002)* & \footnotesize(5.11e-04)* & \footnotesize(0.010)* & \footnotesize(5.19e-04)* & \footnotesize(0.011)*\\
\midrule
\multirow{2}{*}{\textbf{SETAR}} & 0.000 & -0.267 & -0.027 & 0.000 & -0.011 & -0.002 & -0.011 & -0.002\\
 & \footnotesize(0.871) & \footnotesize(0.163) & \footnotesize(0.836) & \footnotesize(3.19e-07)* & \footnotesize(6.61e-17)* & \footnotesize(3.72e-09)* & \footnotesize(7.95e-17)* & \footnotesize(2.86e-09)*\\
\midrule
\multirow{2}{*}{\textbf{WN}} & -0.000 & -0.198 & 0.084 & 0.000 & 0.001 & 0.000 & 0.001 & 0.000\\
 & \footnotesize(0.293) & \footnotesize(0.172) & \footnotesize(0.637) & \footnotesize(0.220) & \footnotesize(0.482) & \footnotesize(0.148) & \footnotesize(0.518) & \footnotesize(0.085)\\
\bottomrule
 \multicolumn{9}{l}{\textsuperscript{}* $p_{FDR} < 0.05$ (Wilcoxon signed-rank test, Benjamini-Hochberg correction, 88 tests).}\\
\label{tab:summ_stat_tests}
\end{tabular}
}
\end{table*}

\begin{figure}[!h]
\centering
  \includegraphics[width=0.94\linewidth]{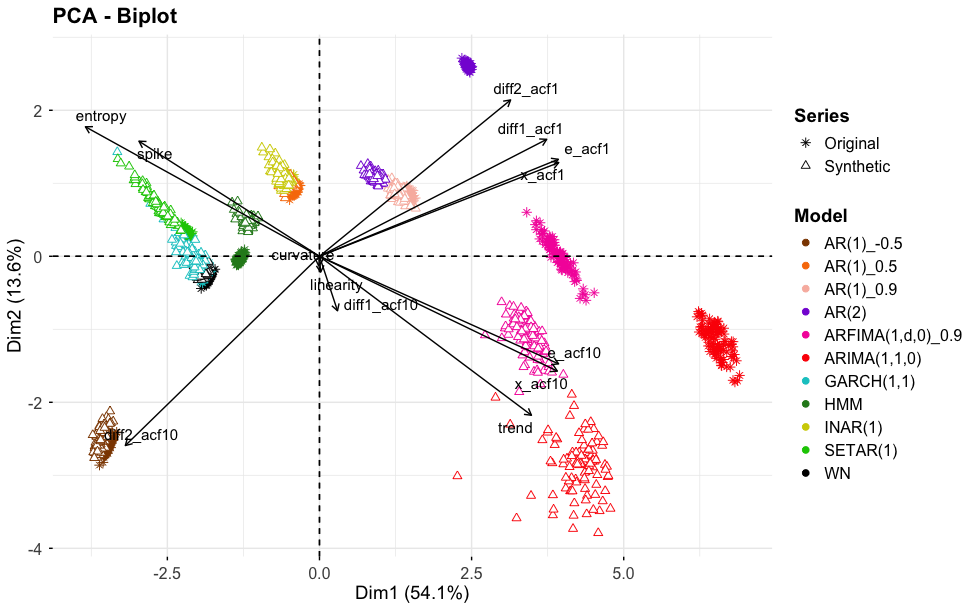}
  \caption[PCA of statistical features (\textit{tsfeatures}) for all models]{PCA projection of statistical features for original (asterisks) and synthetic (triangles) time series, coloured by model. The first two principal components explain 67.7\% of the variance.}
  \label{fig:hyndman_all_models_pca}
\end{figure}

Finally, we examine the joint behavior  of all statistical features using principal component analysis (PCA). Figure~\ref{fig:hyndman_all_models_pca} shows that the original and synthetic series generated from the same model cluster closely together and remain well separated from other models. Although some separation between original and synthetic samples is visible for AR2, ARIMA, and ARFIMA processes, their relative positions in the feature space are preserved. Importantly, the same features drive the separation of both original and synthetic data, indicating that InvQG maintains the statistical identity of each model and supports downstream analytical tasks.

\newpage

Overall, this analysis shows that InvQG provides high statistical fidelity for short‑ and medium‑range temporal properties across diverse time‑series models. Deviations are concentrated in settings where long‑range or higher‑order dependencies dominate, reflecting intrinsic constraints of the first‑order transition formulation rather than instability or lack of robustness of the method.

While the previous analysis characterizes  the statistical fidelity of InvQG in absolute terms, we next assess its practical utility and its performance relative to established deep generative models.

\subsubsection{Utility Analysis via Topological Network Features}

We also evaluate whether the synthetic time series generated by InvQG remain useful for downstream analytical tasks. To this end, we assess utility through network-based representations, examining whether synthetic series preserve the structural characteristics required for meaningful data analysis.
We employ the \textit{NetF} framework~\cite{silva2022novel}, which maps each time series into a set of complex networks --- namely weighted natural visibility graphs (WNVG), weighted horizontal visibility graphs (WHVG), and quantile graphs (QG) -- and extracts corresponding topological feature vectors. These features provide an alternative representation of temporal structure and have been shown to be effective in tasks such as clustering and time series characterization~\cite{silva2022novel}.

\paragraph{Topological Feature Consistency}

\begin{figure*}[!h]
\centering
  \includegraphics[width=0.94\linewidth]{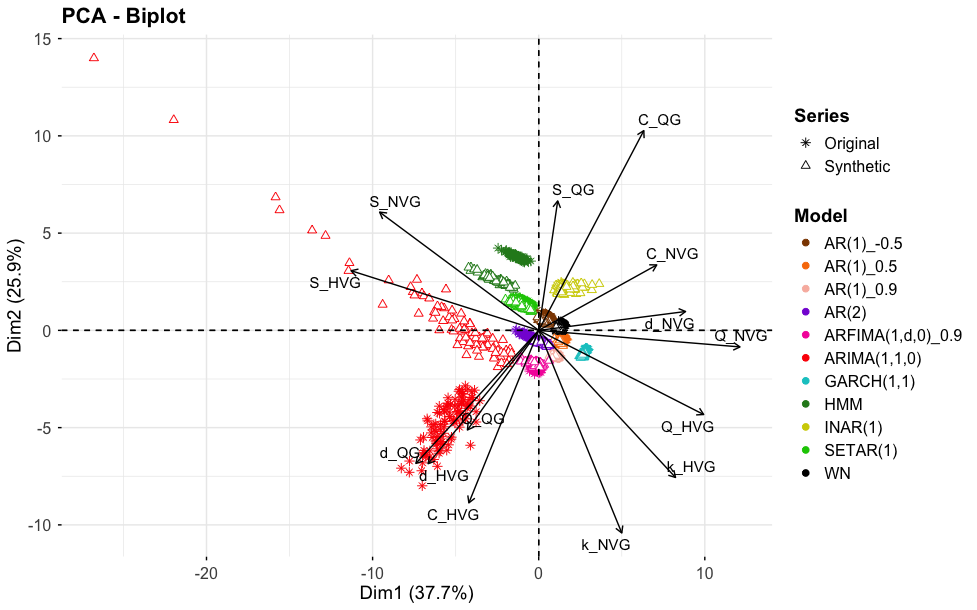}
\caption[PCA projection of network-based features (NetF) for original (asterisks) and synthetic (triangles) time series, coloured by model.]{PCA projection of network-based features (NetF) for original (asterisks) and synthetic (triangles) time series, coloured by model. The first two principal components explain 63.4\% of the variance.}  
  \label{fig:netf_all_models_pca}
\end{figure*}

We first analyze the similarity between original and synthetic data in the space of network derived features. Figure~\ref{fig:netf_all_models_pca} presents a PCA projection of these features for both datasets. The results are broadly consistent with those obtained using statistical features (Figure~\ref{fig:hyndman_all_models_pca}): each time series model forms a distinct cluster, and synthetic samples are generally located within, or in close proximity to, the corresponding original clusters.

For most models, this indicates that InvQG preserves the structural signatures captured by network representations, which are essential for distinguishing between different types of temporal behavior. Some separation between original and synthetic samples is observed for models such as ARIMA, AR2, and HMM; however, the synthetic data remain clearly associated with their respective model classes. Notably, for ARFIMA models, the network-based representation shows stronger overlap between original and synthetic data than the statistical features, suggesting that certain structural aspects are better captured in the network domain.

A closer examination of individual topological characteristics reveals that global properties such as modularity and average degree are well preserved in the synthetic networks. In contrast, more sensitive feature such as clustering coefficient, average path length, and community structure, exhibit greater variability. These discrepancies can be attributed to the sensitivity of network mappings to local amplitude variations and subtle changes in temporal dependence, which may alter visibility relations or quantile transitions. As discussed earlier, such variations are consistent with the first order transition modelling underlying InvQG and primarily affect higher order or long-range structural properties.

\paragraph{Clustering-Based Utility}

To provide a quantitative assessment of utility, we perform a clustering analysis using the \textit{NetF} feature representation. Feature vectors from both original and synthetic data are combined, and clustering is conducted across a range of cluster numbers using the K-means algorithm. Performance is evaluated using the Average Silhouette Score (AS), and metrics relative to the known model labels, the Adjusted Rand Index (ARI) and Normalized Mutual Information (MNI).
The results, reported in Table~\ref{tab:clustering}, indicate that clustering performance on the combined original-synthetic dataset remains strong and close to that obtained using original data alone. While performance metrics do not reach the levels achieved when clustering only original samples --- reflecting the previously observed discrepancies for certain models --- the degradation is limited. Importantly, the optimal number of clusters and the overall cluster structure remain consistent with the underlying data-generating processes.
These findings demonstrate that the synthetic time series preserve the discriminative structure necessary for unsupervised learning tasks. Although some separation between original and synthetic samples persists within certain model classes, this does not prevent correct grouping at the model level, indicating that the essential information required for clustering is retained.

\begin{table*}[!h]
\centering
\begin{tabular}{|c|c|c|c|c|c|}
\hline
\multirow{2}{*}{\textbf{Feature Set}}& \multirow{2}{*}{\textbf{Source Data}} & \textbf{Num. of} & \textbf{AS} & \textbf{ARI} & \textbf{NMI}  \\
            &  & \textbf{Clusters} & {\tiny $[-1,1]$} & {\tiny$[0,1]$} & {\tiny$[-1,1]$}  \\ 
            \hline
            \hline
\multirow{3}{*}{\textbf{\textit{NetF}}} & original + synthetic & $\mathbf{k^* = 13}$  & $0.69 \pm 0.02$ & $0.82 \pm 0.05$ & $0.91 \pm 0.02$ \\
& original + synthetic & $\mathbf{k = 11}$  & $0.66 \pm 0.02$ & $0.76 \pm 0.03$ & $0.88 \pm 0.02$ \\ 
& original & $\mathbf{k = 11}$  & $0.69 \pm 0.03$ & $0.93 \pm 0.05$ & $0.97 \pm 0.03$ \\ 
\hline
\hline
\multirow{3}{*}{\textbf{\textit{tsfeatures}}} & original + synthetic & $\mathbf{k^* = 10}$  & $0.69 \pm 0.02$ & $0.71 \pm 0.05$ & $0.84 \pm 0.03$ \\
& original + synthetic & $\mathbf{k = 11}$  & $0.69 \pm 0.02$ & $0.69 \pm 0.05$ & $0.83 \pm 0.03$ \\
& original & $\mathbf{k = 11}$  & $0.77 \pm 0.02$ & $0.81 \pm 0.05$ & $0.91 \pm 0.03$ \\
\hline
\end{tabular}
\caption{Clustering evaluation metrics. The values reflect the mean and the standard deviation of 100 repetitions of the clustering analysis experiments.}
    \label{tab:clustering}
\end{table*}

Overall, the network-based analysis shows that InvQG-generated synthetic data remain informative and  useful for downstream tasks. The method successfully preserves the structural patterns that distinguish different classes of time series, even when minor discrepancies arise in more sensitive topological features. Consistent with the fidelity analysis, these deviations are concentrated in models with strong long-range or higher-order dependencies and reflect intrinsic modelling constraints rather than instability in the generation process.
These results confirm that InvQG achieves a moderate balance between fidelity and utility: the synthetic data not only resemble the original series in statistical terms but also retain the structural information required for meaningful analysis in realistic settings.

\paragraph{Train on Synthetic, Test on Real}

To further assess the utility of the synthetic data beyond clustering-based  evaluation, we now conducted a supervised classification experiment following the \textit{Train on Synthetic, Test on Real} (TSTR) paradigm. 
This approach aims evaluate whether synthetic data can serve as a substitute  for real data in downstream machine learning tasks. The experiment was also conducted using the two distinct feature representations: \textit{NetF} and \textit{tsfeatures}.

For each feature representation, we formulated a multi-class classification task in which the goal is to identify the generating statistical model (11 class models, as described in Table~\ref{tab_tsmodels}) from the feature representation of a  time series.  Three classical classifiers were considered in both cases: \textit{Random Forest} (RF), \textit{Support Vector Machine} (SVM, radial kernel), and \textit{K-Nearest Neighbours} (KNN). 
The baseline scenario, \textit{Train on Real, Test on Real} (TRTR) was evaluated using 5-fold cross-validation repeated 3 times (15 folds total) on the original data, from which per-fold Accuracy (Acc) and F1-score (F1) were recorded to estimate mean and standard deviation. 
For the TSTR scenario, we performed 30 repeated stratified holdouts: in each repetition, 80\% of the synthetic data was sampled (preserving class proportions via stratified splitting) for training, the full set of 1,100 original time series was used for testing, the  classification performances  was measured using Acc and F1.

Table~\ref{tab:tstr_utility} reports the TRTR and TSTR results for all classifiers under both feature representations. 
For \textit{tsfeatures}, RF and KNN show negligible TSTR degradation ($\Delta$F1 $= -0.007$ and $-0.025$, respectively), while SVM exhibits a larger drop ($\Delta$F1 $= -0.143$), consistent with the known sensitivity of margin-based classifiers to subtle distributional shifts. Averaged across classifiers, degradation remains modest ($\Delta$Acc $= -0.058$, $\Delta$F1 $= -0.062$). 
For \textit{NetF}, the TRTR scenario yields perfect classification across all classifiers (Acc $= 1.000$), reflecting the highly discriminative nature of network-based features with respect to the generating model~\cite{silva2022novel}. In the TSTR scenario, KNN achieves near-perfect performance ($\Delta$Acc $= -0.002$), while RF and SVM show moderate degradation (F1 equal to $-0.072$ and $-0.075$, respectively). 
Overall, these results indicate that the synthetic data preserves a substantial portion of the discriminative structure across both feature representations. While some degradation is observed --- particularly for more sensitive classifiers such as SVM --- the overall performance remains reasonably strong, supporting the utility of the generated data for downstream tasks.

\begin{table*}[!h]
\centering
\caption{\label{tab:tstr_utility}Train on Synthetic, Test on Real (TSTR) \textit{vs.} Train on Real, Test on Real (TRTR) classification results using \textit{tsfeatures} (left) and network features -- \textit{NetF} (right). 
Each block reports the mean over the 30 repetitions of Accuracy (Acc) and macro-averaged F1-score (F1), and the corresponding degradation $\Delta$ (TSTR $-$ TRTR) for three classifiers. }
\footnotesize

\begin{tabular}[t]{lrrrrrrrrrrrr}
\toprule
\multicolumn{1}{c}{ } & \multicolumn{6}{c}{\textit{tsfeatures}} & \multicolumn{6}{c}{\textit{NetF}} \\
\cmidrule(lr){2-7} \cmidrule(lr){8-13}

\multicolumn{1}{c}{ } & \multicolumn{2}{c}{Accuracy} & \multicolumn{2}{c}{F1-score} & \multicolumn{2}{c}{$\Delta$} & 
\multicolumn{2}{c}{Accuracy} & \multicolumn{2}{c}{F1-score} & \multicolumn{2}{c}{$\Delta$} \\
\cmidrule(lr){2-3} \cmidrule(lr){4-5} \cmidrule(lr){6-7}
\cmidrule(lr){8-9} \cmidrule(lr){10-11} \cmidrule(lr){12-13}

Classifier & TRTR & TSTR & TRTR & TSTR & $\Delta$Acc & $\Delta$F1 
           & TRTR & TSTR & TRTR & TSTR & $\Delta$Acc & $\Delta$F1\\

\midrule

RF  & 0.900 & 0.893 & 0.899 & 0.892 & -0.007 & -0.007 
    & 1.000 & 0.914 & 1.000 & 0.928 & -0.086 & -0.072\\

SVM & 0.805 & 0.663 & 0.795 & 0.643 & -0.143 & -0.153 
    & 1.000 & 0.796 & 1.000 & 0.925 & -0.204 & -0.075\\

KNN & 0.843 & 0.818 & 0.842 & 0.817 & -0.025 & -0.025 
    & 1.000 & 0.998 & 1.000 & 0.998 & -0.002 & -0.002\\

\midrule

\textbf{Mean} 
& \textbf{0.849} & \textbf{0.791} & \textbf{0.845} & \textbf{0.784} & \textbf{-0.058} & \textbf{-0.062} 
& \textbf{1.000} & \textbf{0.903} & \textbf{1.000} & \textbf{0.950} & \textbf{-0.097} & \textbf{-0.050}\\

\bottomrule
\end{tabular}

\vspace{2mm}
\begin{minipage}{\textwidth}
\footnotesize
TRTR: 5-fold cross-validation repeated 3 times on real data. TSTR: trained on all synthetic data, tested on all real data. F1-macro averaged over 11 classes. RF: Random Forest; SVM: Support Vector Machine (radial kernel); KNN: K-Nearest Neighbours.
\end{minipage}

\end{table*}

\subsection{Benchmark and Robustness Analysis}

In this section, we perform a comparative visual and computational performance analysis of the InvQG framework against the benchmark GAN-based models, and present a robustness analysis by examining the sensitivity of InvQG to the number of quantiles, $Q$.

\subsubsection{Benchmarking against Generative Models}

We next compare the proposed InvQG framework against two state‑of‑the‑art generative models for time series synthesis --- TimeGAN and DoppelGANger --- in order to assess whether InvQG achieves comparable fidelity in reproducing the statistical and structural characteristics of the original data.

\paragraph{Artificial Dataset}
For each time series model in the artificial dataset, we select a representative original series and generate corresponding synthetic series using InvQG, TimeGAN, and DoppelGANger. To facilitate comparison across high dimensional observations, we apply t-SNE visualizations to the resulting data, where original observations are shown in red and synthetic observations in blue (Figures~\ref{fig:t_sne_time_gan_vs_qg} and~\ref{fig:t_sne_time_gan_vs_qg_2}).
Across most models, InvQG produces synthetic data whose empirical distributions closely overlap with those of the original series. This overlap is consistent across a wide range of temporal behaviors, indicating that InvQG effectively preserves both local dynamics and global distributional structure. Moreover, the synthetic samples generated by InvQG exhibit substantial diversity while remaining concentrated in the same regions of the embedded space as the original data, suggesting a balance between variability and fidelity.
In contrast, TimeGAN and DoppelGANger display less consistent behavior. TimeGAN frequently generates synthetic samples that form clusters partially separated from the original data, indicating difficulties in fully matching the underlying distribution. DoppelGANger often yields tighter clusters, suggesting reduced variability, and in several cases shows noticeable shifts relative to the original observations. These effects are particularly visible for models with pronounced temporal structure, such as ARIMA, ARFIMA, and HMM processes.
The main exception for InvQG is the AR2 model, where the cyclic behavior  driven by higher order dependencies is not fully recovered. As discussed in Section~\ref{sec_fidelity}, this reflects the intrinsic limitation of InvQG’s first order transition formulation rather than an artefact of the comparison.
Overall, these results indicate that InvQG achieves competitive distributional fidelity relative to the considered GAN-based approaches on controlled synthetic data, while avoiding the instability and mode collapse phenomena often associated with adversarial training.

\begin{figure}[!h]
\centering
\captionsetup[subfigure]{justification=centering}

% First row
\hspace*{\fill}
\begin{subfigure}[t]{0.18\textwidth}
  \includegraphics[width=\linewidth]{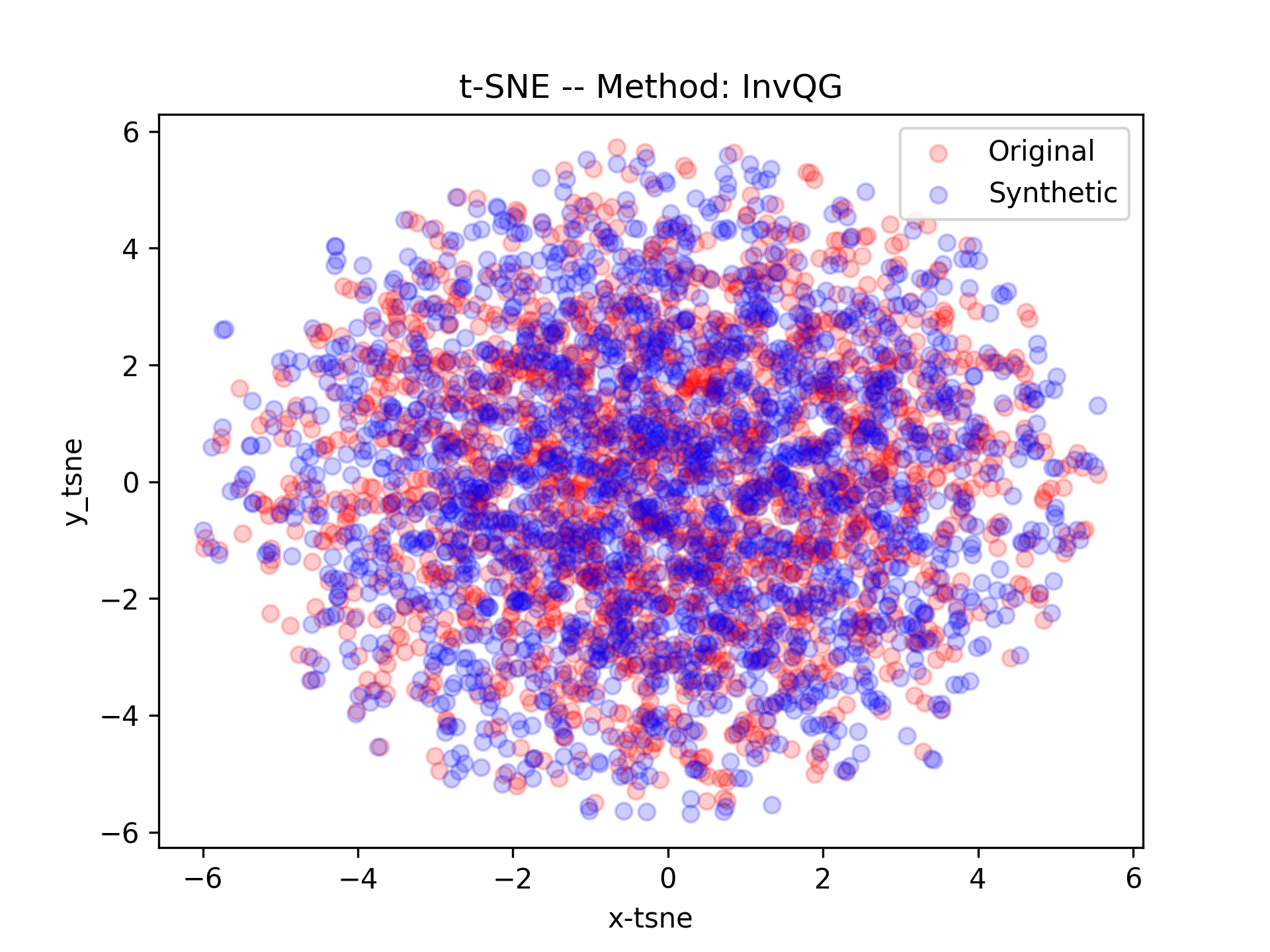}
\end{subfigure}
    \hfill
\begin{subfigure}[t]{0.18\textwidth}
  \includegraphics[width=\linewidth]{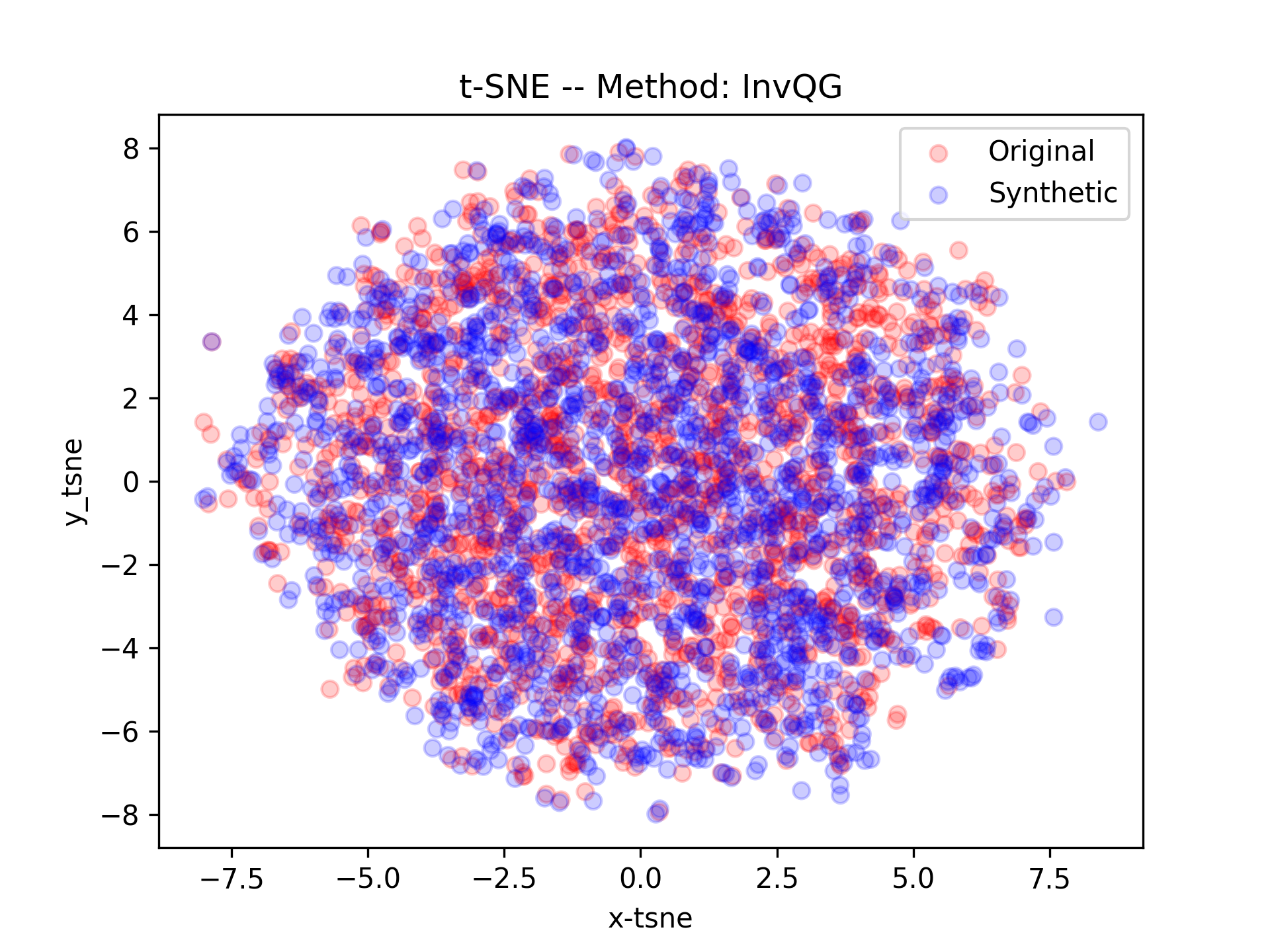}
\end{subfigure}
    \hfill 
\begin{subfigure}[t]{0.18\textwidth}
  \includegraphics[width=\linewidth]{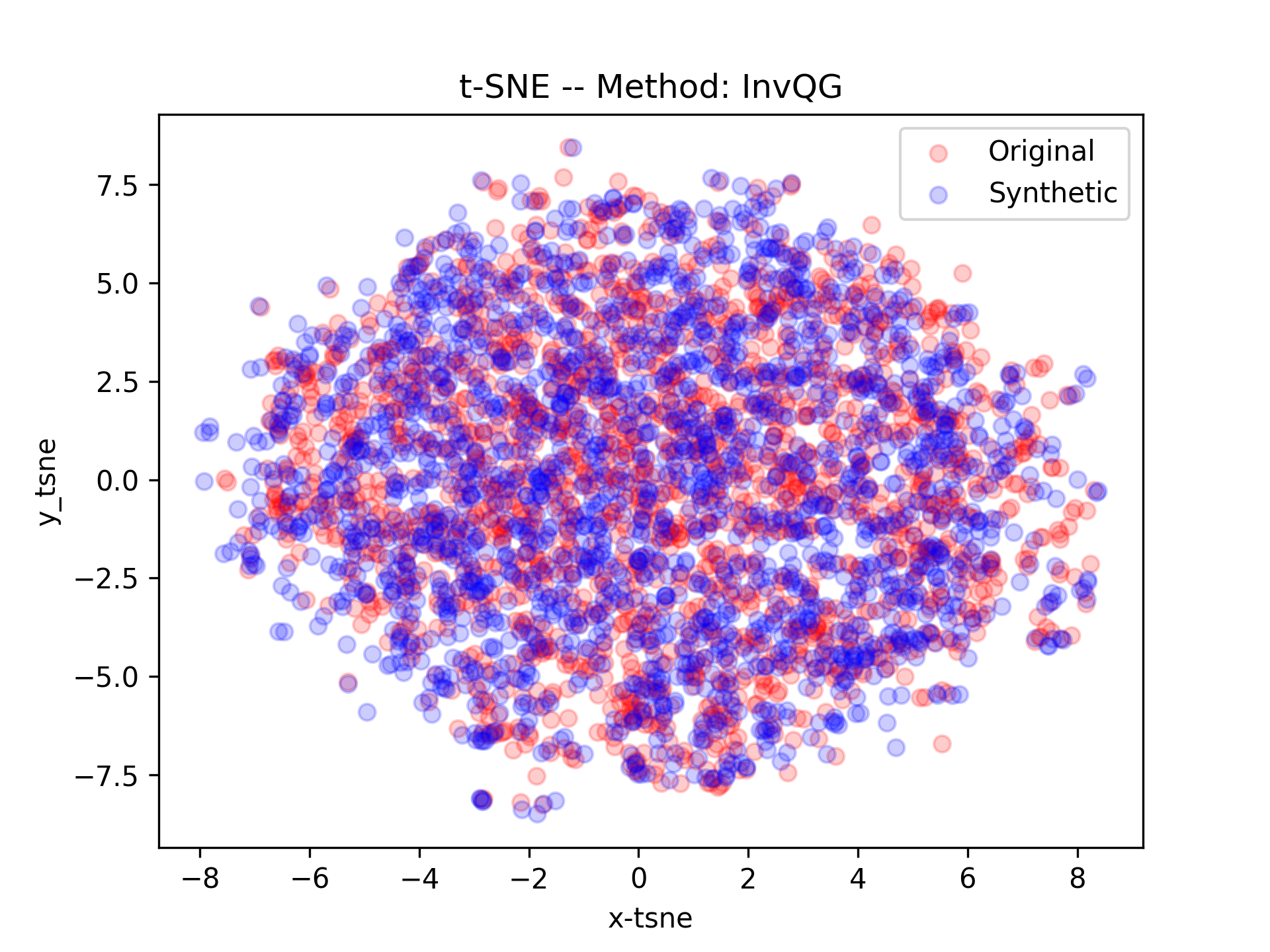}
\end{subfigure}
    \hfill
\begin{subfigure}[t]{0.18\textwidth}
   \includegraphics[width=\linewidth]{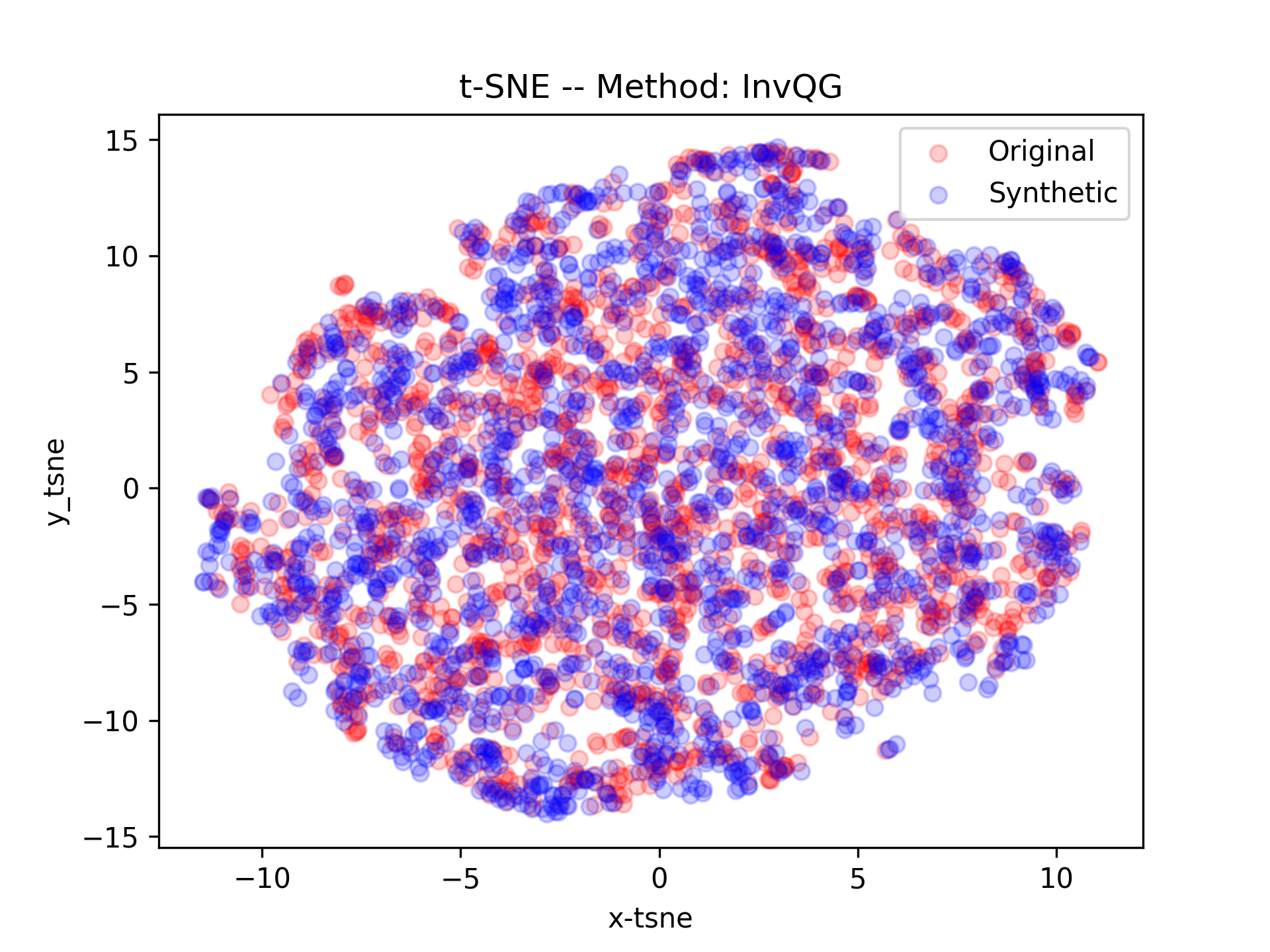}
\end{subfigure}
    \hfill
\begin{subfigure}[t]{0.18\textwidth}
  \includegraphics[width=\linewidth]{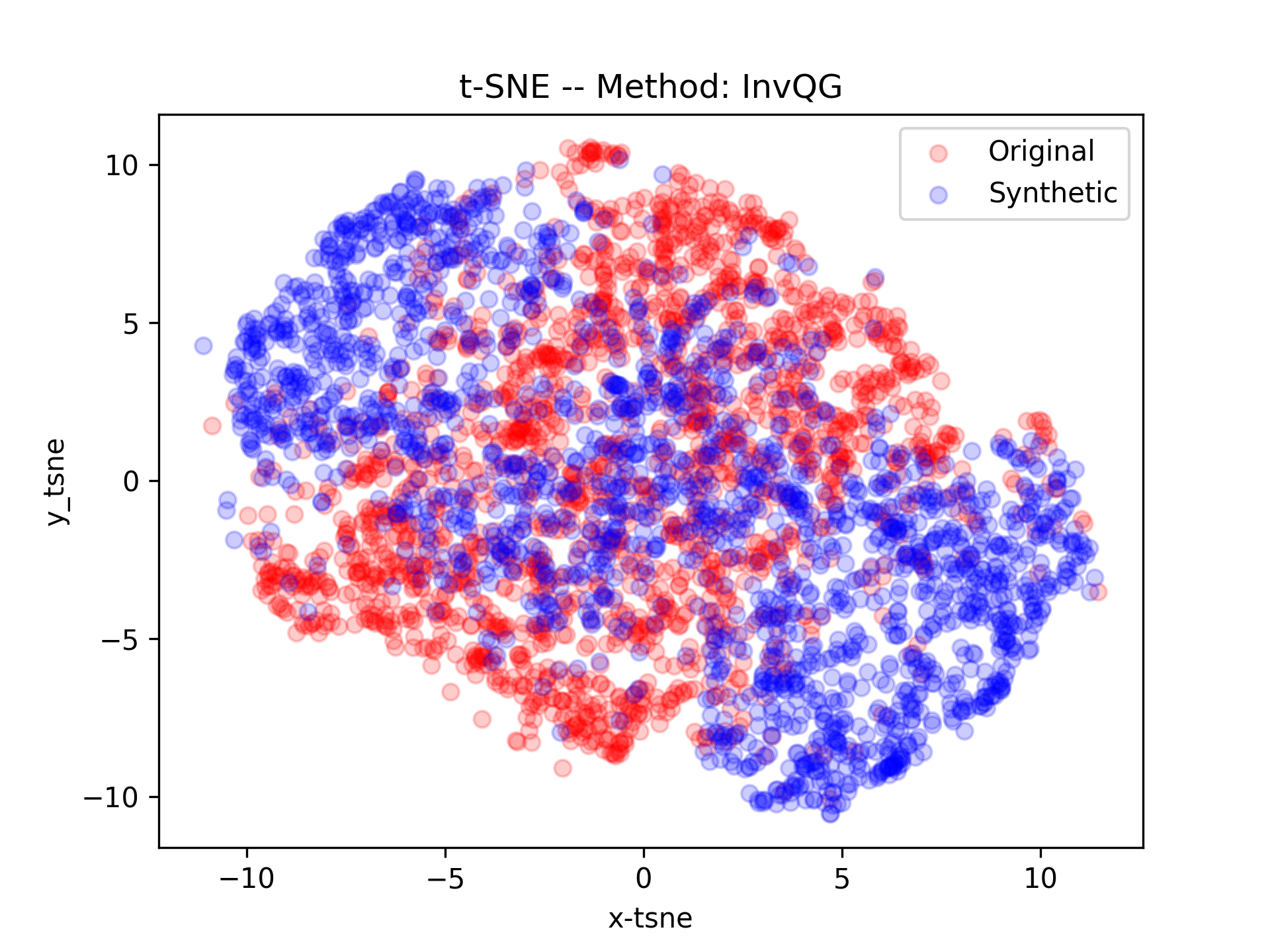}
\end{subfigure}
\hspace*{\fill}

% Second row
\hspace*{\fill}
\begin{subfigure}[t]{0.18\textwidth}
  \includegraphics[width=\linewidth]{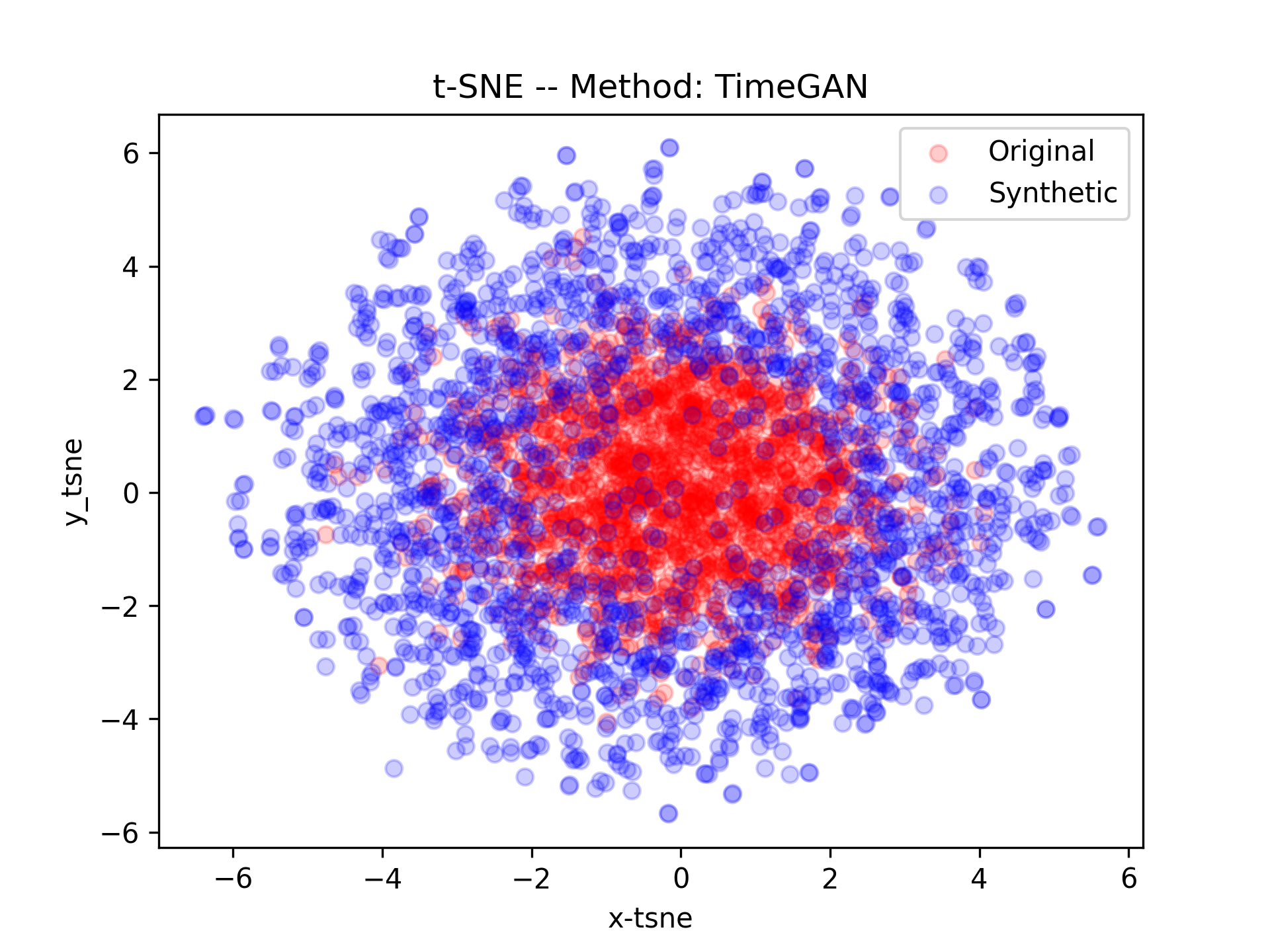}
\end{subfigure}
    \hfill
\begin{subfigure}[t]{0.18\textwidth}
  \includegraphics[width=\linewidth]{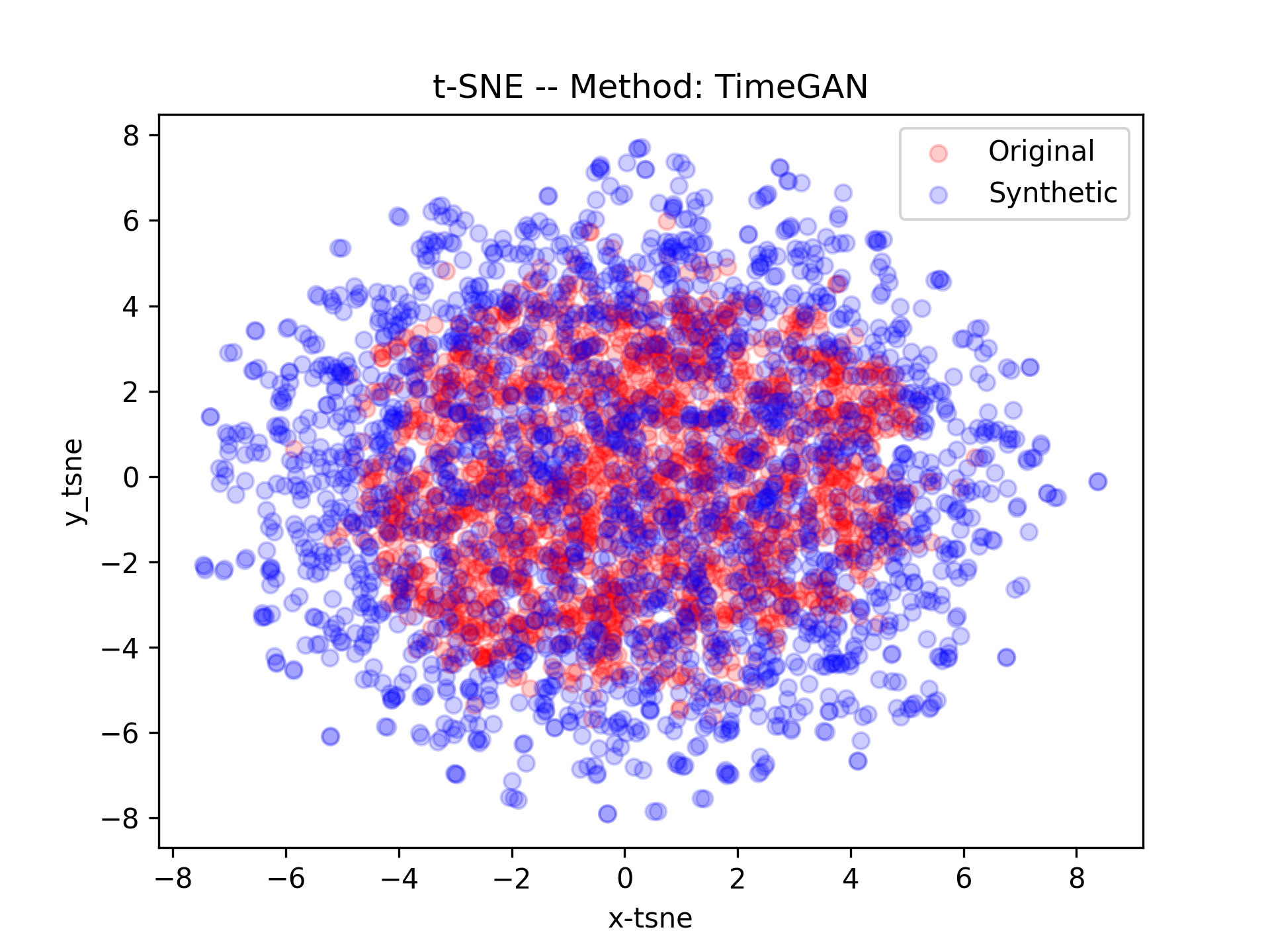}
\end{subfigure}
    \hfill
\begin{subfigure}[t]{0.18\textwidth}
  \includegraphics[width=\linewidth]{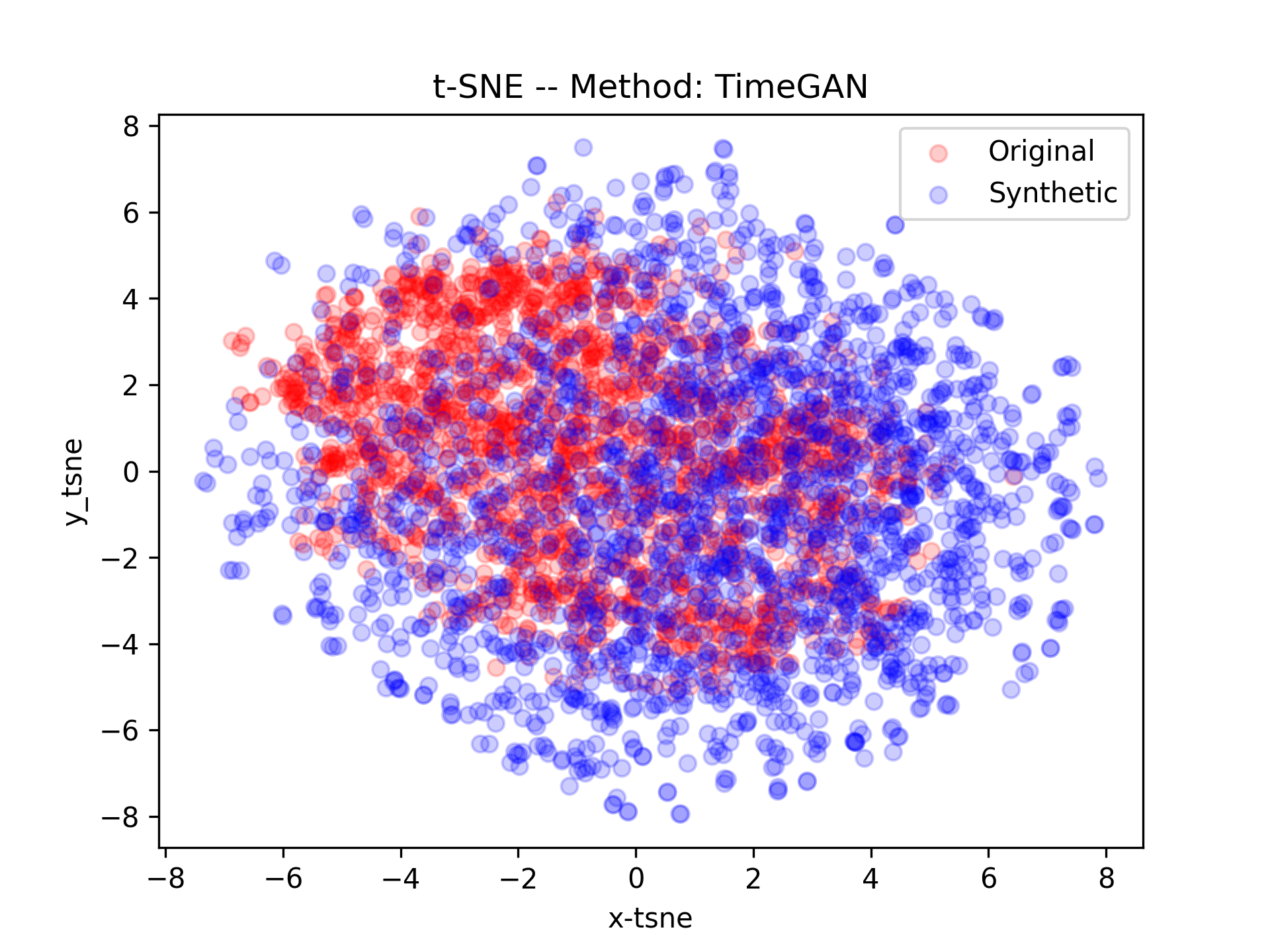}
\end{subfigure}
    \hfill
\begin{subfigure}[t]{0.18\textwidth}
  \includegraphics[width=\linewidth]{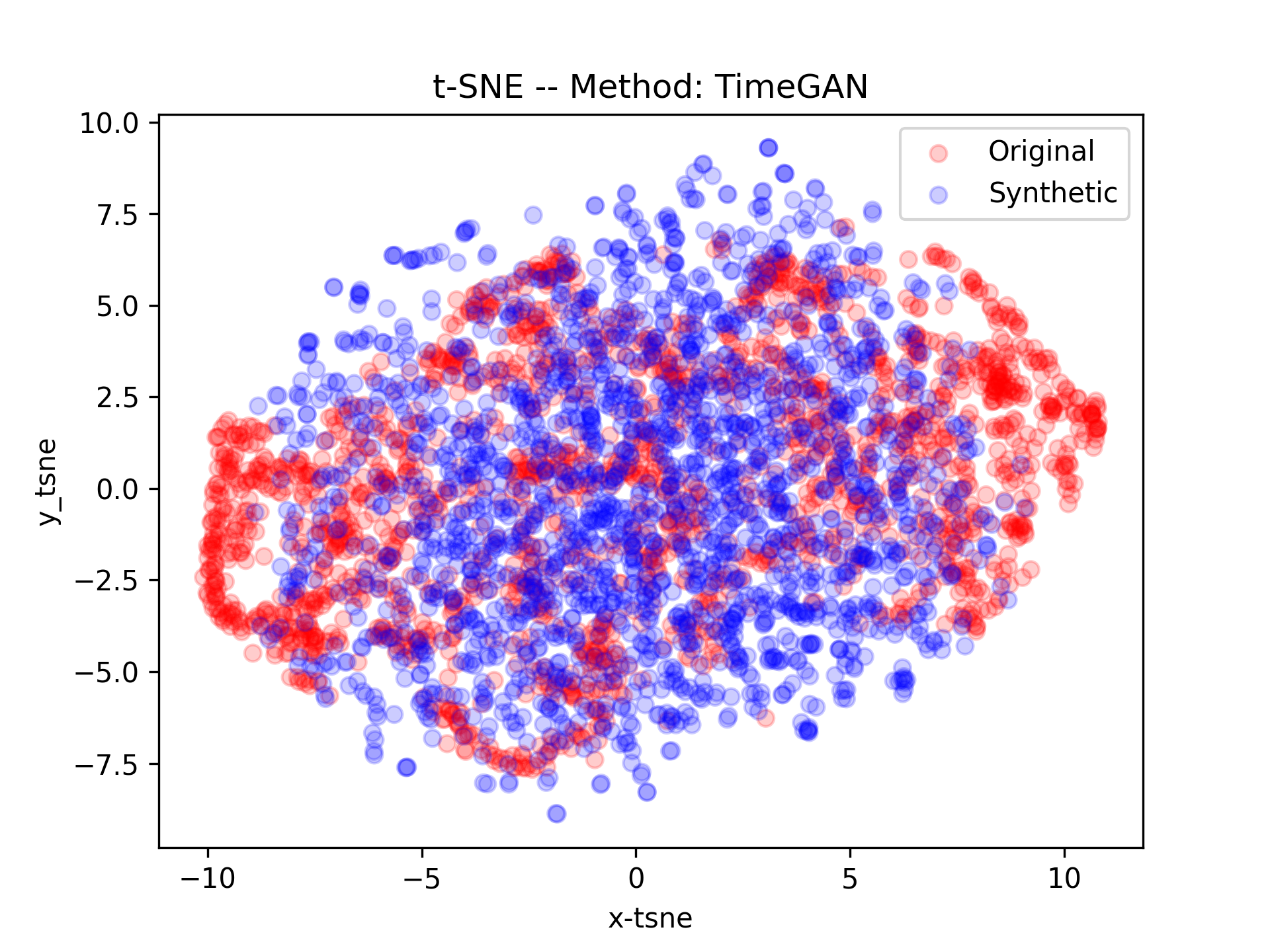}
\end{subfigure}
    \hfill
\begin{subfigure}[t]{0.18\textwidth}
   \includegraphics[width=\linewidth]{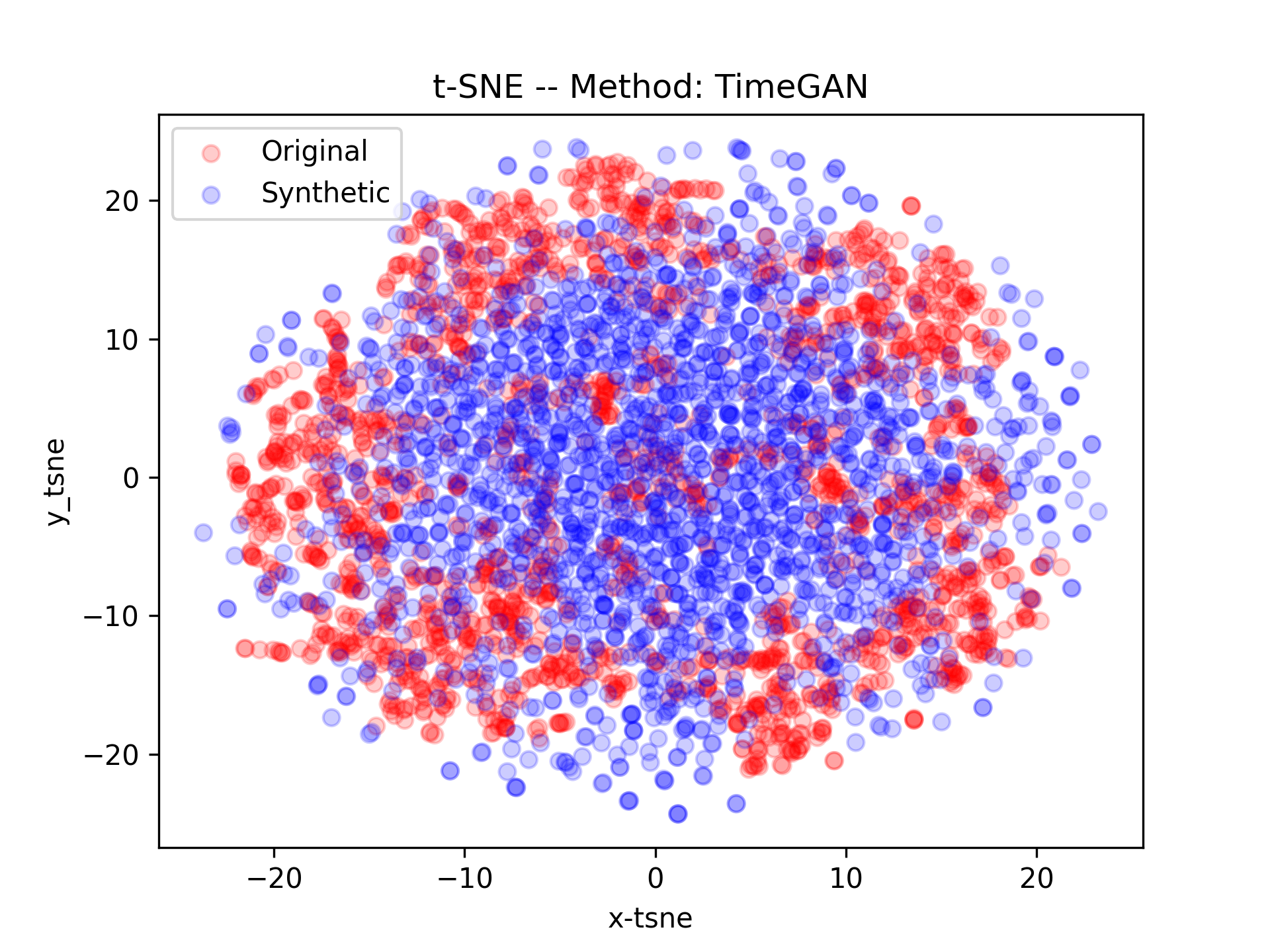}
\end{subfigure}
\hspace*{\fill}%

% 3 row
\hspace*{\fill}
\begin{subfigure}[t]{0.18\textwidth}
  \includegraphics[width=\linewidth]{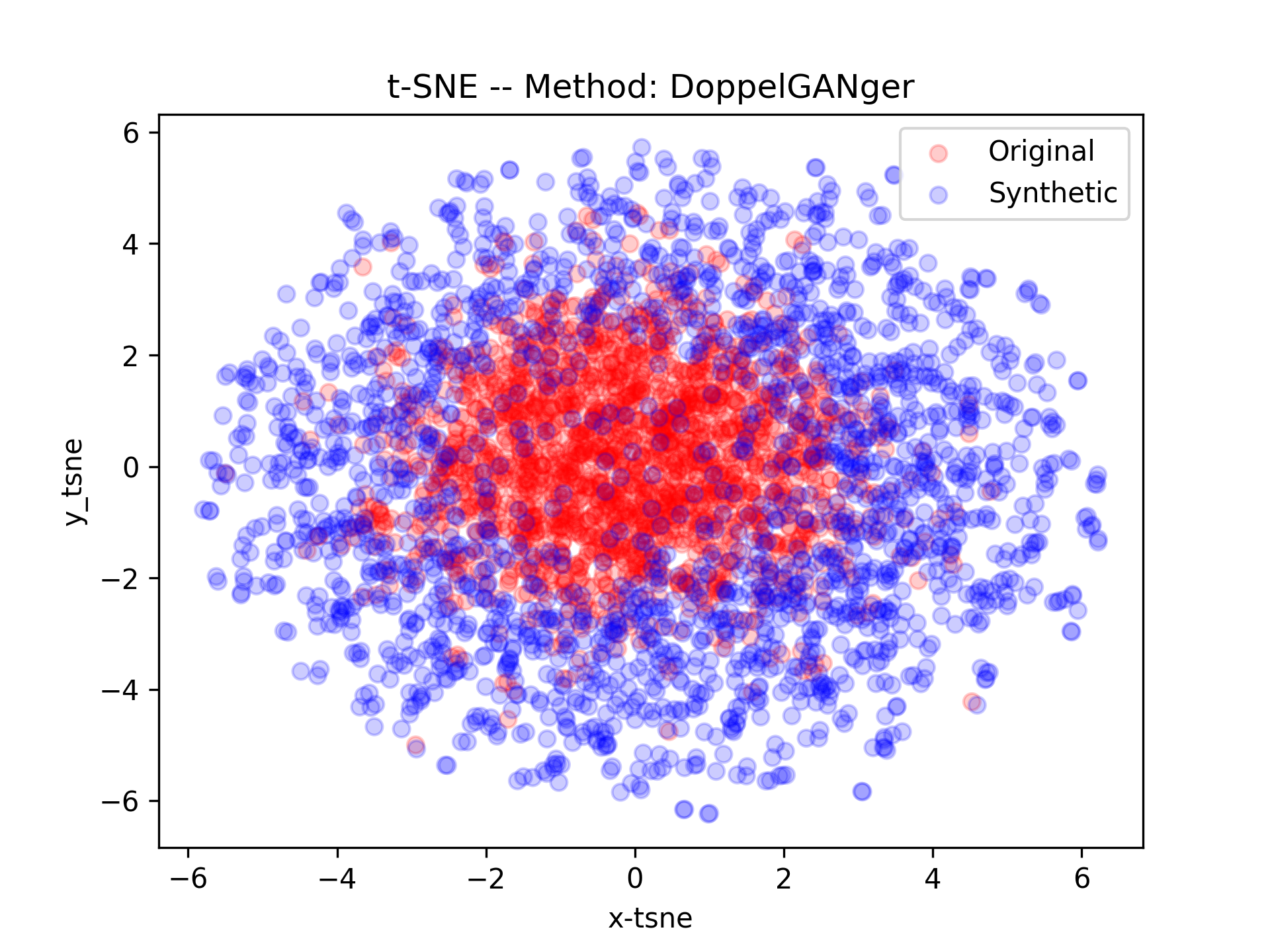}
  \caption*{\scriptsize \textbf{WN}}
\end{subfigure}
    \hfill
\begin{subfigure}[t]{0.18\textwidth}
  \includegraphics[width=\linewidth]{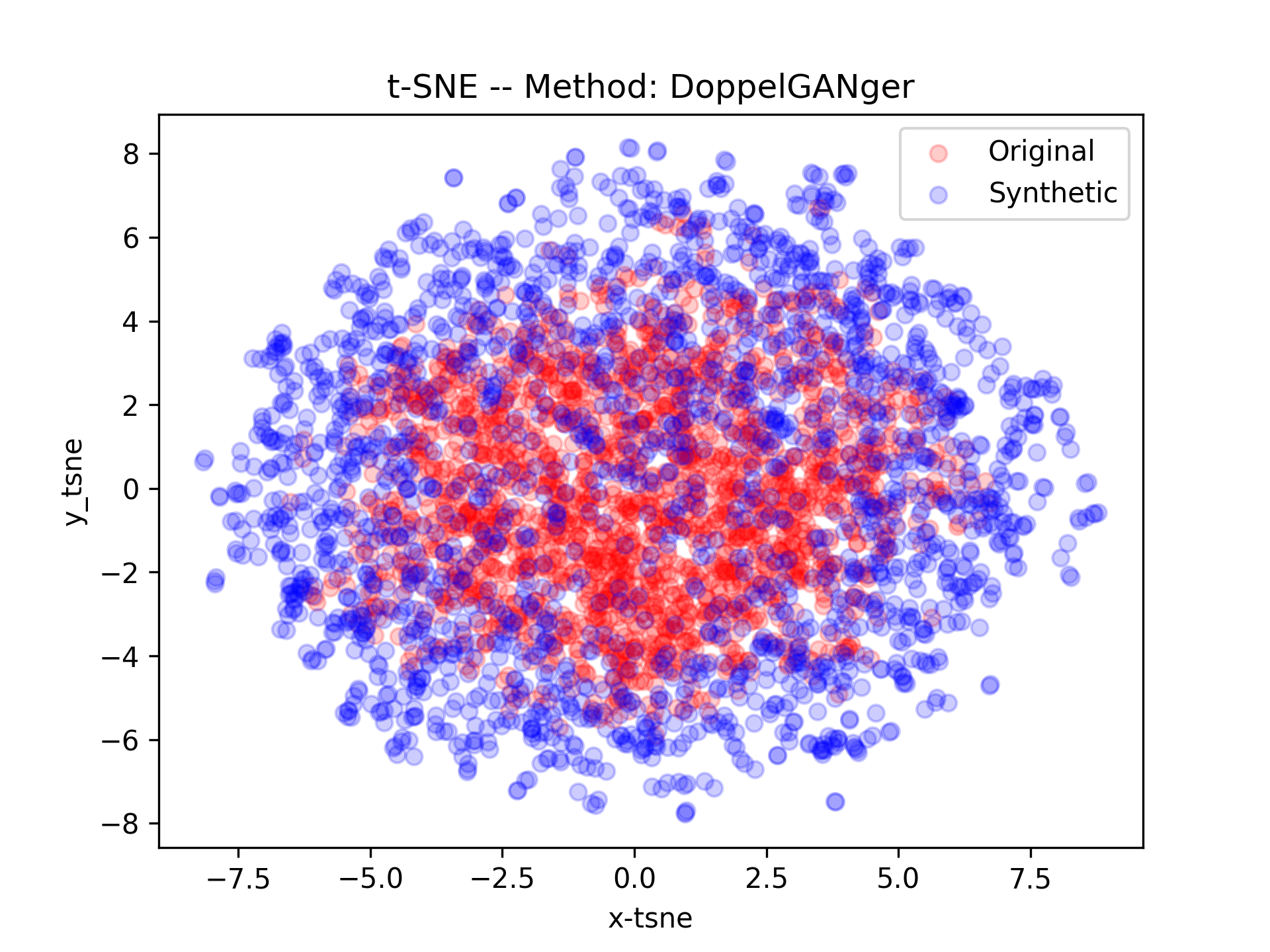}
  \caption*{\scriptsize \textbf{AR -0.5}}
\end{subfigure}
    \hfill
\begin{subfigure}[t]{0.18\textwidth}
  \includegraphics[width=\linewidth]{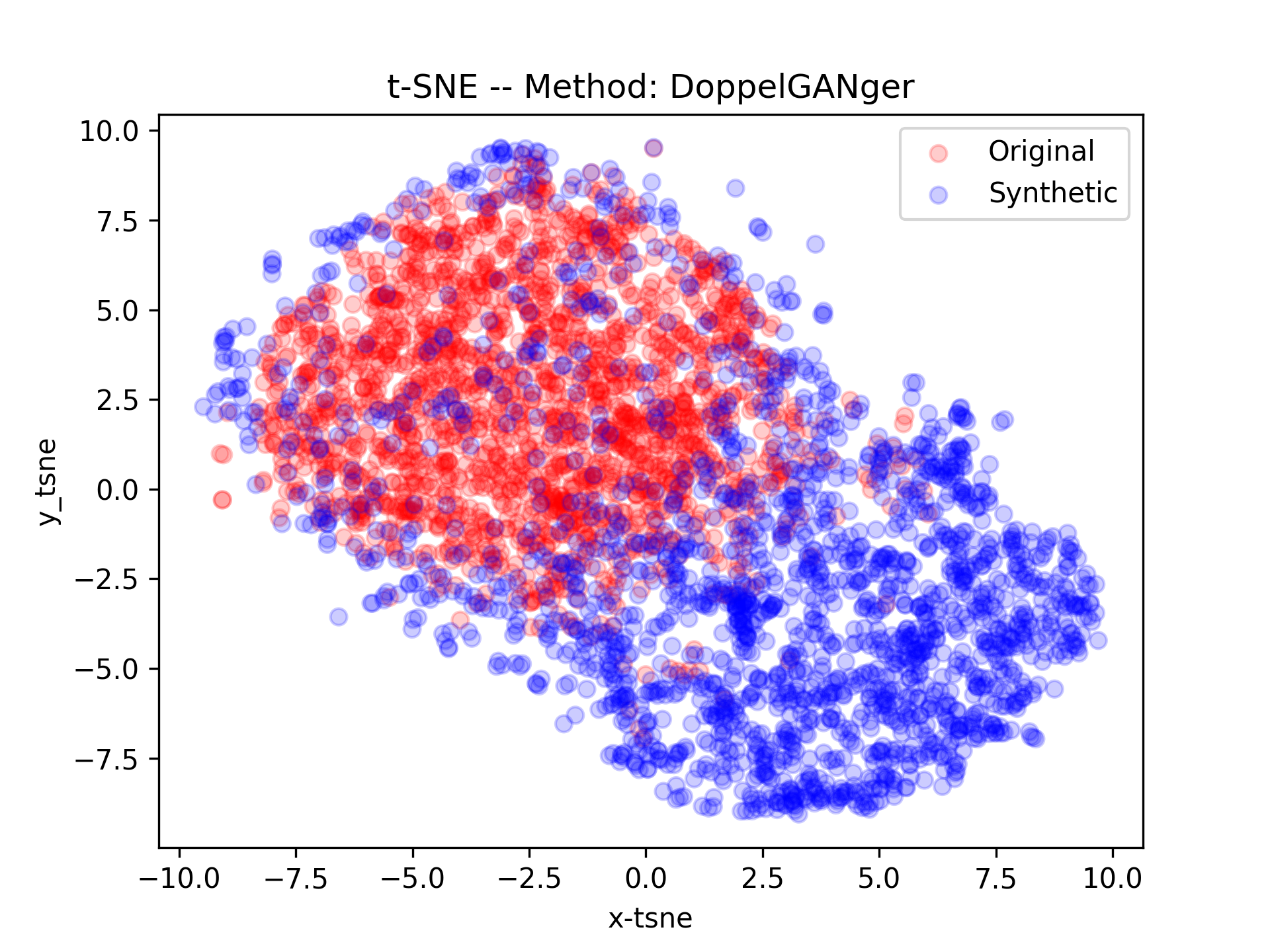}
  \caption*{\scriptsize \textbf{AR 0.5}}
\end{subfigure}
    \hfill
\begin{subfigure}[t]{0.18\textwidth}
  \includegraphics[width=\linewidth]{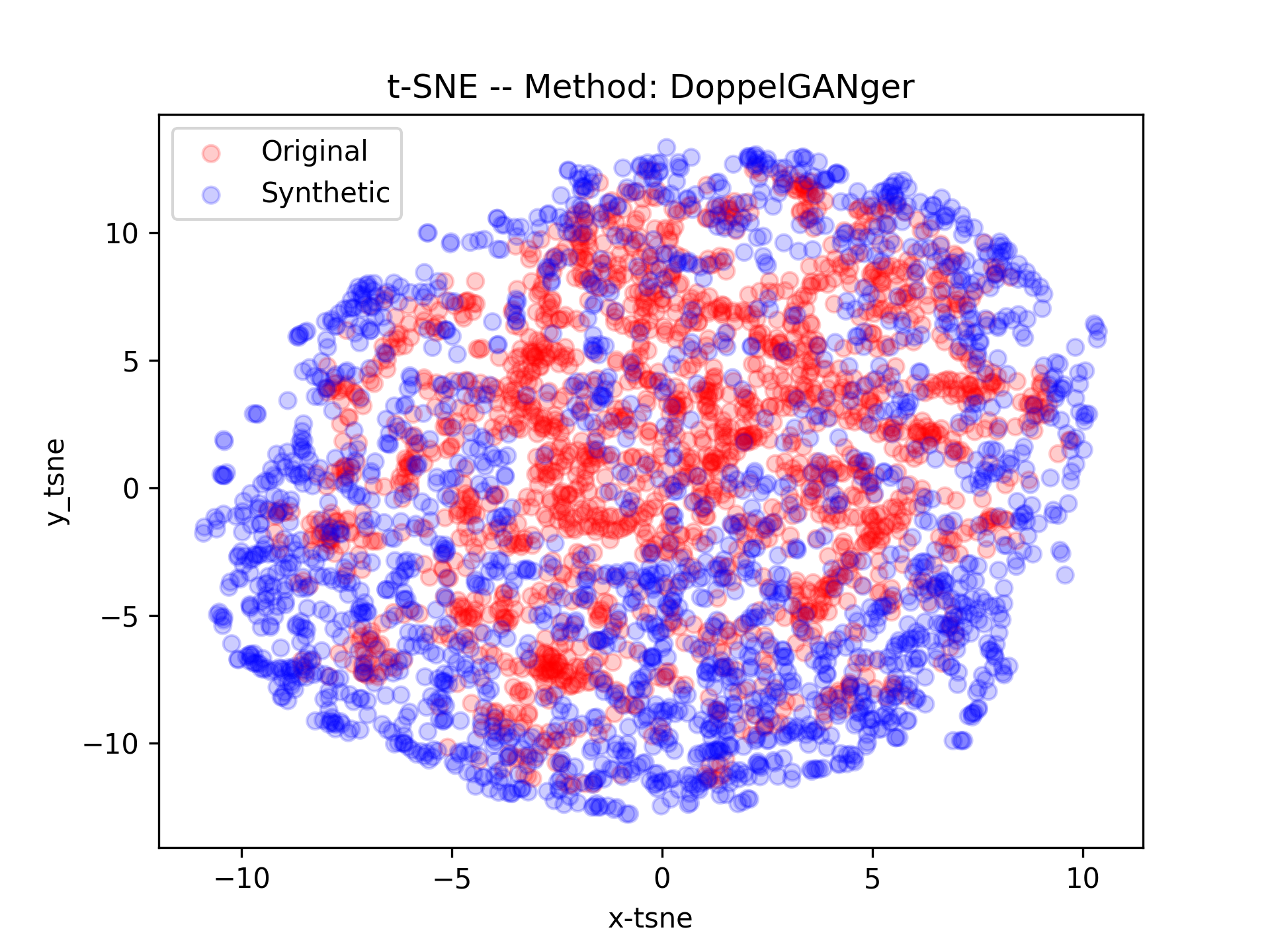}
  \caption*{\scriptsize \textbf{AR 0.9}}
\end{subfigure}
    \hfill
\begin{subfigure}[t]{0.18\textwidth}
   \includegraphics[width=\linewidth]{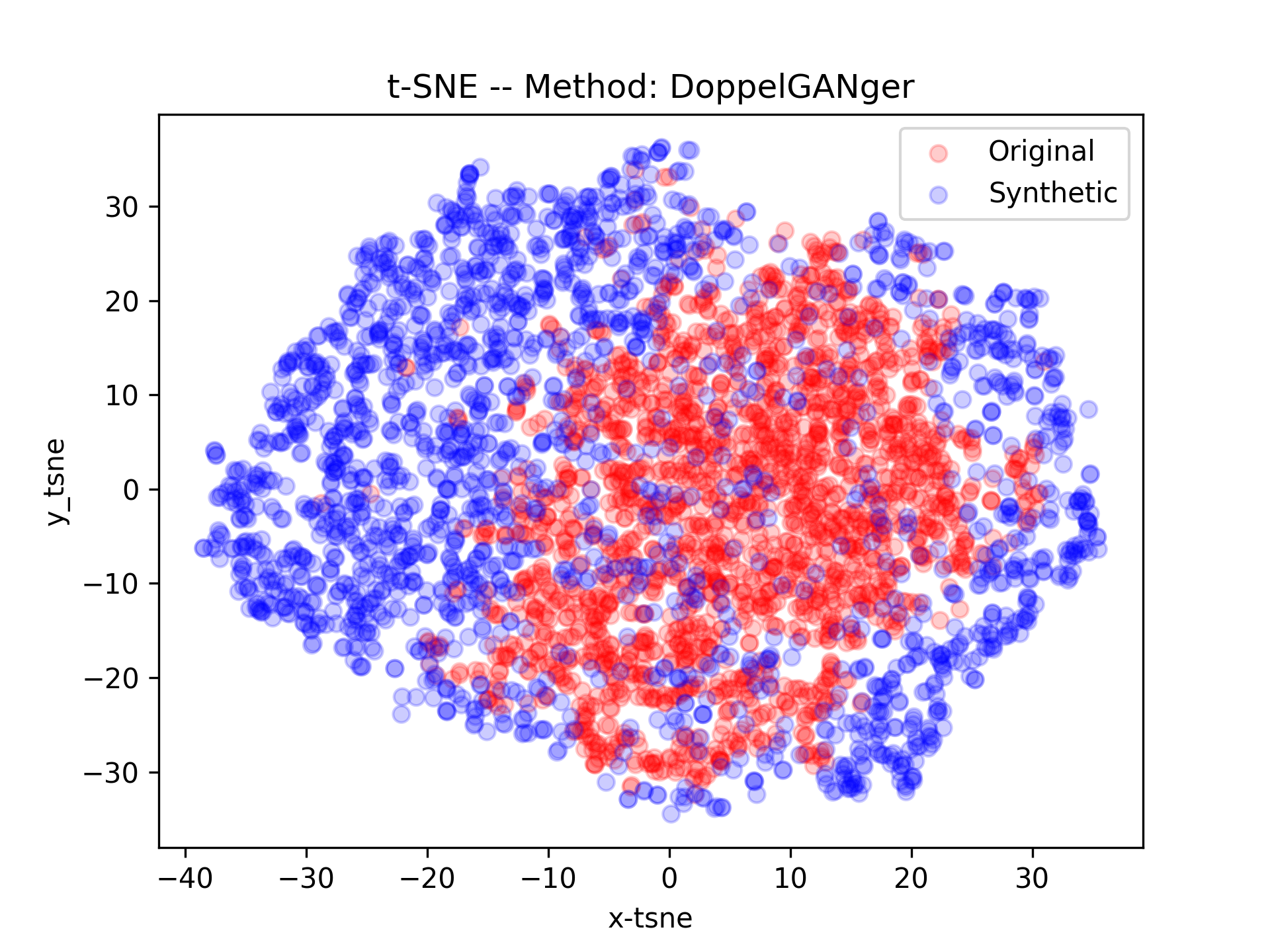}
  \caption*{\scriptsize \textbf{AR2}}
\end{subfigure}
\hspace*{\fill}%
\caption[t-SNE projections for synthetic data generated by InvQG, TimeGAN and DoppelGranger]{t-SNE projections for synthetic data generated by InvQG (top row), TimeGAN (middle row), and DoppelGANger (bottom row) for selected models (WN, AR1-0.5, AR10.5, AR10.9, AR2).}
\label{fig:t_sne_time_gan_vs_qg}
\end{figure}

\begin{figure}[!h]
\centering
\captionsetup[subfigure]{justification=centering}

% First row
\hspace*{\fill}
\begin{subfigure}[t]{0.15\textwidth}
  \includegraphics[width=\linewidth]{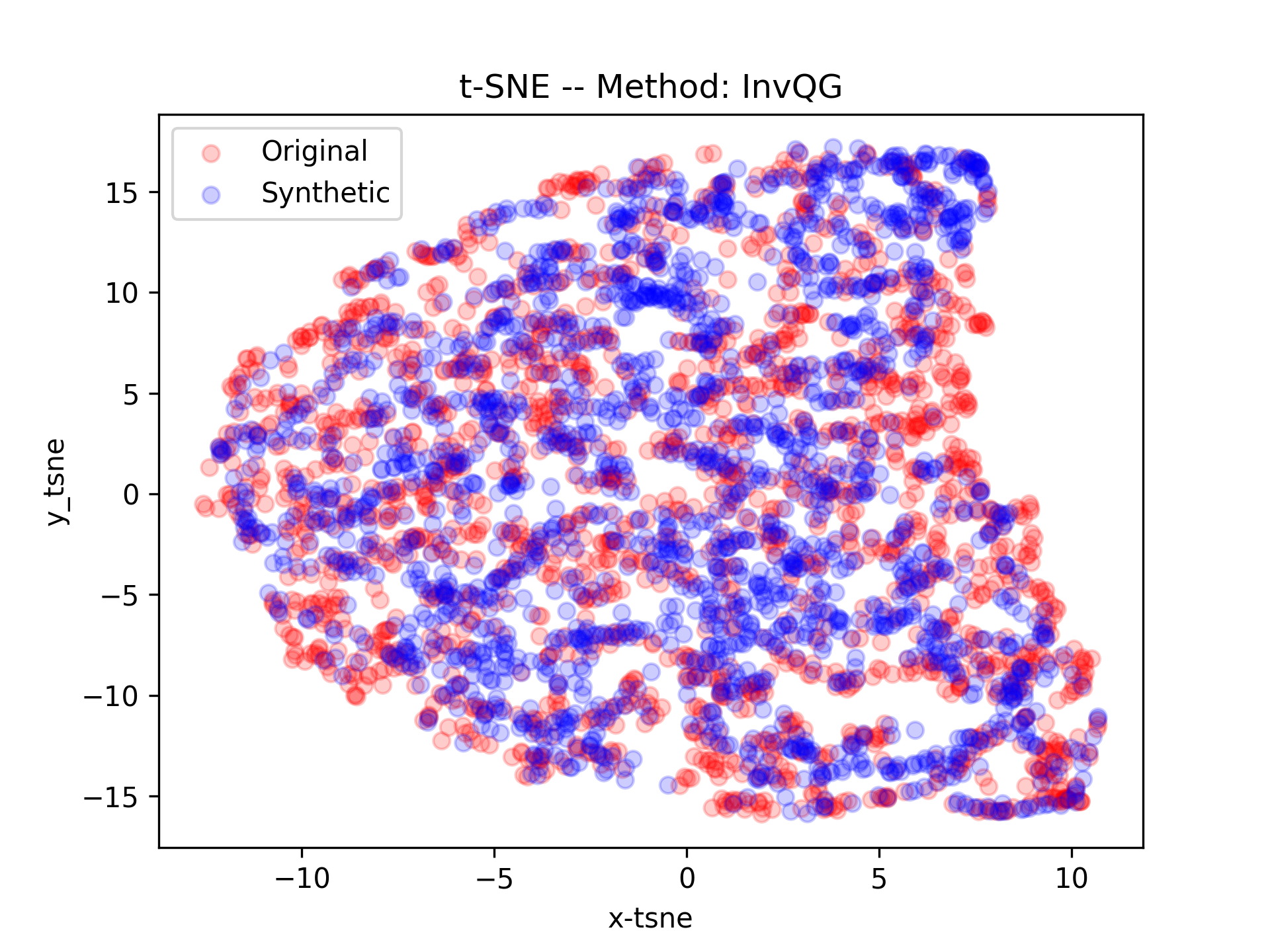}
\end{subfigure}
    \hfill
\begin{subfigure}[t]{0.15\textwidth}
  \includegraphics[width=\linewidth]{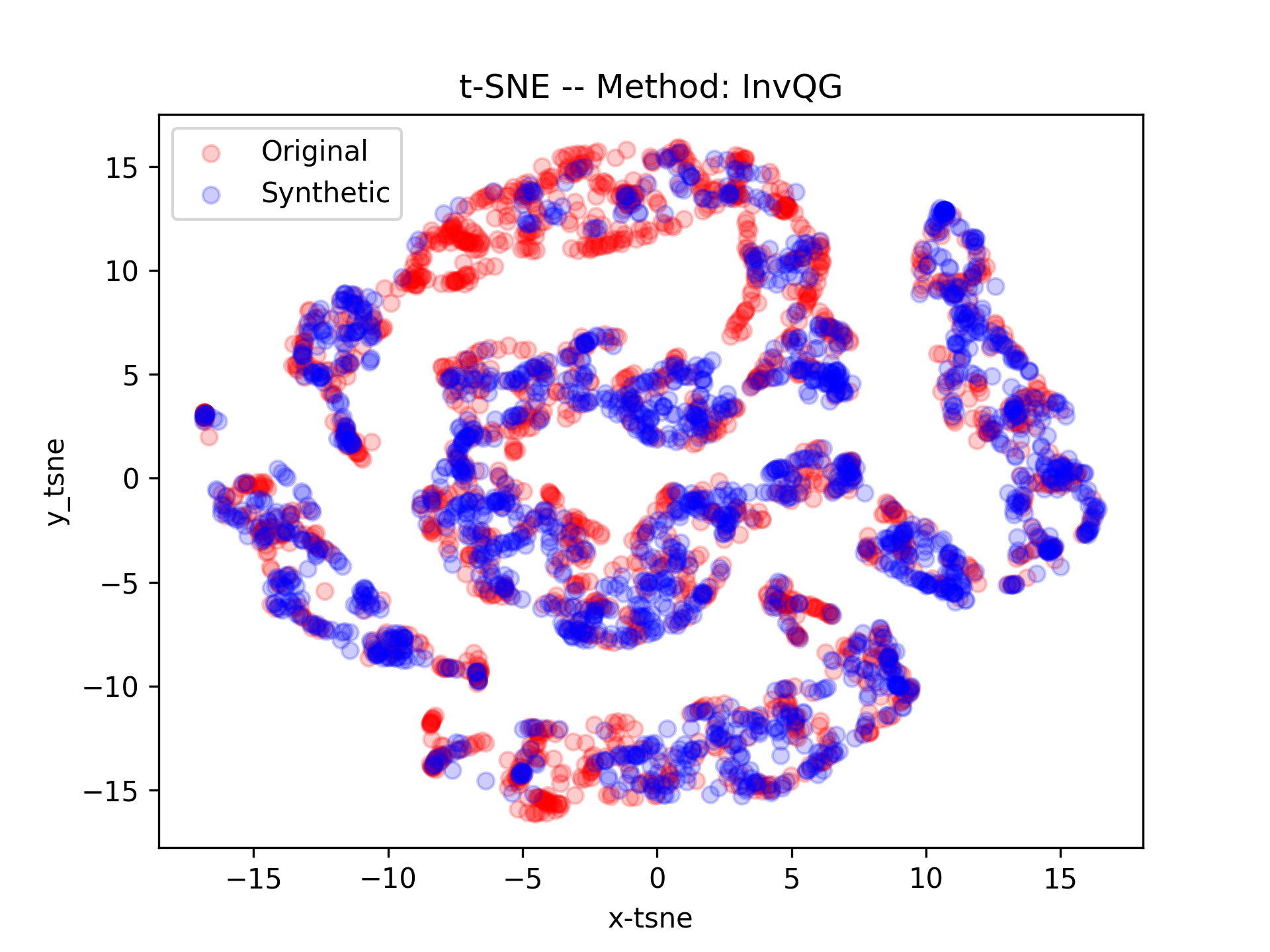}
\end{subfigure}
    \hfill
\begin{subfigure}[t]{0.15\textwidth}
  \includegraphics[width=\linewidth]{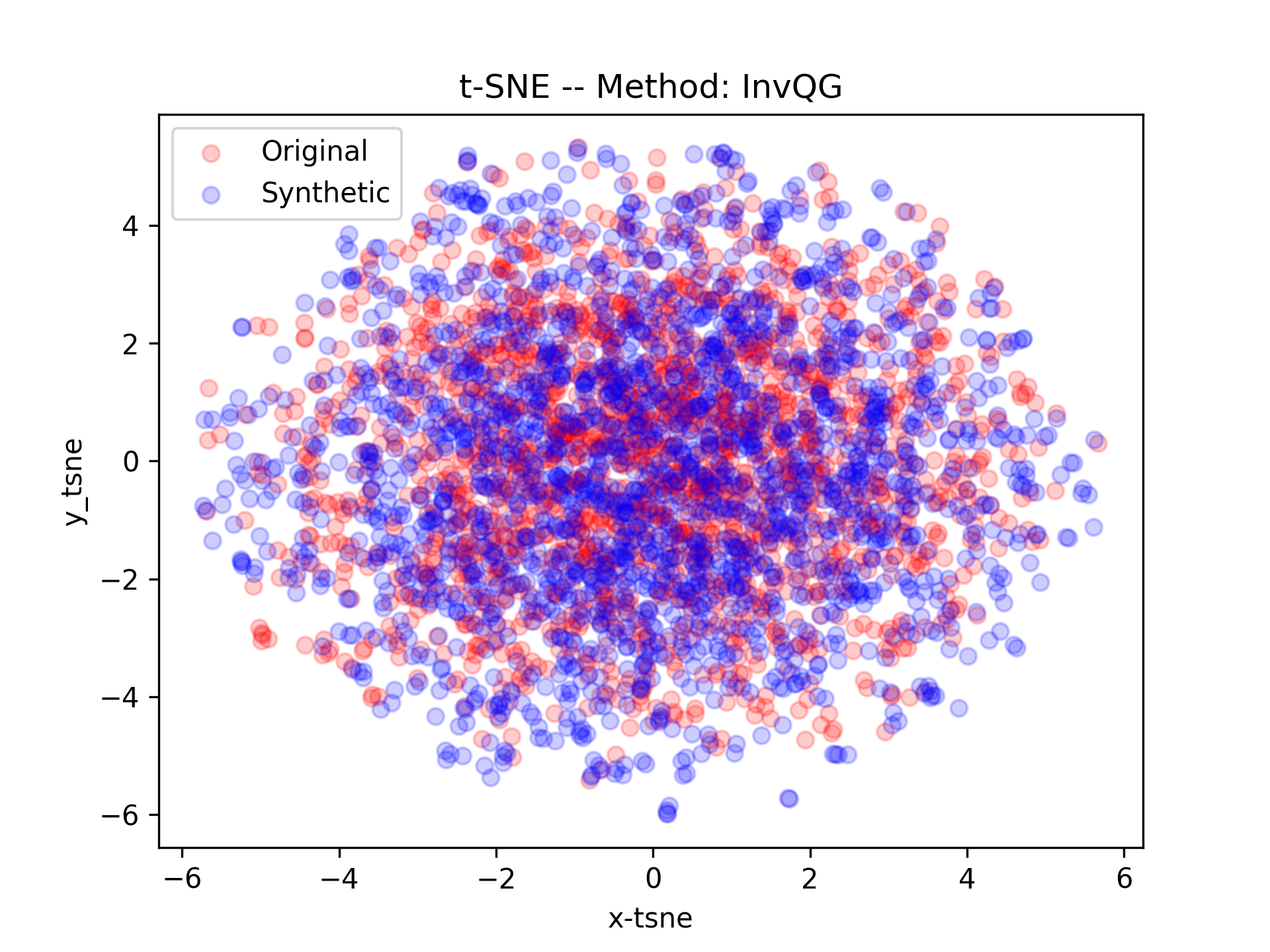}
\end{subfigure}
    \hfill
\begin{subfigure}[t]{0.15\textwidth}
  \includegraphics[width=\linewidth]{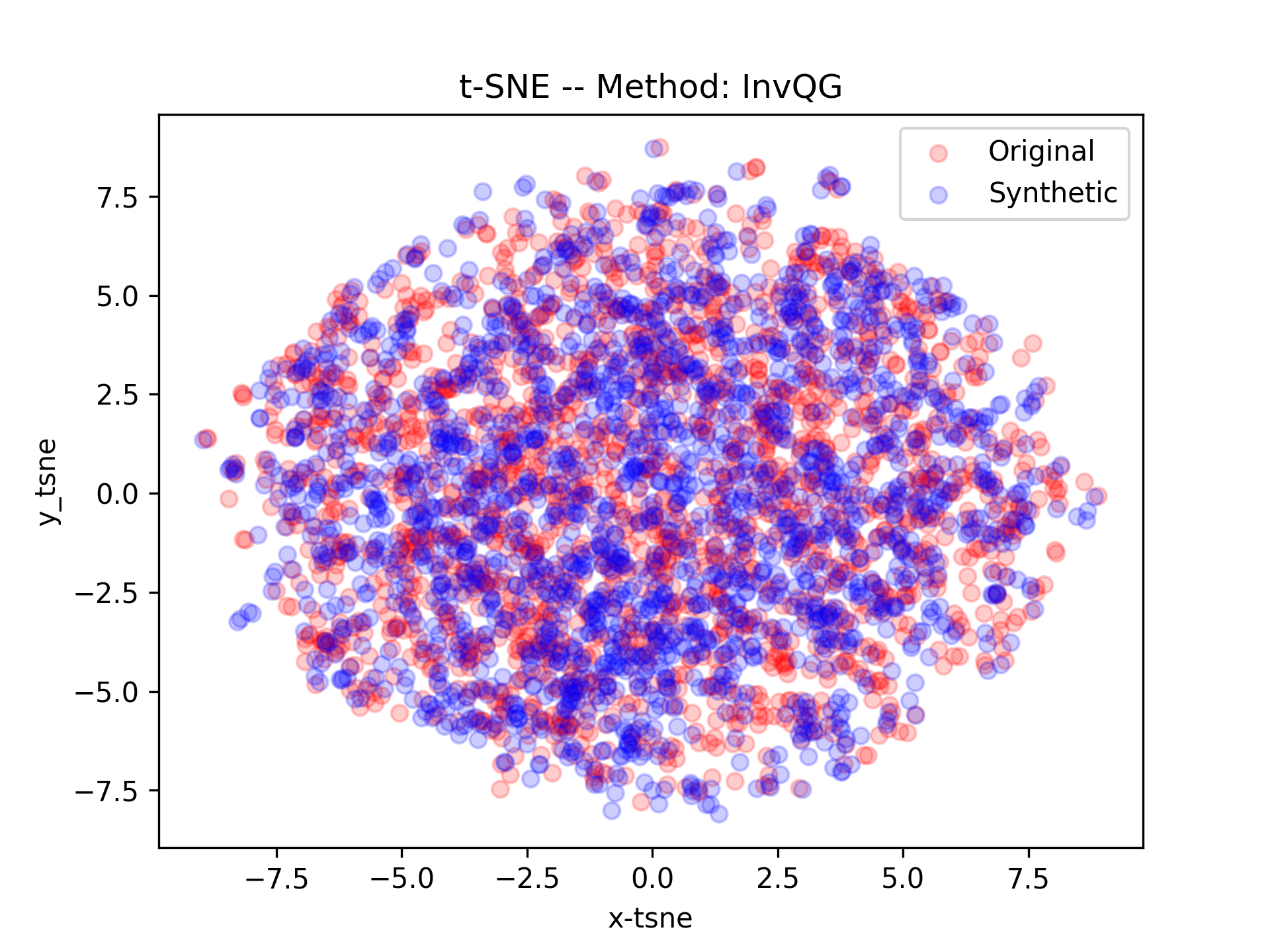}
\end{subfigure}
    \hfill
\begin{subfigure}[t]{0.15\textwidth}
  \includegraphics[width=\linewidth]{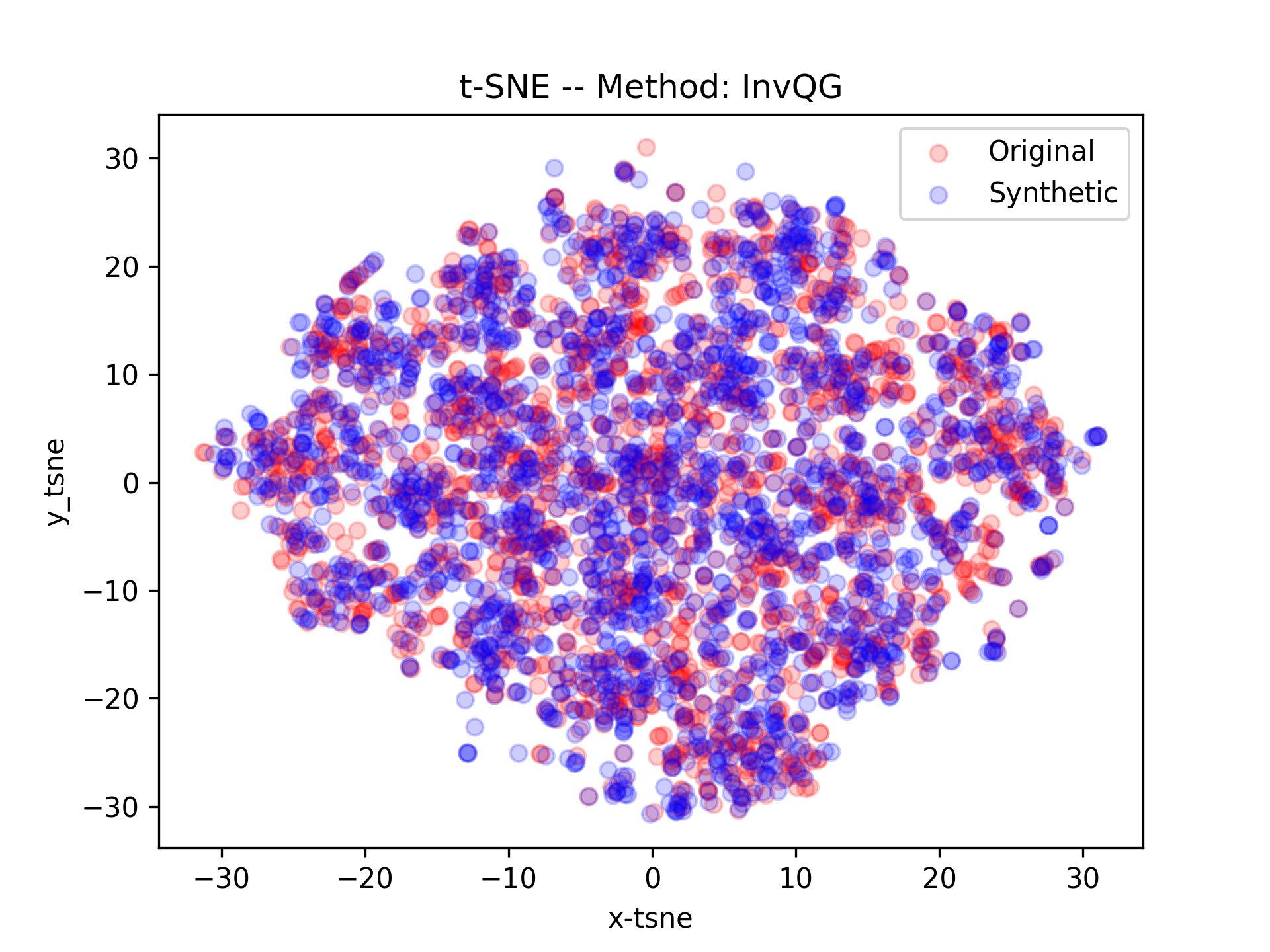}
\end{subfigure}
    \hfill
\begin{subfigure}[t]{0.15\textwidth}
  \includegraphics[width=\linewidth]{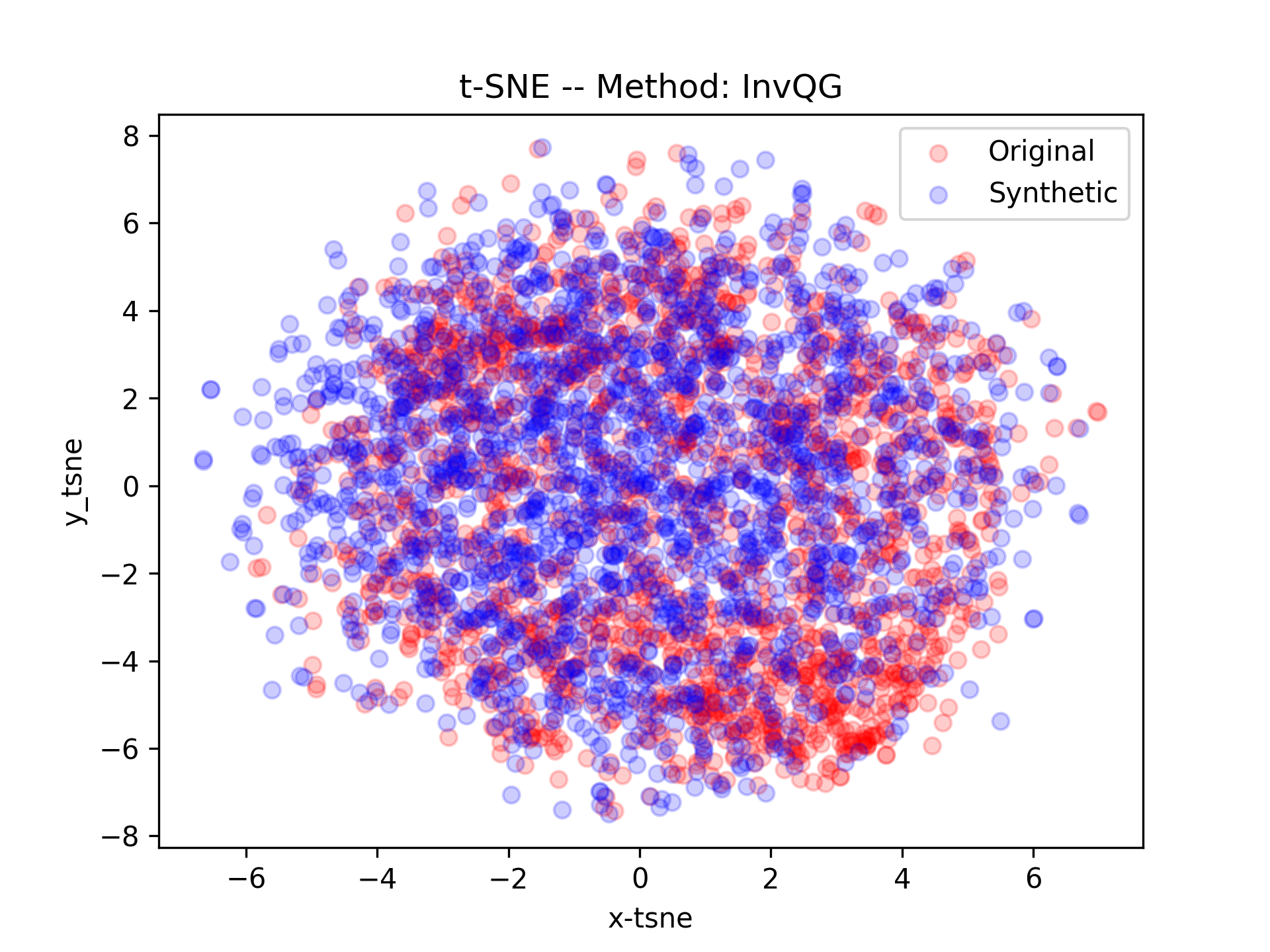}
\end{subfigure}
\hspace*{\fill}

% Second row
\hspace*{\fill}
\begin{subfigure}[t]{0.15\textwidth}
  \includegraphics[width=\linewidth]{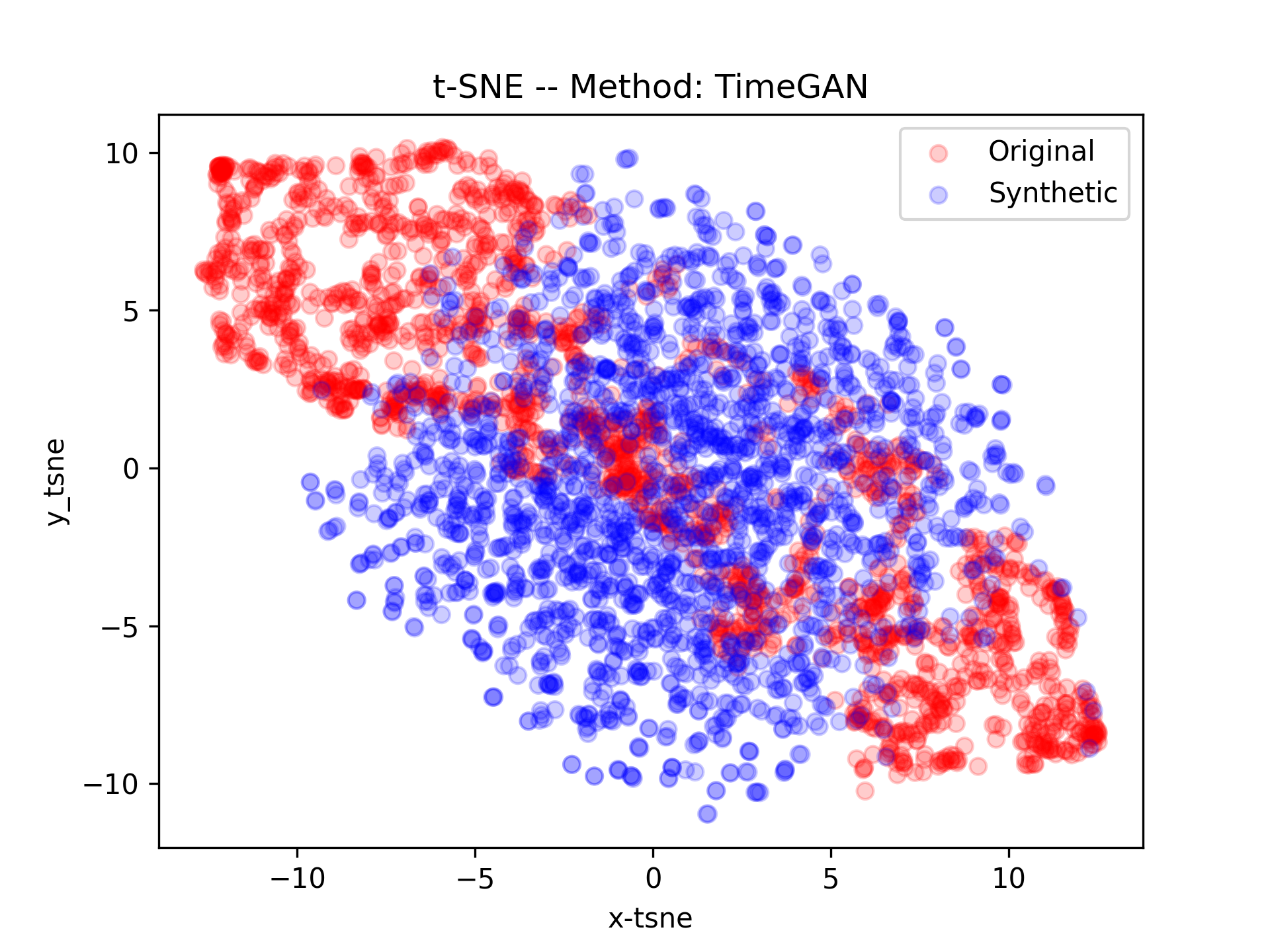}
\end{subfigure}
    \hfill
\begin{subfigure}[t]{0.15\textwidth}
  \includegraphics[width=\linewidth]{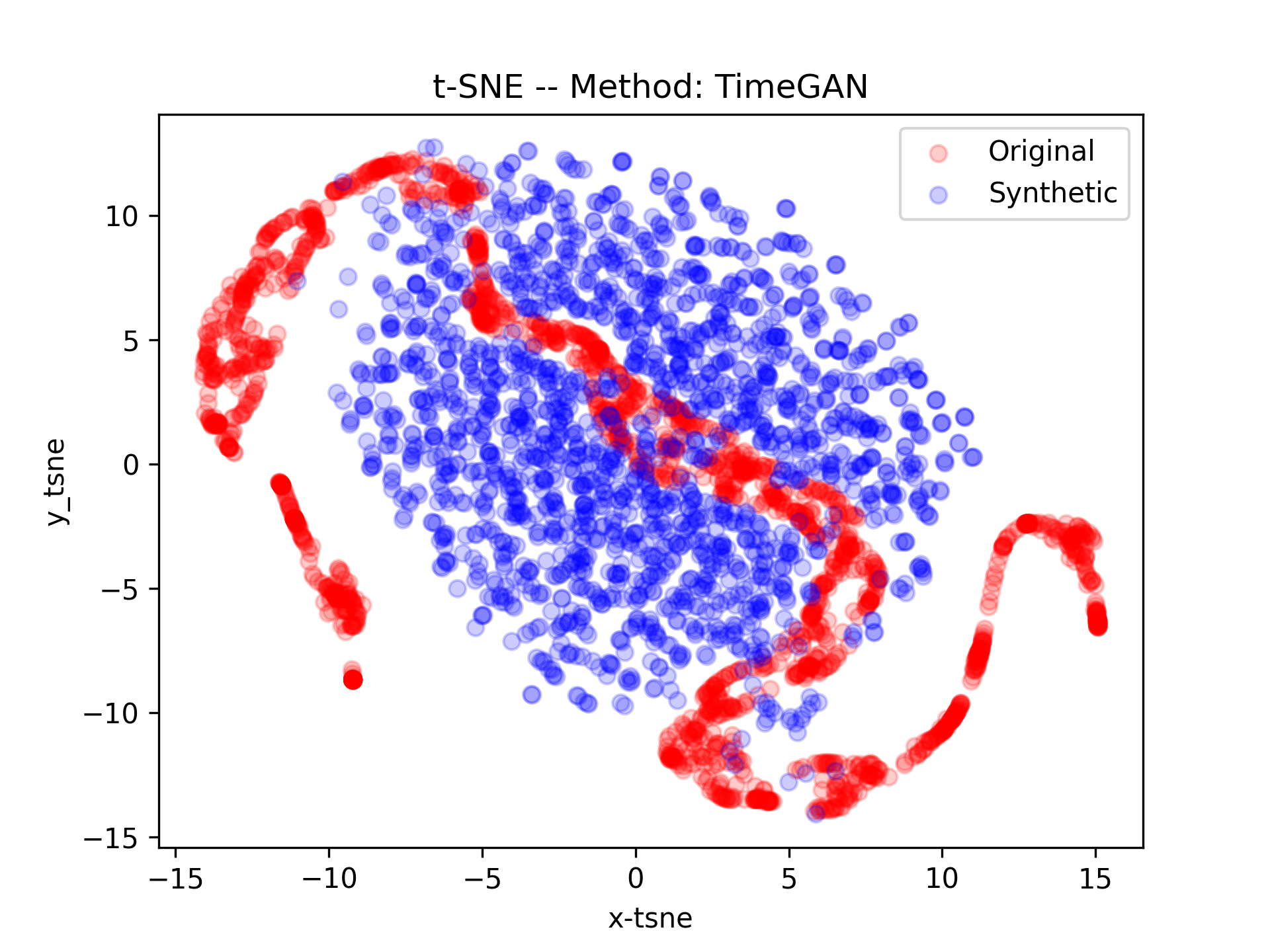}
\end{subfigure}
    \hfill
\begin{subfigure}[t]{0.15\textwidth}
  \includegraphics[width=\linewidth]{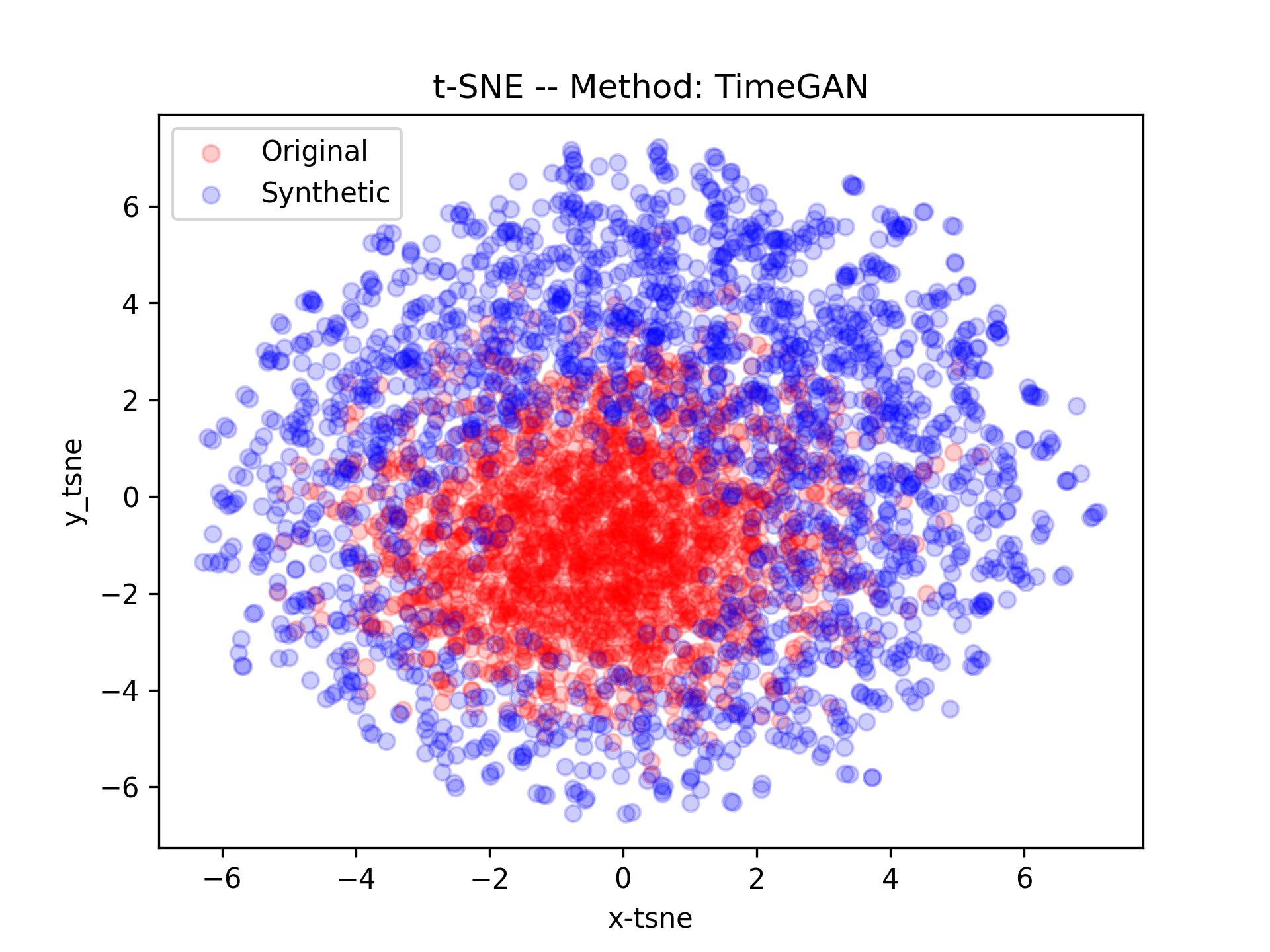}
\end{subfigure}
    \hfill
\begin{subfigure}[t]{0.15\textwidth}
  \includegraphics[width=\linewidth]{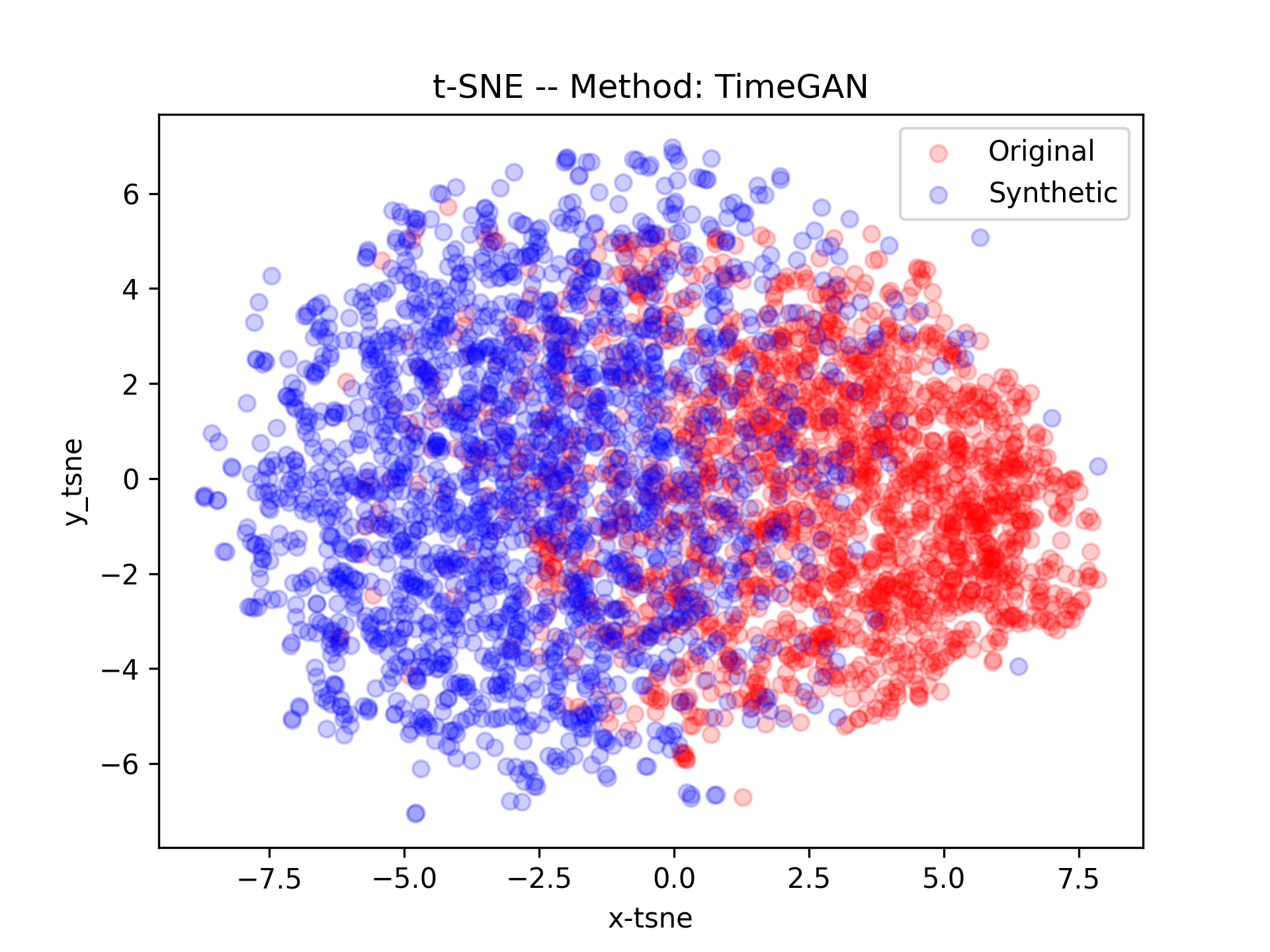}
\end{subfigure}
    \hfill
\begin{subfigure}[t]{0.15\textwidth}
  \includegraphics[width=\linewidth]{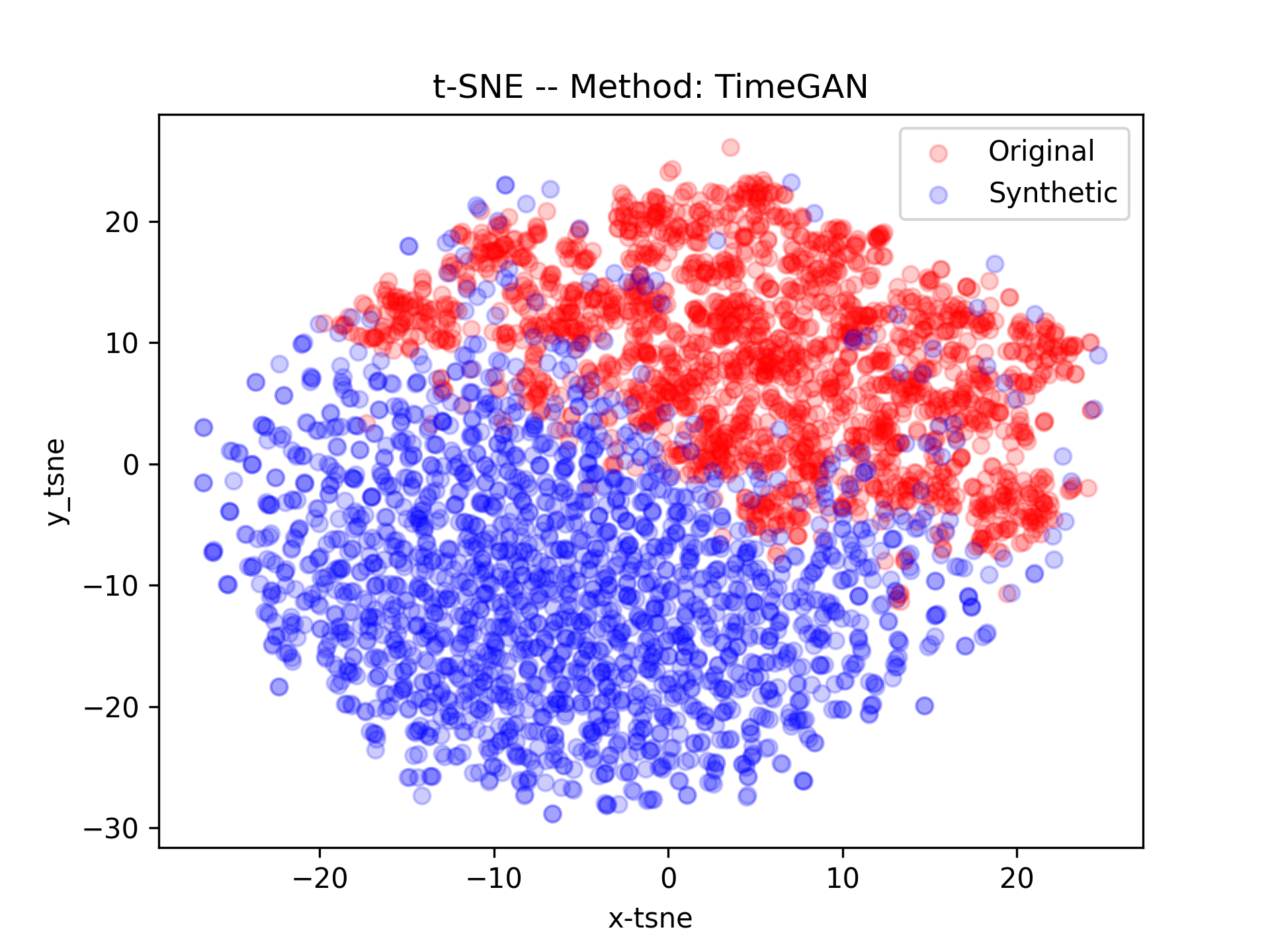}
\end{subfigure}
    \hfill
\begin{subfigure}[t]{0.15\textwidth}
  \includegraphics[width=\linewidth]{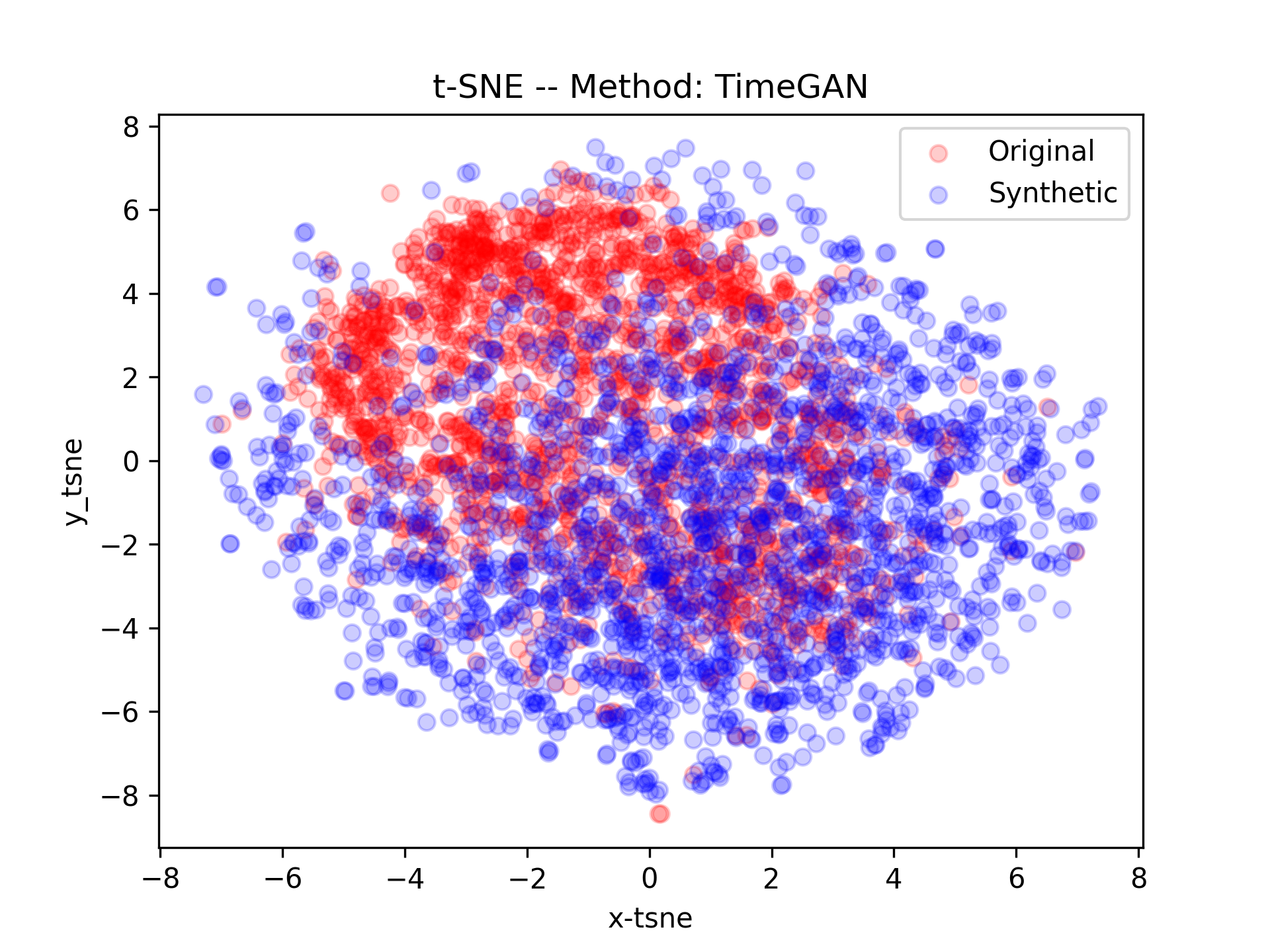}
\end{subfigure}
\hspace*{\fill}

% 3 row
\hspace*{\fill}
\begin{subfigure}[t]{0.15\textwidth}
  \includegraphics[width=\linewidth]{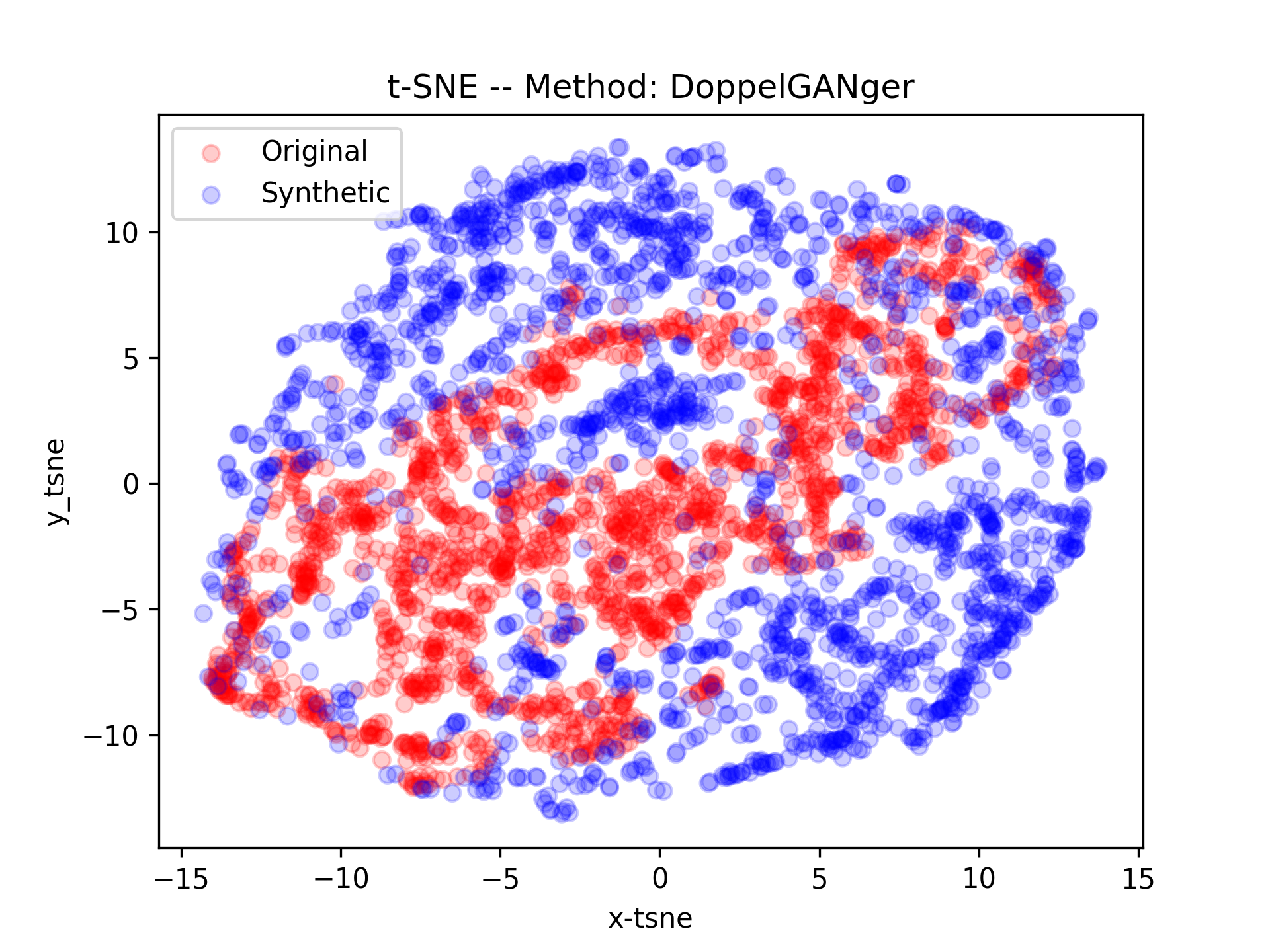}
  \caption*{\scriptsize \textbf{ARFIMA}}
\end{subfigure}
    \hfill
\begin{subfigure}[t]{0.15\textwidth}
  \includegraphics[width=\linewidth]{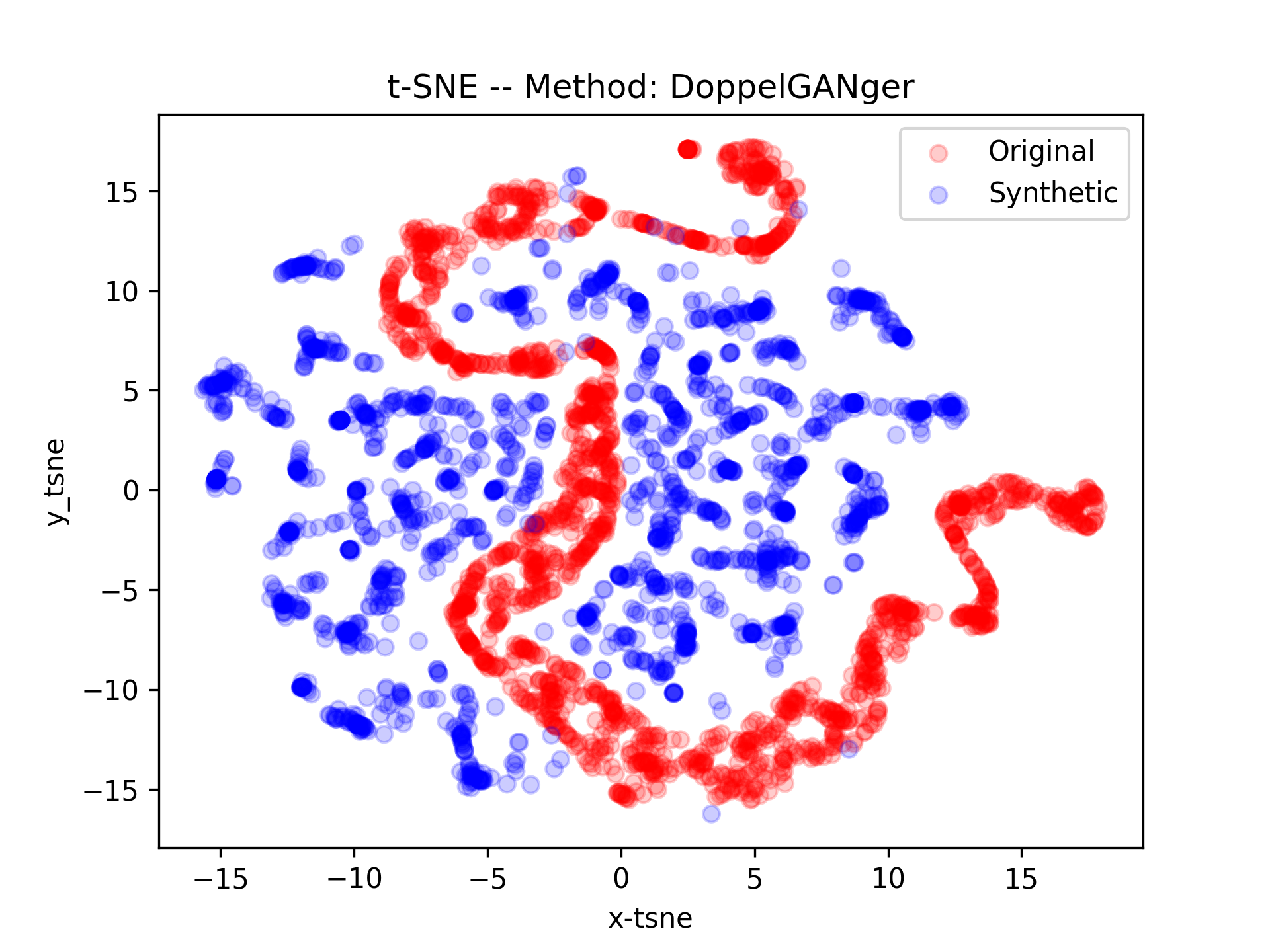}
  \caption*{\scriptsize \textbf{ARIMA}}
\end{subfigure}
    \hfill
\begin{subfigure}[t]{0.15\textwidth}
  \includegraphics[width=\linewidth]{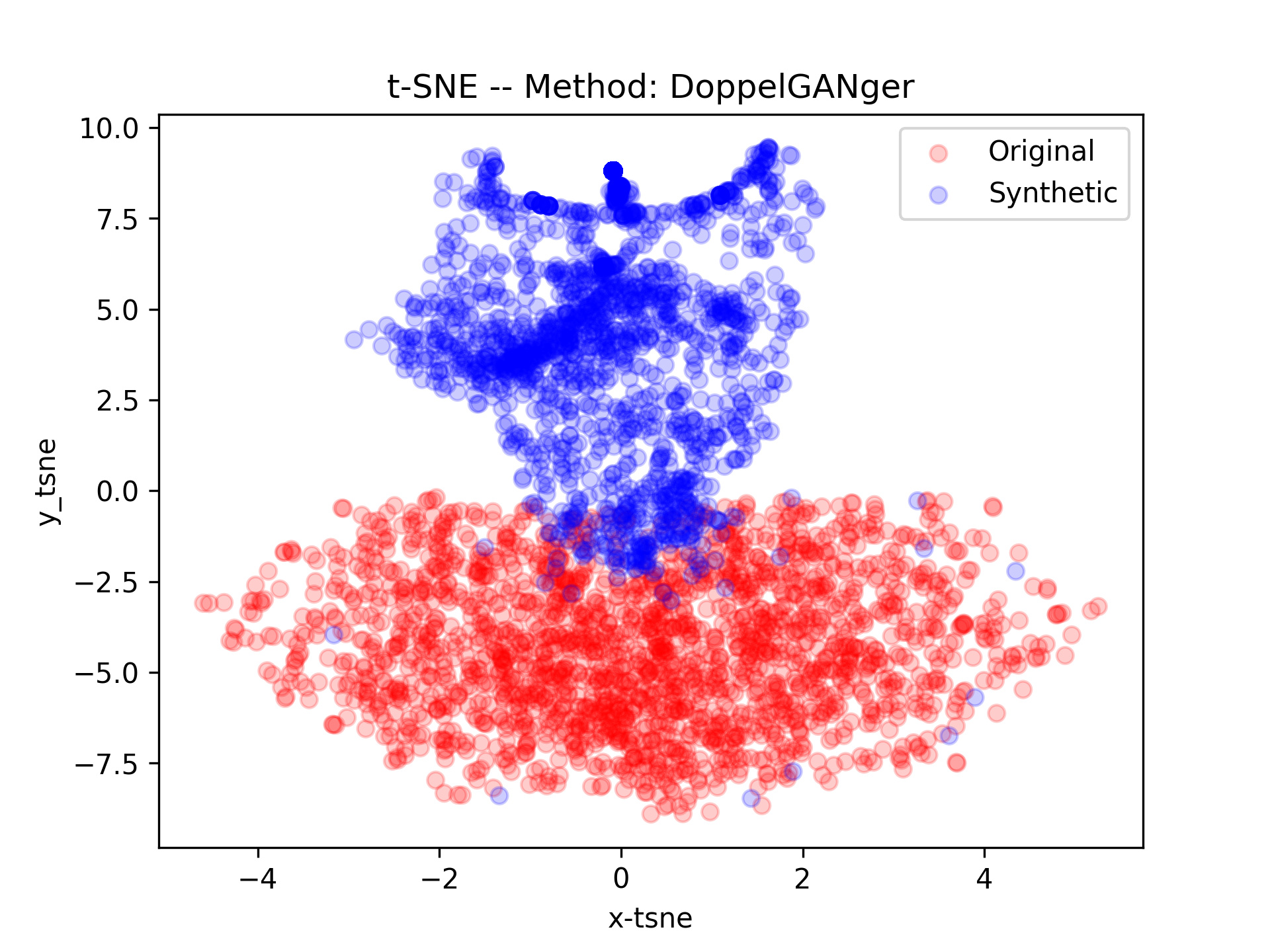}
  \caption*{\scriptsize \textbf{GARCH}}
\end{subfigure}
    \hfill
\begin{subfigure}[t]{0.15\textwidth}
  \includegraphics[width=\linewidth]{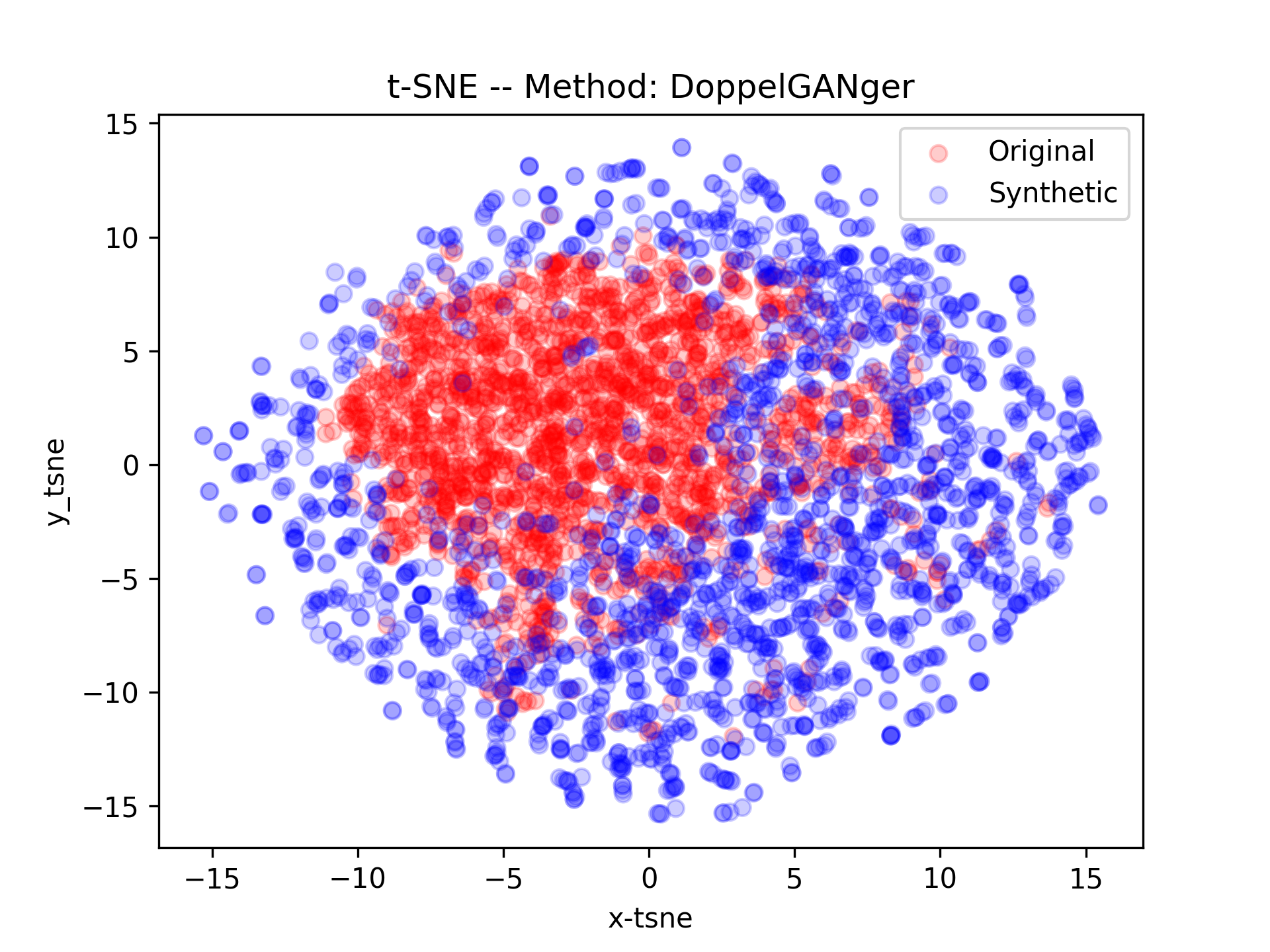}
  \caption*{\scriptsize \textbf{INAR}}
\end{subfigure}
    \hfill
\begin{subfigure}[t]{0.15\textwidth}
  \includegraphics[width=\linewidth]{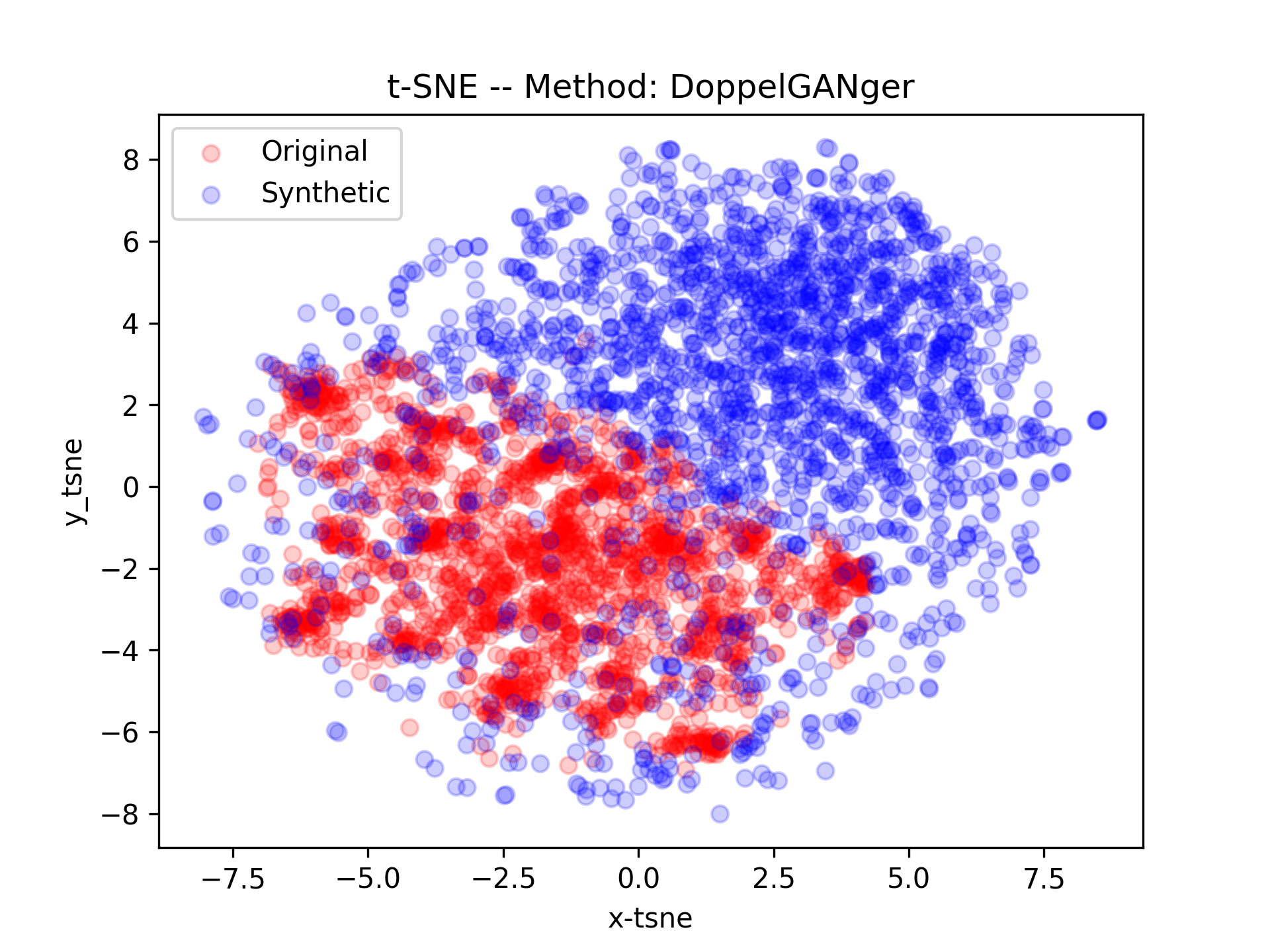}
  \caption*{\scriptsize \textbf{SETAR}}
\end{subfigure}
    \hfill
\begin{subfigure}[t]{0.15\textwidth}
  \includegraphics[width=\linewidth]{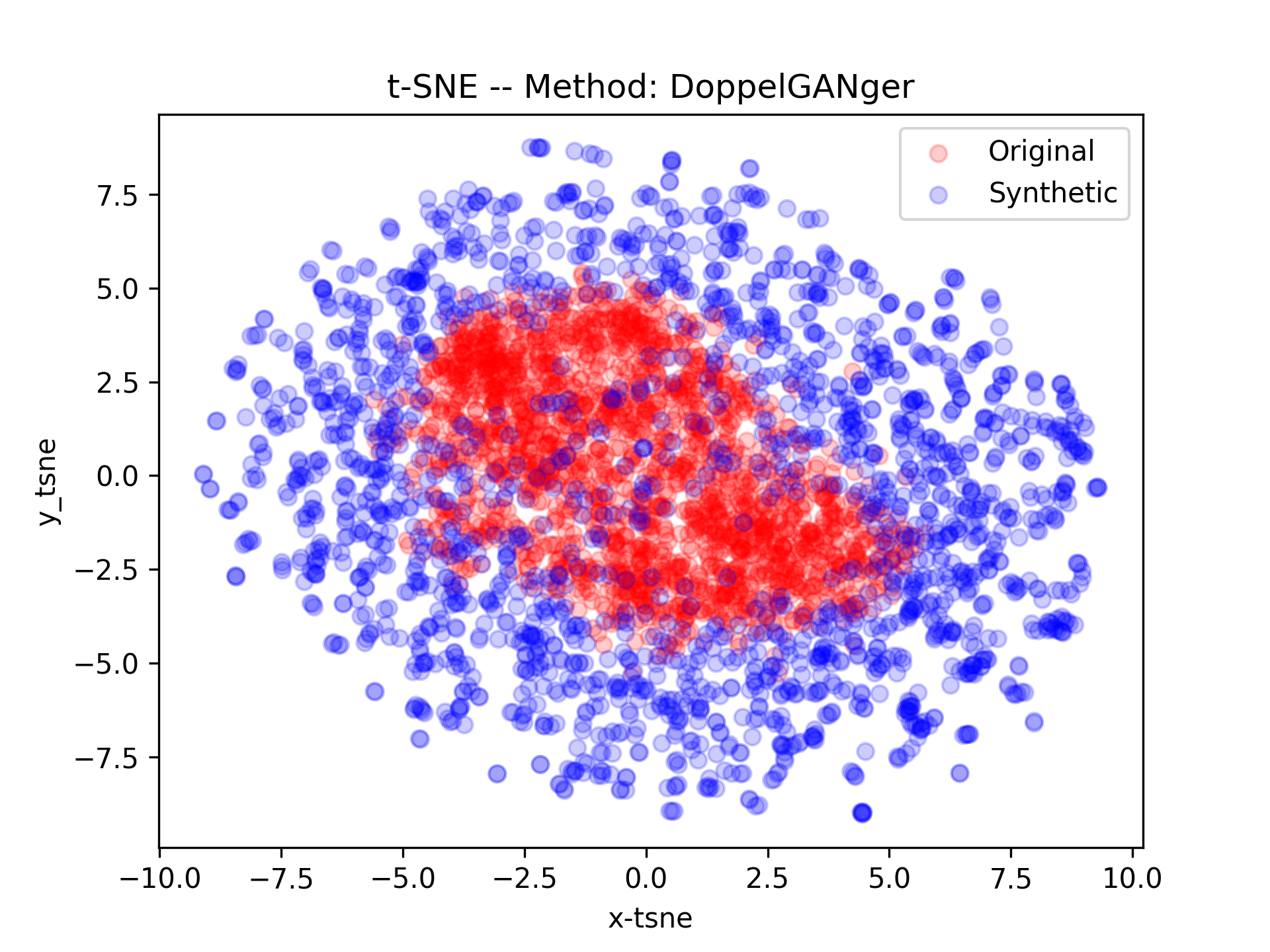}
  \caption*{\scriptsize \textbf{HMM}}
\end{subfigure}
\hspace*{\fill}
\caption[t-SNE comparison between QG-based mapping and GAN methods \textit{(continued)}]{\textit{(continued)} t-SNE projections for synthetic data generated by InvQG (top row), TimeGAN (middle row), and DoppelGANger (bottom row) for selected models  ({ARIMA}, {GARCH}, {INAR}, {SETAR} and {HMM}).}
\label{fig:t_sne_time_gan_vs_qg_2}
\end{figure}

\paragraph{Real Dataset}

\begin{figure*}[!h]
\centering
\captionsetup[subfigure]{justification=centering}

% First row
\hspace*{\fill}
\begin{subfigure}[t]{0.15\textwidth}
  \includegraphics[width=\linewidth]{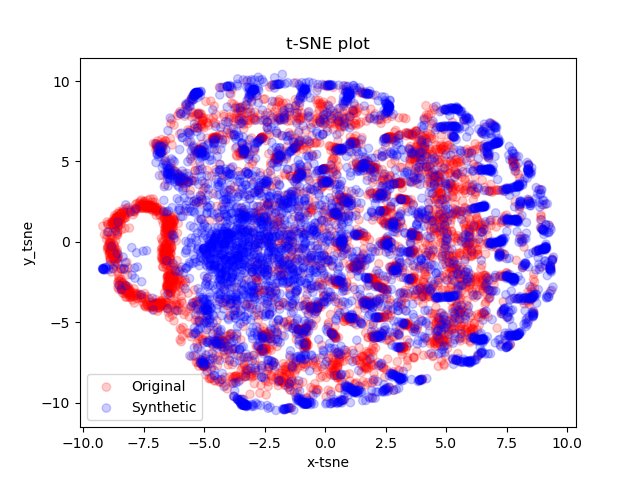}
\end{subfigure}
    \hfill
\hspace*{\fill}
\begin{subfigure}[t]{0.15\textwidth}
  \includegraphics[width=\linewidth]{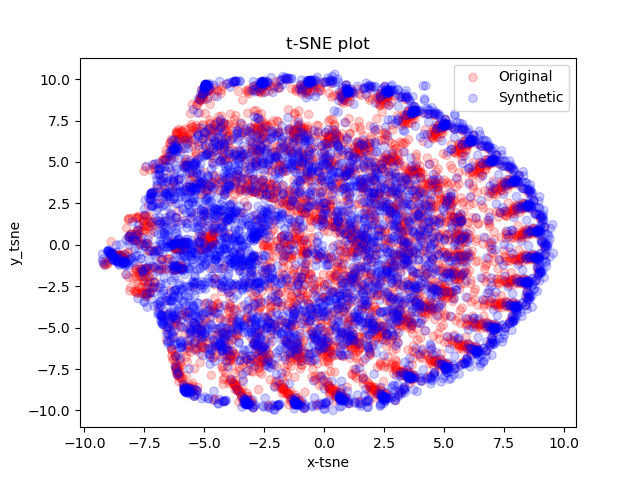}
\end{subfigure}
    \hfill
\begin{subfigure}[t]{0.15\textwidth}
  \includegraphics[width=\linewidth]{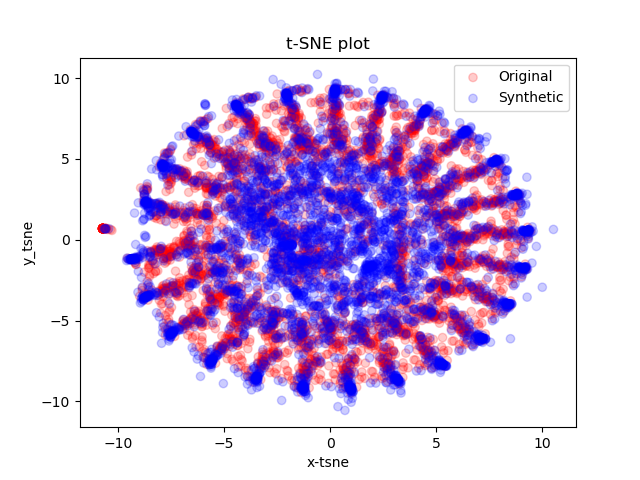}
\end{subfigure}
    \hfill
\begin{subfigure}[t]{0.15\textwidth}
  \includegraphics[width=\linewidth]{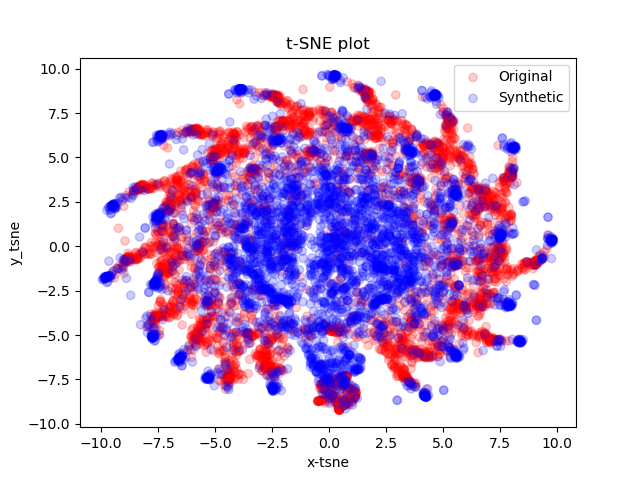}
\end{subfigure}
    \hfill
\begin{subfigure}[t]{0.15\textwidth}
  \includegraphics[width=\linewidth]{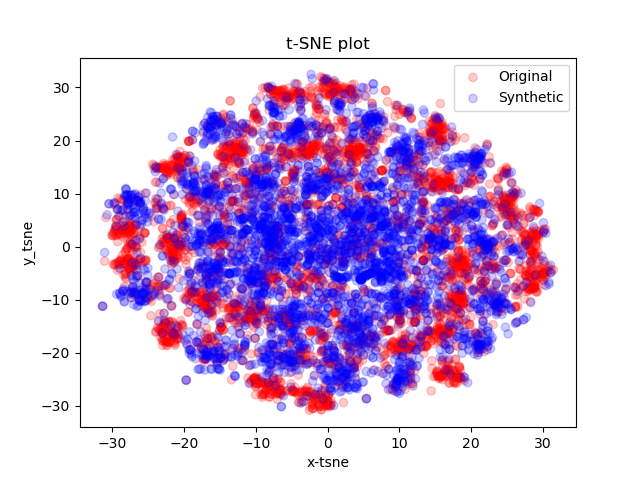}
\end{subfigure}
    \hfill
\begin{subfigure}[t]{0.15\textwidth}
  \includegraphics[width=\linewidth]{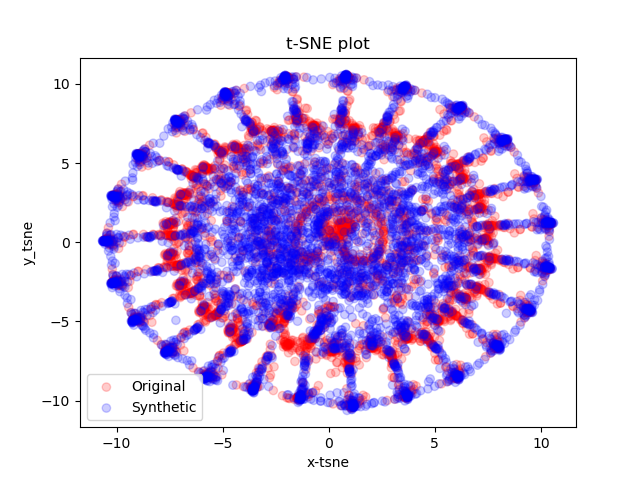}
\end{subfigure}
\hspace*{\fill}

% Second row
\hspace*{\fill}
\begin{subfigure}[t]{0.15\textwidth}
  \includegraphics[width=\linewidth]{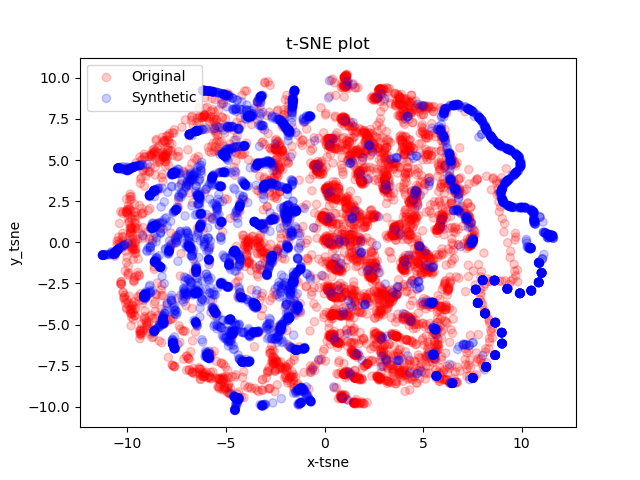}
\end{subfigure}
    \hfill
\begin{subfigure}[t]{0.15\textwidth}
  \includegraphics[width=\linewidth]{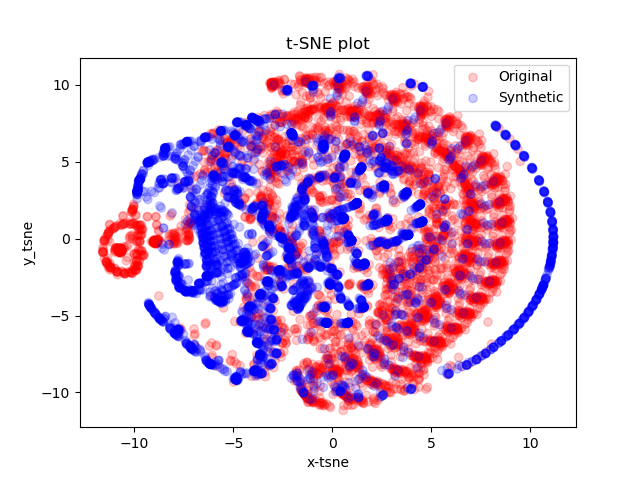}
\end{subfigure}
    \hfill
\begin{subfigure}[t]{0.15\textwidth}
  \includegraphics[width=\linewidth]{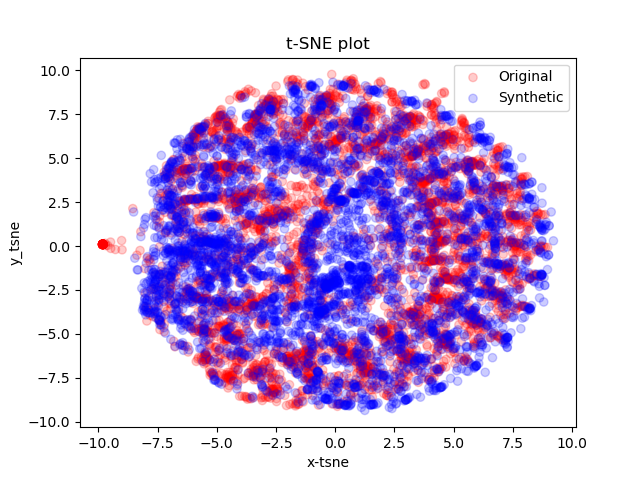}
\end{subfigure}
    \hfill
\begin{subfigure}[t]{0.15\textwidth}
  \includegraphics[width=\linewidth]{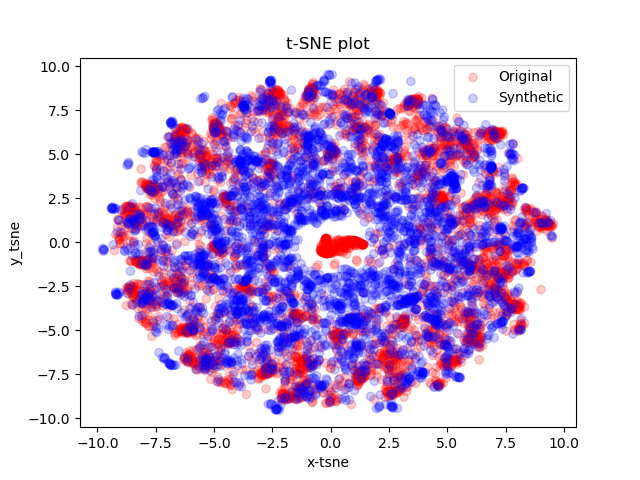}
\end{subfigure}
    \hfill
\begin{subfigure}[t]{0.15\textwidth}
  \includegraphics[width=\linewidth]{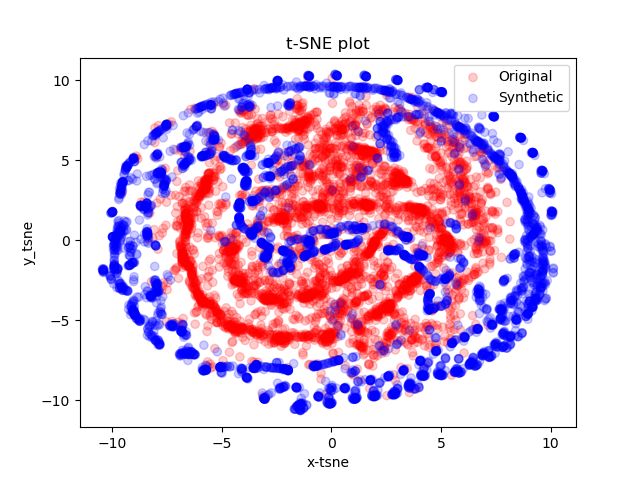}
\end{subfigure}
    \hfill
\begin{subfigure}[t]{0.15\textwidth}
  \includegraphics[width=\linewidth]{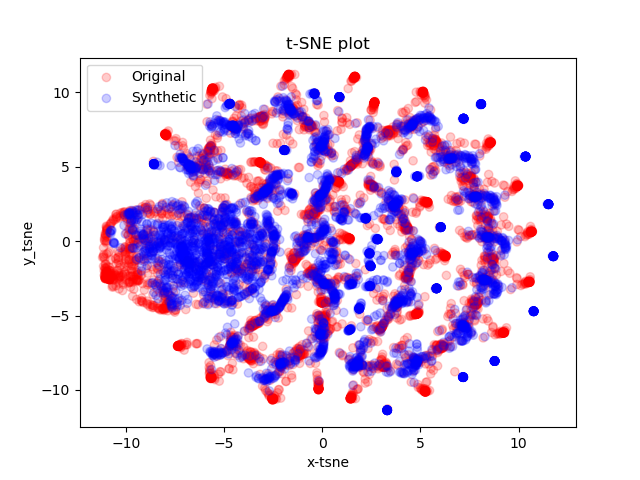}
\end{subfigure}
\hspace*{\fill}

\hspace*{\fill}
\caption[t‑SNE projections for six representative households (columns), comparing synthetic data generated by InvQG (top row)%, 
 and TimeGAN  (bottom row). t-SNE comparison between QG-based mapping and GAN methods]{t‑SNE projections for households 5, 6, 7, 8, 14, and 17 (columns), comparing synthetic data generated by InvQG (top row)%, 
 and TimeGAN  (bottom row) (see Appendix~\ref{app:t_sne_time_gan_vs_qg_real} for remaining households).}
\label{fig:t_sne_time_gan_vs_qg_real}
\end{figure*}

We extend the benchmarking analysis to the real-world smart meter dataset comprising twenty-two household electricity consumption series. For each household, we generate synthetic counterparts using InvQG and GAN-based methods. Figure~\ref{fig:t_sne_time_gan_vs_qg_real} presents t-SNE representations for six representative households (additional results are provided in Appendix~\ref{app:t_sne_time_gan_vs_qg_real}).

The results are similar to those observed for the artificial dataset. InvQG consistently generates synthetic samples that align well with the original data distribution across all households, capturing both variability and consumption patterns. 
TimeGAN generally shows the weakest alignment, with synthetic samples often occupying regions distinct from the original data.

These findings suggest that InvQG offers robust and stable performance across heterogeneous real-world time series, despite its conceptual simplicity. Importantly, it attains this level of fidelity without requiring iterative training, extensive hyperparameter tuning, or large computational resources --- factors that significantly affect the practical applicability of GAN-based methods.

\paragraph{Computational Performance}

In addition to the theoretical complexity analysis in Section~\ref{sec3}, we evaluated the empirical computational performance of the InvQG method and compared it with TimeGAN.

We reported empirical times for both InvQG and TimeGAN methods. 
For the \textit{artificial dataset}, the execution times correspond to the average execution time across all eleven time series models, and for \textit{real dataset} (smart meter dataset), the reported values correspond to the average execution time across the twenty-two household time series. 
The times for InvQG includes the construction of the QG and the generation of synthetic data, while for TimeGAN corresponds to the total time required for model training and data generation. 

For the \textit{artificial dataset} (time series of length $T = 10,000$), InvQG requires on average approximately \textbf{0.213} seconds per instance, whereas TimeGAN requires approximately \textbf{14,531.19} seconds (ie., $\approx$ 4 hours and 2 minutes). Similarly, for the \textit{real dataset} (with $T = 4,262$), InvQG requires on average  \textbf{0.066} seconds per instance, compared to \textbf{7,3302.60} seconds (ie., $\approx$ 20 hours and 21 minutes) for TimeGAN.

\newpage

Although the absolute values depend on implementation details and hardware, the results consistently show that InvQG is substantially more computationally efficient than TimeGAN. 
This difference is primarily due to the fact that InvQG does not require iterative training and operates through a direct mapping and sampling process.

\subsubsection{Sensitivity Analysis on the Number of Quantiles}

The InvQG method depends on a single parameter, $Q$, which determines the number of quantiles used to construct the transition graph and thus controls the granularity of the underlying representation. In this section, we assess the sensitivity of the generated synthetic time series to this parameter in order to evaluate the robustness of the method.

We consider a range of values  $Q \in \{8, 16, 32, 64, 100, 128\}$ and generate synthetic series for the \textit{artificial dataset} following the same experimental protocol as in Section~\ref{sec_fidelity}. For each configuration, statistical features are extracted and analyzed  through paired differences (\textit{synthetic} $-$ \textit{original}).

Across all values of $Q,$ the overall behavior  of the statistical features remains relatively stable, with differences consistently centered  close to zero for most models. This indicates that InvQG preserves the main statistical properties of the original data across a broad range of quantile resolutions.

More detailed analysis reveals two expected effects. For small values of $Q$, the representation becomes coarse, leading to a loss of fine-grained temporal structure and slightly larger deviations in autocorrelation-related features. Conversely, increasing $Q$ improves the resolution of the transition structure, but the gains quickly diminish beyond moderate values, with no substantial improvement in overall fidelity.

\begin{figure*}[th]
\centering
  \includegraphics[width=0.94\linewidth]{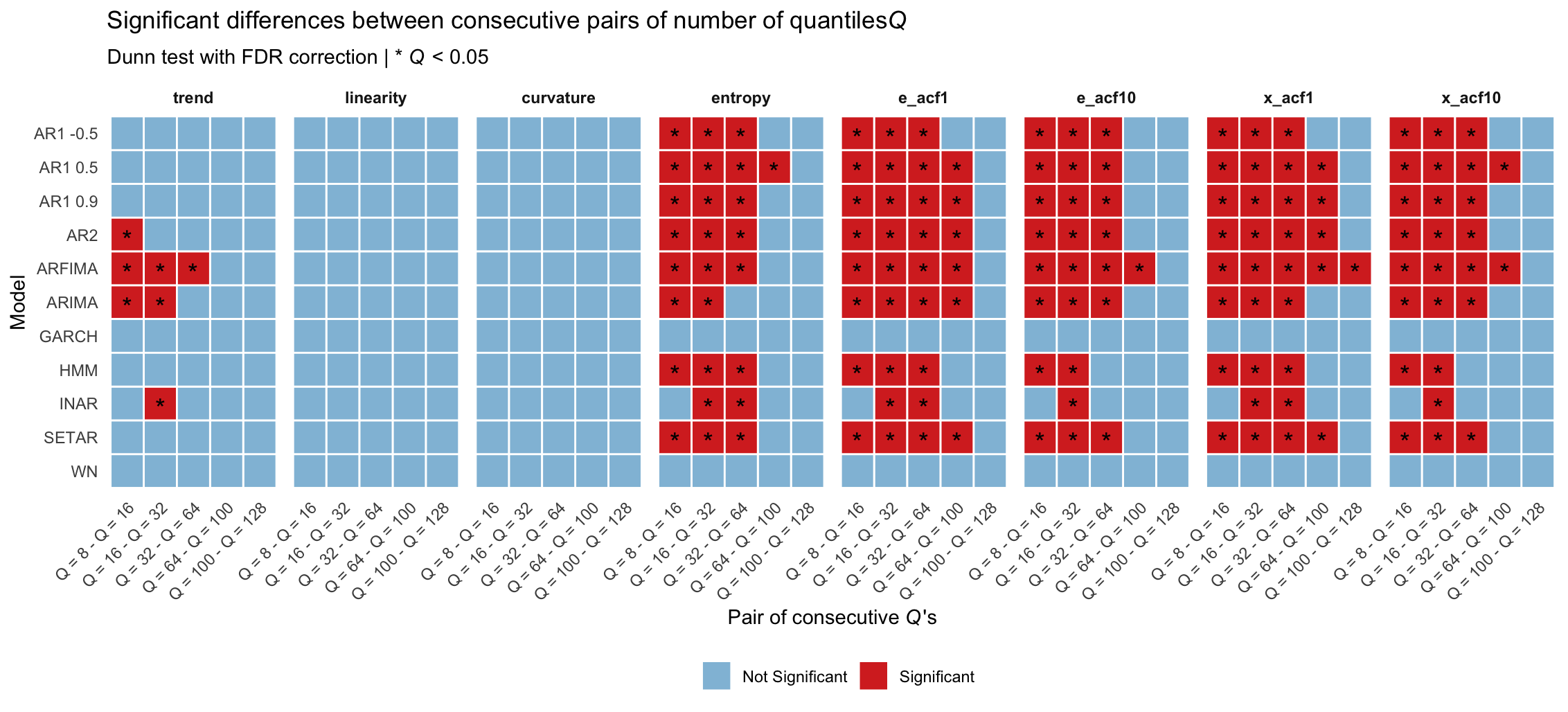}
\caption{Kruskal-Wallis test results per model, feature and pair of number of quantiles ($Q = i - Q = j$). Red cells indicate statistically significant median (\textit{synthetic-original}) differences for quantiles $Q = i$ and $Q = j$ (Dunn test, FDR-corrected $p < 0.05$). }  
  \label{fig:hmap}
\end{figure*}

To formally assess the influence of $Q$, we apply non-parametric statistical tests to compare feature differences across parameter settings. The results summarized in Figure~\ref{fig:hmap} indicate that, although some statistically significant differences can be detected for specific features and models, these differences are generally small in magnitude and do not alter the qualitative behavior  of the synthetic data.

Overall, this analysis shows that InvQG is robust to the choice of $Q$ within a wide and practically relevant range --- in particular, the literature suggests the rule $Q=2T^\frac{1}{3}$~\cite{campanharo2020application} (for $T=10,000$ we have $Q \sim 43$). 
Moderate values of $Q$ provide a good trade-off between fidelity and computational efficiency, while extreme values either reduce detail (small $Q$) or yield diminishing gains (large $Q$). This confirms that the method is stable and does not require fine parameter tuning for effective use.

The results should be interpreted considering  the modelling assumptions underlying InvQG. The method relies on a first order transition representation, which inherently prioritizes the preservation of short-term temporal dependencies and marginal distributional properties. While this formulation limits its ability to fully capture long range or higher order dynamics --- particularly in processes exhibiting strong persistence or cyclic structure --- it provides a transparent and computationally efficient mechanism for synthetic data generation. In contrast to GAN-based approaches, which offer greater expressive flexibility but require extensive training and may suffer from instability or reduced reproducibility. InvQG achieves a balance between fidelity, interpretability, and efficiency. Importantly, the observed discrepancies are systematic and well understood, rather than arbitrary, and remain localized to specific classes of models. Furthermore, the sensitivity analysis confirms that the method is robust to the choice of its single parameter, $Q,$ supporting its practical applicability.

\section{Conclusion}\label{sec6}

Synthetic time series generation plays a critical role in enabling data‑driven modelling in domains where access to real data is limited or restricted. In this work, we investigated a framework for synthetic time series generation based on complex network mappings, leveraging the Quantile Graph (QG) representation and its inverse mapping (InvQG). This approach provides a structured and interpretable mechanism for generating synthetic data directly from the transition dynamics of the original series.

Our empirical evaluation indicates that the proposed method achieves high statistical fidelity across a diverse set of time series models, effectively preserving marginal distributions, trend behavior, and short‑term temporal dependencies. In addition, the synthetic data retain useful structural information, as evidenced by both statistical feature analysis and network‑based representations, supporting their applicability in downstream tasks such as clustering.

When compared with state‑of‑the‑art GAN-based approaches, namely TimeGAN and DoppelGANger, InvQG shows competitive performance in reproducing the empirical distribution of the data, while offering substantial advantages in terms of interpretability, computational efficiency, and ease of use. In particular, the absence of a training process and the presence of a single, robust parameter make the method attractive in practical settings where reliability and reproducibility are critical. 

This study also highlights important limitations. Due to its reliance on a first‑order transition framework, InvQG has a limited ability to capture long‑range or higher‑order temporal dependencies, which becomes evident in processes with strong persistence or cyclic dynamics. However, these limitations are systematic and well understood, and do not compromise the method’s effectiveness in preserving the essential characteristics required for many analytical tasks.
Overall, the results position InvQG as a competitive, efficient, and interpretable alternative for synthetic time series generation, particularly in scenarios where computational cost, stability, and transparency are key considerations.

\newpage
Future work may explore extensions of the framework to address its current limitations, including the incorporation of higher‑order transition structures, alternative quantile representations, and extensions to multivariate time series. Further evaluation across additional real‑world domains may also provide deeper insight into the applicability of the approach in complex data environments.

\section*{Availability of data and materials}

The raw data are available from the corresponding author upon request and available at \url{https://github.com/vanessa-silva/InvQG}.

\section{Acknowledgments}

We thank the reviewers for their constructive comments and suggestions, which have helped improve the manuscript. This work is funded by national funds through FCT – Fundação para a Ciência e a Tecnologia, I.P., under the project 2023.13039.PEX (\url{https://doi.org/10.54499/2023.13039.PEX}) and the support UID/50014/2023 (\url{https://doi.org/10.54499/UID/50014/2023}).

\bibliographystyle{apalike}
\bibliography{sn-bibliography}

\appendix

\section{Mean and SD Tables of Differences}

\begin{table}[H]
\centering
\caption[Mean and standard deviation of the paired samples (\textit{synthetic}-\textit{original}) of statistical features.]
{Mean and standard deviation of the paired samples (\textit{synthetic}-\textit{original}) of statistical features. The values correspond to the mean and standard deviation of the 100 instances of each original and synthetic time series models for each feature.}
\label{tab:mean_sd_diffs}
\begin{tabular}{|l|c|c|c|c|c|c|c|c|}
  \hline
\multicolumn{1}{|c|}{\textbf{Model}} & \multicolumn{1}{|c|}{\textbf{trend}} & \multicolumn{1}{|c|}{\textbf{linea.}} & \multicolumn{1}{|c|}{\textbf{curvat.}} & \multicolumn{1}{|c|}{\textbf{entro.}} & \multicolumn{1}{|c|}{\textbf{e\_acf1}} & \multicolumn{1}{|c|}{\textbf{e\_acf10}} & \multicolumn{1}{|c|}{\textbf{x\_acf1}} & \multicolumn{1}{|c|}{\textbf{x\_acf10}} \\  
  \hline
\multirow{2}{*}{\textbf{AR -0.5}} & 0.000 & 0.045 & 0.039 & 0.001 & 0.004 & -0.004 & 0.004 & -0.004 \\ 
                                  & (0.000) &  (0.854) &  (0.712) & (0.001) & (0.009) & (0.019) & (0.009) & (0.019) \\ 
\hline
\multirow{2}{*}{\textbf{AR 0.5}} & 0.001 & 0.074 & -0.001 & 0.001 & -0.005 & -0.007 & -0.005 & -0.006 \\ 
                                 & (0.003) &  (2.467) &  (2.453) & (0.001) & (0.008) & (0.017) & (0.008) & (0.017) \\ 
\hline
\multirow{2}{*}{\textbf{AR 0.9}} & -0.001 & 0.827 & 0.536 & 0.004 & -0.007 & -0.081 & -0.006 & -0.087 \\ 
                                 & (0.017) &  (6.019) &  (5.906) & (0.005) & (0.005) & (0.175) & (0.004) & (0.187) \\ 
\hline
\multirow{2}{*}{\textbf{AR2}} & 0.026 & -0.525 & 0.269 & -0.007 & -0.011 & 0.833 & -0.007 & 0.957 \\ 
                              & (0.010) &  (3.824) &  (3.845) & (0.004) & (0.006) & (0.113) & (0.006) & (0.122) \\ 
\hline
\multirow{2}{*}{\textbf{ARFIMA}} & 0.039 & -0.439 & 0.891 & -0.074 & -0.007 & 0.886 & -0.003 & 0.655 \\ 
                                 & (0.087) & (27.135) & (23.596) & (0.019) & (0.003) & (0.252) & (0.002) & (0.175) \\ 
\hline
\multirow{2}{*}{\textbf{ARIMA}} & -0.043 & 6.021 & 5.470 & 0.065 & -0.042 & -1.026 & -0.004 & -0.135 \\ 
                                & (0.086) & (63.112) & (59.249) & (0.108) & (0.095) & (1.199) & (0.009) & (0.216) \\ 
\hline
\multirow{2}{*}{\textbf{GARCH}} & 0.000 & 0.124 & 0.171 & 0.000 & -0.001 & 0.000 & -0.001 & 0.000 \\ 
                                & (0.001) &  (1.439) &  (1.368) & (0.000) & (0.012) & (0.001) & (0.012) & (0.001) \\ 
\hline
\multirow{2}{*}{\textbf{HMM}} & -0.006 & -0.333 & 0.351 & 0.007 & 0.000 & -0.104 & -0.004 & -0.118 \\ 
                              & (0.003) &  (2.358) &  (1.997) & (0.001) & (0.012) & (0.013) & (0.011) & (0.015) \\ 
\hline
\multirow{2}{*}{\textbf{INAR}} & 0.000 & 0.163 & -0.115 & 0.001 & -0.004 & -0.007 & -0.004 & -0.007 \\ 
                               & (0.001) &  (2.086) &  (1.779) & (0.001) & (0.009) & (0.020) & (0.009) & (0.021) \\ 
\hline
\multirow{2}{*}{\textbf{SETAR}} & 0.000 & -0.142 & 0.009 & 0.000  & -0.011 & -0.003 & -0.011 & -0.003 \\ 
                                & (0.000) &  (0.973) &  (0.950) & (0.000) & (0.007) & (0.004) & (0.007) & (0.004) \\ 
\hline
\multirow{2}{*}{\textbf{WN}} & 0.000 & -0.230 & -0.024 & 0.000 & 0.001 & 0.000 & 0.001 & 0.000 \\ 
                             & (0.001) &  (1.481) &  (1.207) & (0.000) & (0.010) & (0.001) & (0.010) & (0.001) \\ 
   \hline
\end{tabular}
\end{table}

\newpage
\section{Paired Statistically Significant}

{
\begin{longtable}{llccccc}

\caption{Full Wilcoxon signed-rank test results for all model--feature combinations (88 tests). Median: median difference (synthetic $-$ original). $W$: test statistic. $p$: raw $p$-value. $p_{\text{Bonf}}$: Bonferroni-corrected $p$-value. $p_{\text{FDR}}$: Benjamini--Hochberg corrected $p$-value. $^{*}$ $p_{\text{FDR}} < 0.05$.}
\label{app_tab:stat_tests}\\

\toprule
\textbf{Model} & \textbf{Feature} & \textbf{Median} &
$\boldsymbol{W}$ & $\boldsymbol{p}$ &
$\boldsymbol{p}_{\text{Bonf}}$ &
$\boldsymbol{p}_{\text{FDR}}$\\
\midrule
\endfirsthead

\caption{Full Wilcoxon signed-rank test results for all model--feature combinations (88 tests).  \textit{(continued)}}\\
\toprule
\textbf{Model} & \textbf{Feature} & \textbf{Median} &
$\boldsymbol{W}$ & $\boldsymbol{p}$ &
$\boldsymbol{p}_{\text{Bonf}}$ &
$\boldsymbol{p}_{\text{FDR}}$\\
\midrule
\endhead

\bottomrule
\endfoot

\bottomrule
\endlastfoot
\multirow{8}{*}{\textbf{AR -0.5}} & trend & -0.0000 & 2463 & 0.833 & 1.000 & 0.872\\
 & linearity & 0.0603 & 2668 & 0.624 & 1.000 & 0.687\\
 & curvature & 0.0697 & 2716 & 0.512 & 1.000 & 0.609\\
 & entropy & 0.0005 & 3560 & 3.75e-04 & 0.033 & 8.05e-04$^{*}$\\
 & e\_acf1 & 0.0031 & 3726 & 3.66e-05 & 0.003 & 8.96e-05$^{*}$\\
 & e\_acf10 & -0.0034 & 1953 & 0.049 & 1.000 & 0.089\\
 & x\_acf1 & 0.0031 & 3726 & 3.66e-05 & 0.003 & 8.96e-05$^{*}$\\
 & x\_acf10 & -0.0035 & 1952 & 0.049 & 1.000 & 0.089\\
\midrule
\multirow{8}{*}{\textbf{AR 0.5}} & trend & 0.0004 & 2978 & 0.120 & 1.000 & 0.192\\
 & linearity & 0.2242 & 2699 & 0.551 & 1.000 & 0.629\\
 & curvature & -0.2634 & 2507 & 0.952 & 1.000 & 0.974\\
 & entropy & 0.0005 & 3888 & 2.80e-06 & 2.47e-04 & 7.71e-06$^{*}$\\
 & e\_acf1 & -0.0045 & 1035 & 3.03e-07 & 2.67e-05 & 8.90e-07$^{*}$\\
 & e\_acf10 & -0.0056 & 1504 & 4.50e-04 & 0.040 & 9.43e-04$^{*}$\\
 & x\_acf1 & -0.0042 & 1062 & 4.94e-07 & 4.35e-05 & 1.40e-06$^{*}$\\
 & x\_acf10 & -0.0055 & 1558 & 8.90e-04 & 0.078 & 0.002$^{*}$\\
\midrule
\multirow{8}{*}{\textbf{AR 0.9}} & trend & -0.0017 & 2261 & 0.365 & 1.000 & 0.482\\
 & linearity & 0.3603 & 2835 & 0.287 & 1.000 & 0.408\\
 & curvature & 0.4751 & 2764 & 0.412 & 1.000 & 0.518\\
 & entropy & 0.0037 & 4322 & 6.54e-10 & 5.75e-08 & 2.40e-09$^{*}$\\
 & e\_acf1 & -0.0066 & 114 & 1.15e-16 & 1.01e-14 & 5.06e-16$^{*}$\\
 & e\_acf10 & -0.0870 & 1232 & 8.83e-06 & 7.77e-04 & 2.29e-05$^{*}$\\
 & x\_acf1 & -0.0066 & 119 & 1.33e-16 & 1.17e-14 & 5.57e-16$^{*}$\\
 & x\_acf10 & -0.1111 & 1183 & 3.98e-06 & 3.50e-04 & 1.06e-05$^{*}$\\
\midrule
\multirow{8}{*}{\textbf{AR2}} & trend & 0.0250 & 5050 & 3.96e-18 & 3.48e-16 & 2.39e-17$^{*}$\\
 & linearity & -0.3576 & 2163 & 0.214 & 1.000 & 0.319\\
 & curvature & 0.1042 & 2700 & 0.549 & 1.000 & 0.629\\
 & entropy & -0.0063 & 32 & 1.03e-17 & 9.10e-16 & 5.36e-17$^{*}$\\
 & e\_acf1 & -0.0120 & 16 & 6.41e-18 & 5.64e-16 & 3.52e-17$^{*}$\\
 & e\_acf10 & 0.8241 & 5050 & 3.96e-18 & 3.48e-16 & 2.39e-17$^{*}$\\
 & x\_acf1 & -0.0076 & 171 & 5.86e-16 & 5.16e-14 & 2.24e-15$^{*}$\\
 & x\_acf10 & 0.9550 & 5050 & 3.96e-18 & 3.48e-16 & 2.39e-17$^{*}$\\
\newpage
\midrule
\multirow{8}{*}{\textbf{ARFIMA}} & trend & 0.0354 & 3715 & 4.32e-05 & 0.004 & 1.03e-04$^{*}$\\
 & linearity & 0.5163 & 2520 & 0.988 & 1.000 & 0.988\\
 & curvature & -0.6419 & 2576 & 0.862 & 1.000 & 0.893\\
 & entropy & -0.0735 & 0 & 3.96e-18 & 3.48e-16 & 2.39e-17$^{*}$\\
 & e\_acf1 & -0.0064 & 0 & 3.96e-18 & 3.48e-16 & 2.39e-17$^{*}$\\
 & e\_acf10 & 0.8381 & 5050 & 3.96e-18 & 3.48e-16 & 2.39e-17$^{*}$\\
 & x\_acf1 & -0.0031 & 0 & 3.96e-18 & 3.48e-16 & 2.39e-17$^{*}$\\
 & x\_acf10 & 0.6579 & 5050 & 3.96e-18 & 3.48e-16 & 2.39e-17$^{*}$\\
\midrule
\multirow{8}{*}{\textbf{ARIMA}} & trend & -0.0333 & 988 & 1.27e-07 & 1.12e-05 & 3.86e-07$^{*}$\\
 & linearity & 1.8082 & 2746 & 0.448 & 1.000 & 0.556\\
 & curvature & 4.8283 & 2844 & 0.273 & 1.000 & 0.401\\
 & entropy & 0.0656 & 4082 & 8.71e-08 & 7.67e-06 & 2.84e-07$^{*}$\\
 & e\_acf1 & -0.0213 & 0 & 3.96e-18 & 3.48e-16 & 2.39e-17$^{*}$\\
 & e\_acf10 & -0.7503 & 0 & 3.96e-18 & 3.48e-16 & 2.39e-17$^{*}$\\
 & x\_acf1 & -0.0018 & 0 & 3.96e-18 & 3.48e-16 & 2.39e-17$^{*}$\\
 & x\_acf10 & -0.0871 & 149 & 3.14e-16 & 2.77e-14 & 1.26e-15$^{*}$\\
\midrule
\multirow{8}{*}{\textbf{GARCH}} & trend & -0.0001 & 2245 & 0.337 & 1.000 & 0.470\\
 & linearity & 0.2034 & 2800 & 0.345 & 1.000 & 0.475\\
 & curvature & 0.2506 & 2909 & 0.187 & 1.000 & 0.289\\
 & entropy & 0.0000 & 888 & 0.118 & 1.000 & 0.192\\
 & e\_acf1 & 0.0005 & 2358 & 0.567 & 1.000 & 0.637\\
 & e\_acf10 & -0.0000 & 2779 & 0.383 & 1.000 & 0.496\\
 & x\_acf1 & 0.0000 & 2342 & 0.530 & 1.000 & 0.622\\
 & x\_acf10 & 0.0000 & 2789 & 0.365 & 1.000 & 0.482\\
\midrule
\multirow{8}{*}{\textbf{HMM}} & trend & -0.0053 & 1 & 4.08e-18 & 3.59e-16 & 2.39e-17$^{*}$\\
 & linearity & -0.1991 & 2214 & 0.286 & 1.000 & 0.408\\
 & curvature & 0.3502 & 3009 & 0.096 & 1.000 & 0.163\\
 & entropy & 0.0074 & 5050 & 3.96e-18 & 3.48e-16 & 2.39e-17$^{*}$\\
 & e\_acf1 & 0.0002 & 2539 & 0.963 & 1.000 & 0.974\\
 & e\_acf10 & -0.1055 & 0 & 3.96e-18 & 3.48e-16 & 2.39e-17$^{*}$\\
 & x\_acf1 & -0.0044 & 1431 & 1.70e-04 & 0.015 & 3.94e-04$^{*}$\\
 & x\_acf10 & -0.1184 & 0 & 3.96e-18 & 3.48e-16 & 2.39e-17$^{*}$\\
 \newpage
\midrule
\multirow{8}{*}{\textbf{INAR}} & trend & 0.0000 & 2607 & 0.779 & 1.000 & 0.836\\
 & linearity & 0.0552 & 2734 & 0.473 & 1.000 & 0.579\\
 & curvature & -0.0782 & 2329 & 0.501 & 1.000 & 0.605\\
 & entropy & 0.0004 & 3482 & 0.001 & 0.089 & 0.002$^{*}$\\
 & e\_acf1 & -0.0023 & 1452 & 2.26e-04 & 0.020 & 5.11e-04$^{*}$\\
 & e\_acf10 & -0.0057 & 1711 & 0.005 & 0.454 & 0.010$^{*}$\\
 & x\_acf1 & -0.0029 & 1455 & 2.36e-04 & 0.021 & 5.19e-04$^{*}$\\
 & x\_acf10 & -0.0057 & 1723 & 0.006 & 0.515 & 0.011$^{*}$\\
\midrule
\multirow{8}{*}{\textbf{SETAR}} & trend & 0.0000 & 2591 & 0.822 & 1.000 & 0.871\\
 & linearity & -0.2670 & 2040 & 0.096 & 1.000 & 0.163\\
 & curvature & -0.0272 & 2441 & 0.774 & 1.000 & 0.836\\
 & entropy & 0.0002 & 4074 & 1.01e-07 & 8.92e-06 & 3.19e-07$^{*}$\\
 & e\_acf1 & -0.0112 & 41 & 1.35e-17 & 1.19e-15 & 6.61e-17$^{*}$\\
 & e\_acf10 & -0.0024 & 752 & 1.10e-09 & 9.66e-08 & 3.72e-09$^{*}$\\
 & x\_acf1 & -0.0113 & 49 & 1.72e-17 & 1.51e-15 & 7.95e-17$^{*}$\\
 & x\_acf10 & -0.0025 & 738 & 8.12e-10 & 7.15e-08 & 2.86e-09$^{*}$\\
\midrule
\multirow{8}{*}{\textbf{WN}} & trend & -0.0001 & 2146 & 0.193 & 1.000 & 0.293\\
 & linearity & -0.1983 & 2051 & 0.104 & 1.000 & 0.172\\
 & curvature & 0.0840 & 2360 & 0.572 & 1.000 & 0.637\\
 & entropy & 0.0000 & 405 & 0.140 & 1.000 & 0.220\\
 & e\_acf1 & 0.0014 & 2788 & 0.367 & 1.000 & 0.482\\
 & e\_acf10 & 0.0001 & 3028 & 0.084 & 1.000 & 0.148\\
 & x\_acf1 & 0.0009 & 2764 & 0.412 & 1.000 & 0.518\\
 & x\_acf10 & 0.0001 & 3108 & 0.045 & 1.000 & 0.085\\
\bottomrule
\end{longtable}
}

\newpage
\section{Simulated Dataset: Original \textit{vs} Synthetic Counterparts}

\begin{figure}[!h]
    \centering
    \begin{subfigure}{0.5\textwidth}
        \centering
        \includegraphics[width=\linewidth]{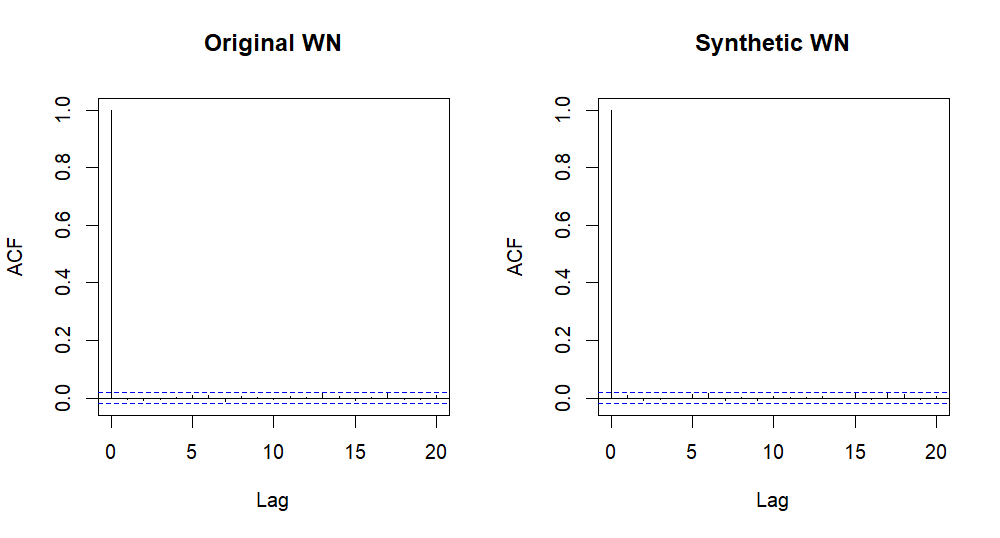}
        \caption{ACF Plots}
        \label{fig:acf_plot_wn}
    \end{subfigure}
    \hfill
    \begin{subfigure}{0.45\textwidth}
        \centering
        \includegraphics[width=\linewidth]{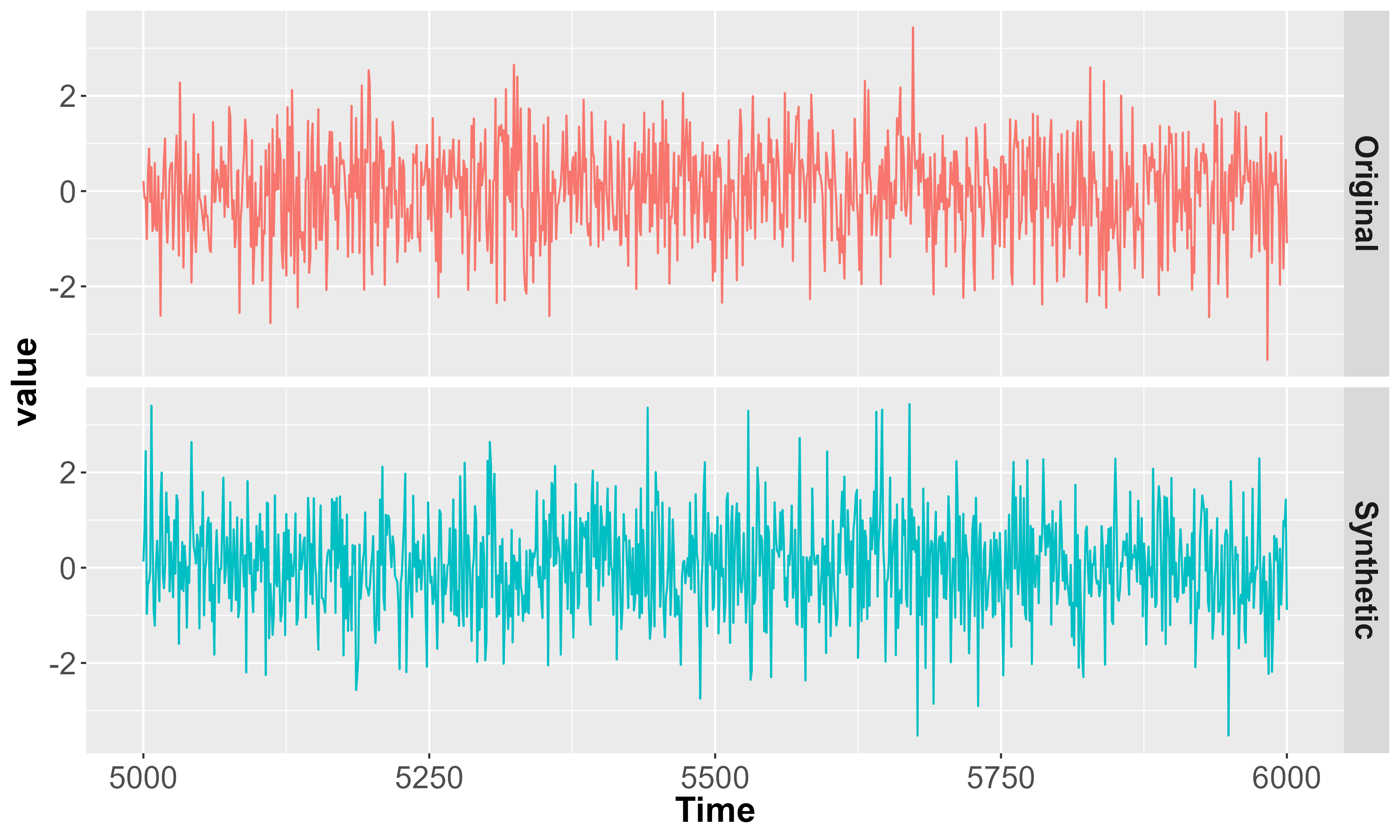}
        \caption{Time Series}
        \label{fig:ts_plot_wn}
    \end{subfigure}
    \caption{Comparison between an original and synthetic \textbf{WN} time series. (a) Autocorrelation function (ACF). (b) Time series (first 1000 observations).}
    \label{fig:acf_ts_plot_wn}
\end{figure}

\begin{figure}[!h]
    \centering
    \begin{subfigure}{0.5\textwidth}
        \centering
        \includegraphics[width=\linewidth]{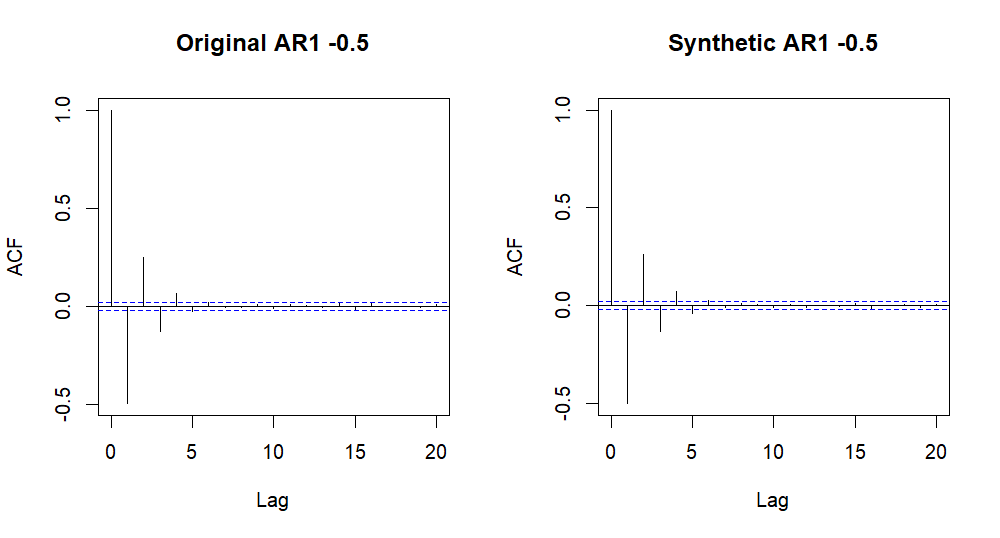}
        \caption{ACF Plots}
        \label{fig:acf_plot_ar1-5}
    \end{subfigure}
    \hfill
    \begin{subfigure}{0.45\textwidth}
        \centering
        \includegraphics[width=\linewidth]{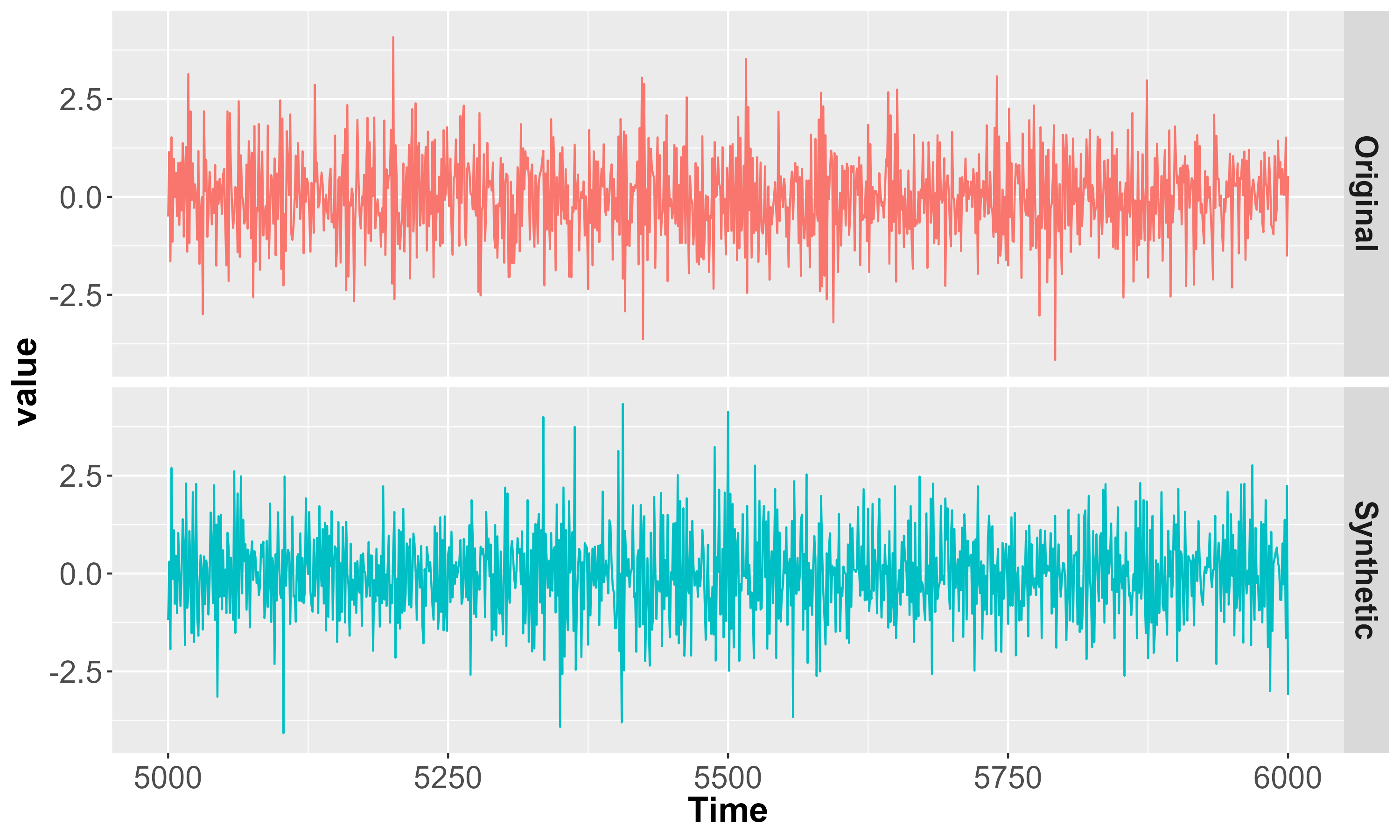}
        \caption{Time Series}
        \label{fig:ts_plot_ar1-5}
    \end{subfigure}
    \caption{Comparison between an original and synthetic \textbf{AR1 -0.5} time series. (a) Autocorrelation function (ACF). (b) Time series (first 1000 observations).}
    \label{fig:acf_ts_plot_ar1-5}
\end{figure}

\begin{figure}[!h]
    \centering
    \begin{subfigure}{0.5\textwidth}
        \centering
        \includegraphics[width=\linewidth]{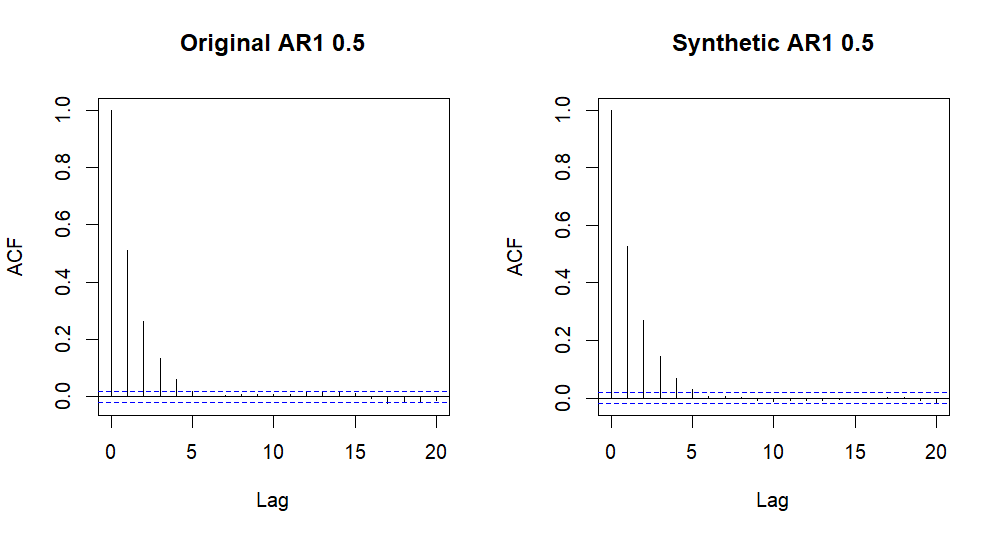}
        \caption{ACF Plots}
        \label{fig:acf_plot_ar15}
    \end{subfigure}
    \hfill
    \begin{subfigure}{0.45\textwidth}
        \centering
        \includegraphics[width=\linewidth]{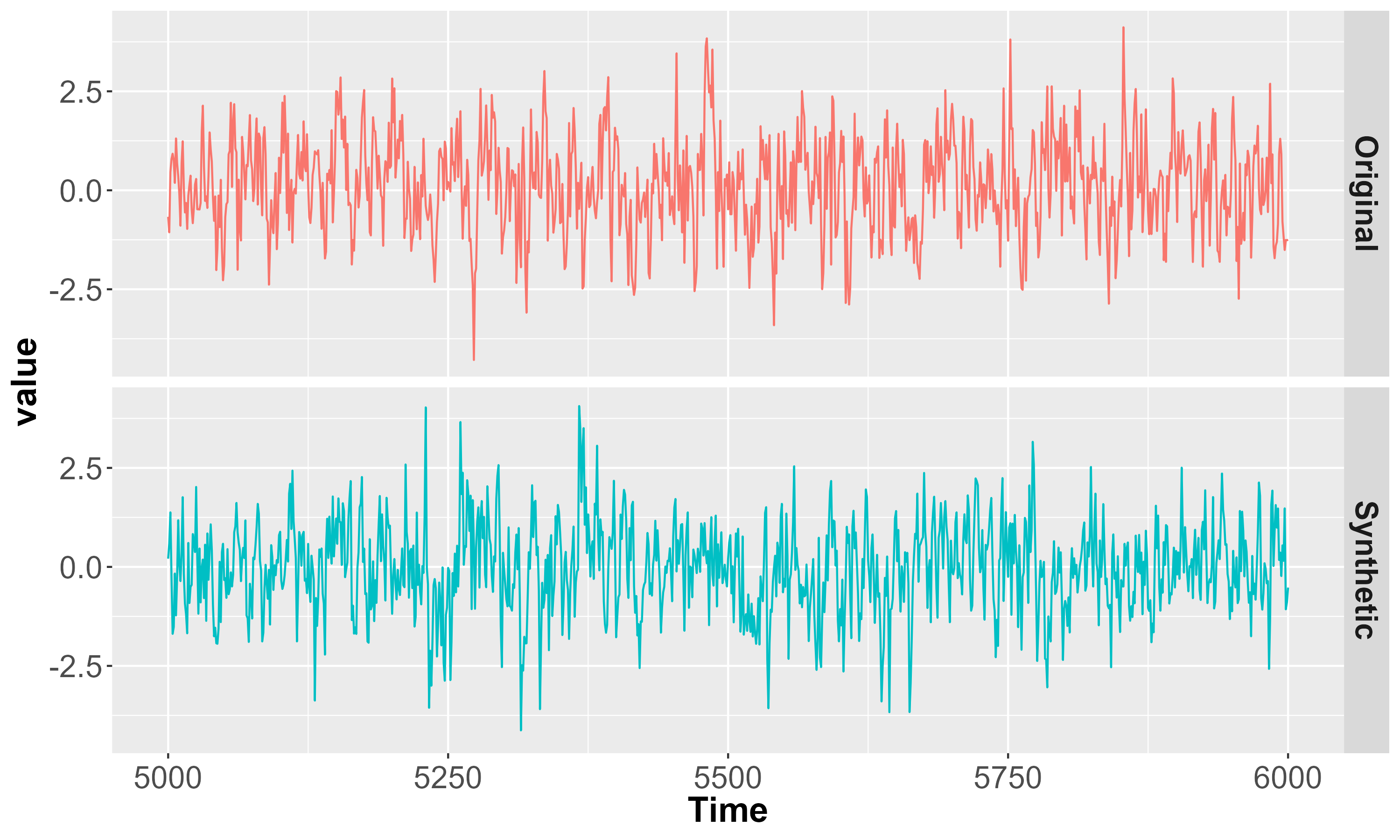}
        \caption{Time Series}
        \label{fig:ts_plot_ar15}
    \end{subfigure}
    \caption{Comparison between an original and synthetic \textbf{AR1 0.5} time series. (a) Autocorrelation function (ACF). (b) Time series (first 1000 observations).  }
    \label{fig:acf_ts_plot_ar15}
\end{figure}

\begin{figure}[!h]
    \centering
    \begin{subfigure}{0.5\textwidth}
        \centering
        \includegraphics[width=\linewidth]{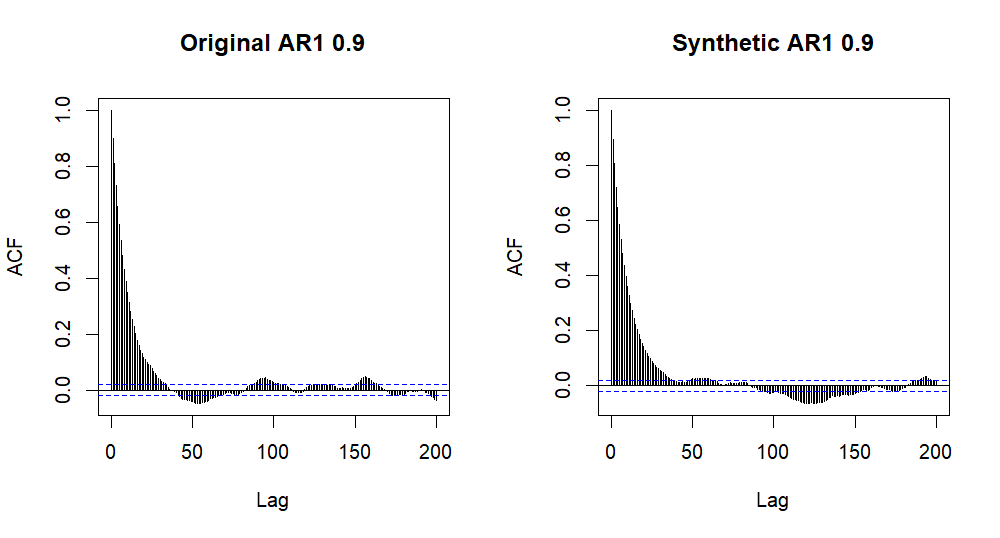}
        \caption{ACF Plots}
        \label{fig:acf_plot_ar19}
    \end{subfigure}
    \hfill
    \begin{subfigure}{0.45\textwidth}
        \centering
        \includegraphics[width=\linewidth]{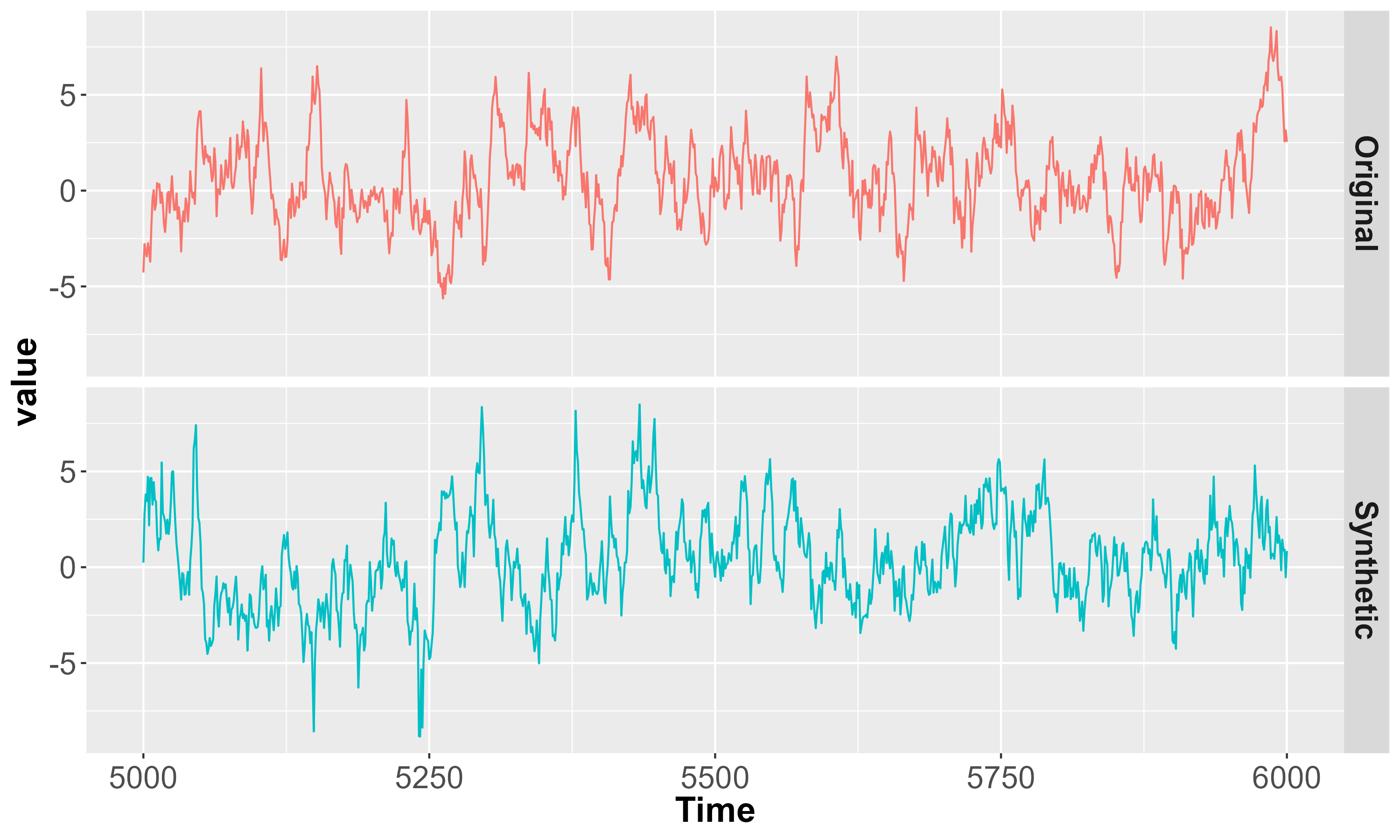}
        \caption{Time Series}
        \label{fig:ts_plot_ar19}
    \end{subfigure}
    \caption{Comparison between an original and synthetic \textbf{AR1 0.9} time series. (a) Autocorrelation function (ACF). (b) Time series (first 1000 observations).}
    \label{fig:acf_ts_plot_ar19}
\end{figure}

\begin{figure}[!h]
    \centering
    \begin{subfigure}{0.5\textwidth}
        \centering
        \includegraphics[width=\linewidth]{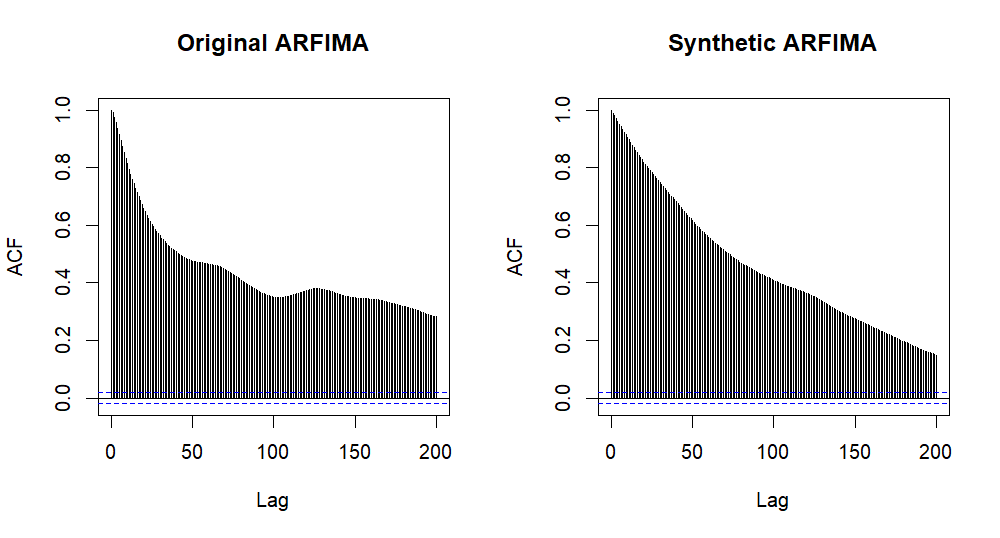}
        \caption{ACF Plots}
        \label{fig:acf_plot_arfima}
    \end{subfigure}
    \hfill
    \begin{subfigure}{0.45\textwidth}
        \centering
        \includegraphics[width=\linewidth]{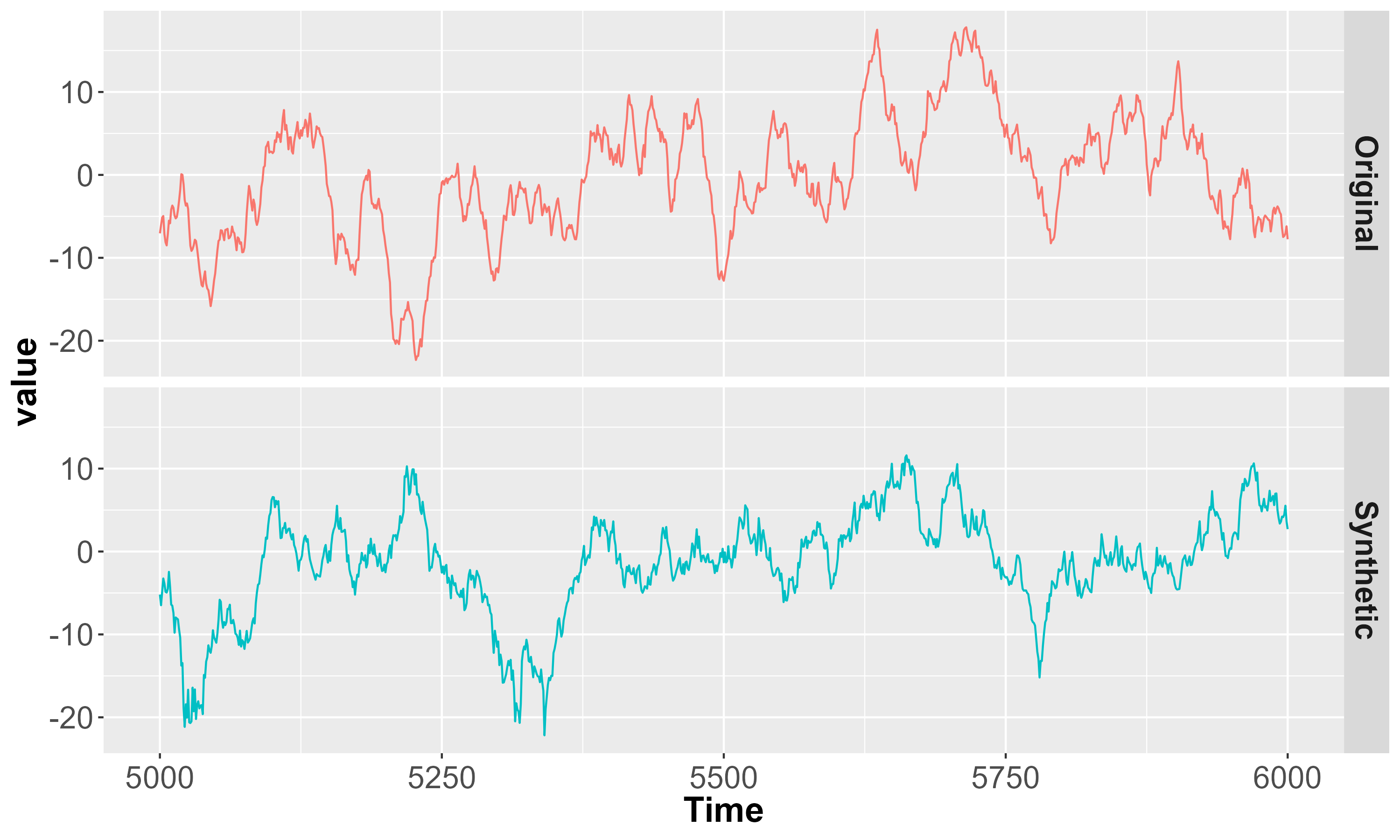}
        \caption{Time Series}
        \label{fig:ts_plot_arfima2}
    \end{subfigure}
    \caption{Comparison between an original and synthetic \textbf{ARFIMA} time series. (a) Autocorrelation function (ACF). (b) Time series (first 1000 observations).}
    \label{fig:acf_ts_plot_arfima}
\end{figure}

\begin{figure}[!h]
    \centering
    \begin{subfigure}{0.5\textwidth}
        \centering
        \includegraphics[width=\linewidth]{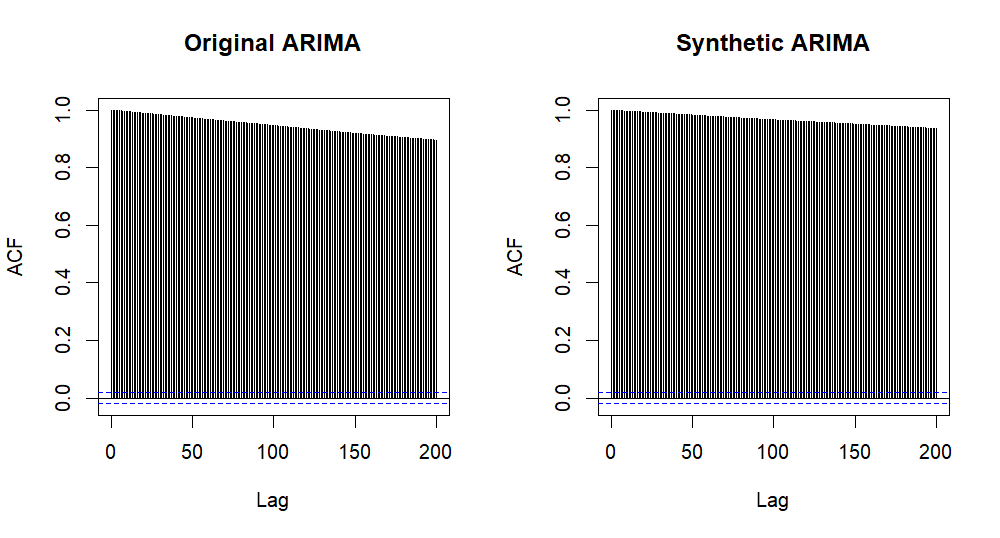}
        \caption{ACF Plots}
        \label{fig:acf_plot_arima}
    \end{subfigure}
    \hfill
    \begin{subfigure}{0.45\textwidth}
        \centering
        \includegraphics[width=\linewidth]{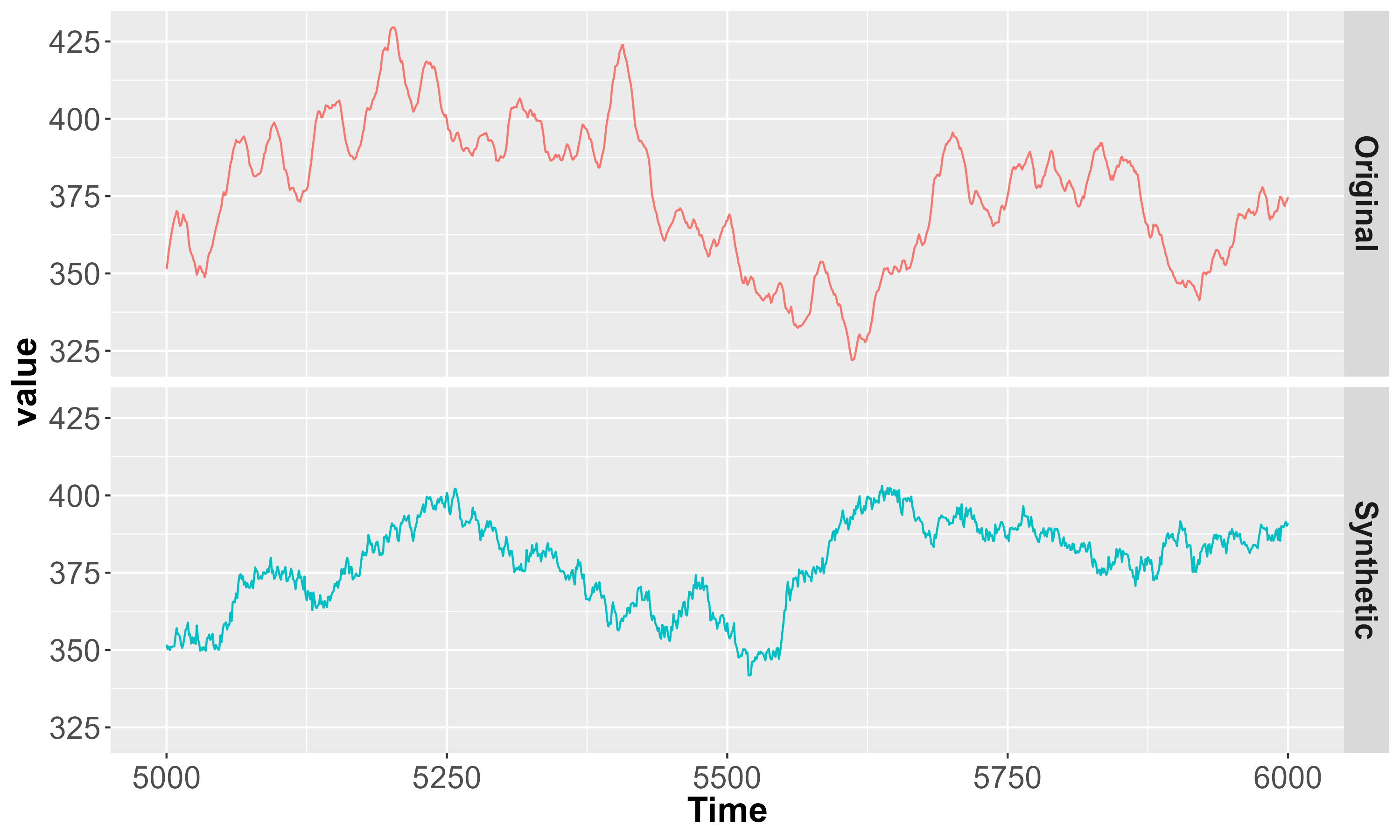}
        \caption{Time Series}
        \label{fig:ts_plot_arima2}
    \end{subfigure}
    \caption{Comparison between an original and synthetic \textbf{ARIMA} time series. (a) Autocorrelation function (ACF). (b) Time series (first 1000 observations).}
    \label{fig:acf_ts_plot_arima}
\end{figure}

\begin{figure}[h]
    \centering
    \begin{subfigure}{0.5\textwidth}
        \centering
        \includegraphics[width=\linewidth]{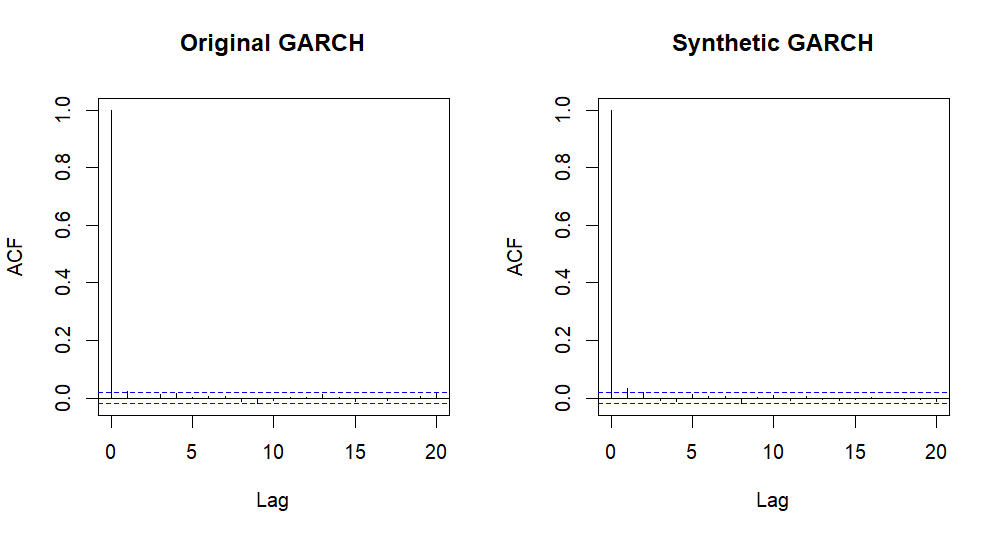}
        \caption{ACF Plots}
        \label{fig:acf_plot_garch}
    \end{subfigure}
    \hfill
    \begin{subfigure}{0.45\textwidth}
        \centering
        \includegraphics[width=\linewidth]{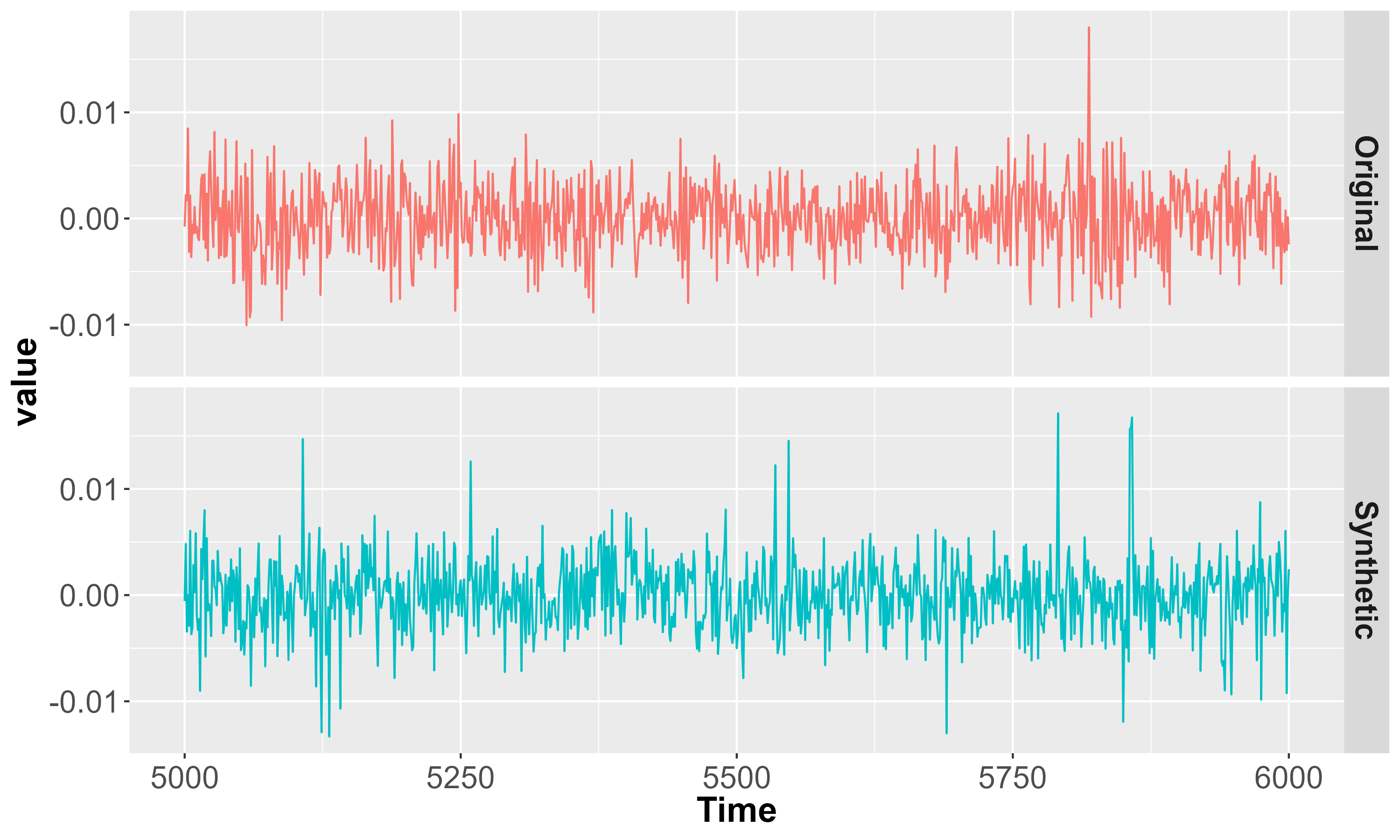}
        \caption{Time Series}
        \label{fig:ts_plot_garch}
    \end{subfigure}
    \caption{Comparison between an original and synthetic \textbf{GARCH} time series. (a) Autocorrelation function (ACF). (b) Time series (first 1000 observations).}
    \label{fig:acf_ts_plot_garch}
\end{figure}

\begin{figure}[h]
    \centering
    \begin{subfigure}{0.5\textwidth}
        \centering
        \includegraphics[width=\linewidth]{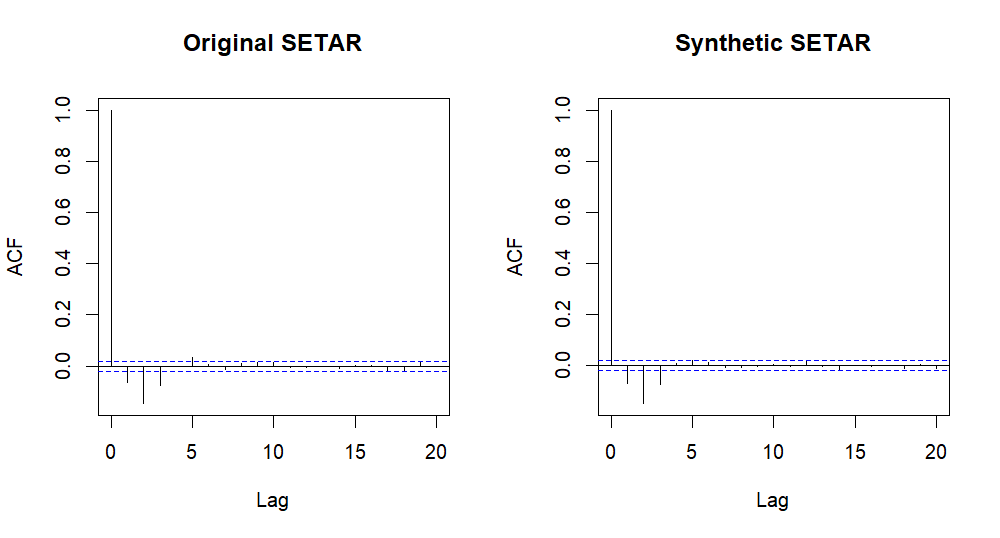}
        \caption{ACF Plots}
        \label{fig:acf_plot_setar}
    \end{subfigure}
    \hfill
    \begin{subfigure}{0.45\textwidth}
        \centering
        \includegraphics[width=\linewidth]{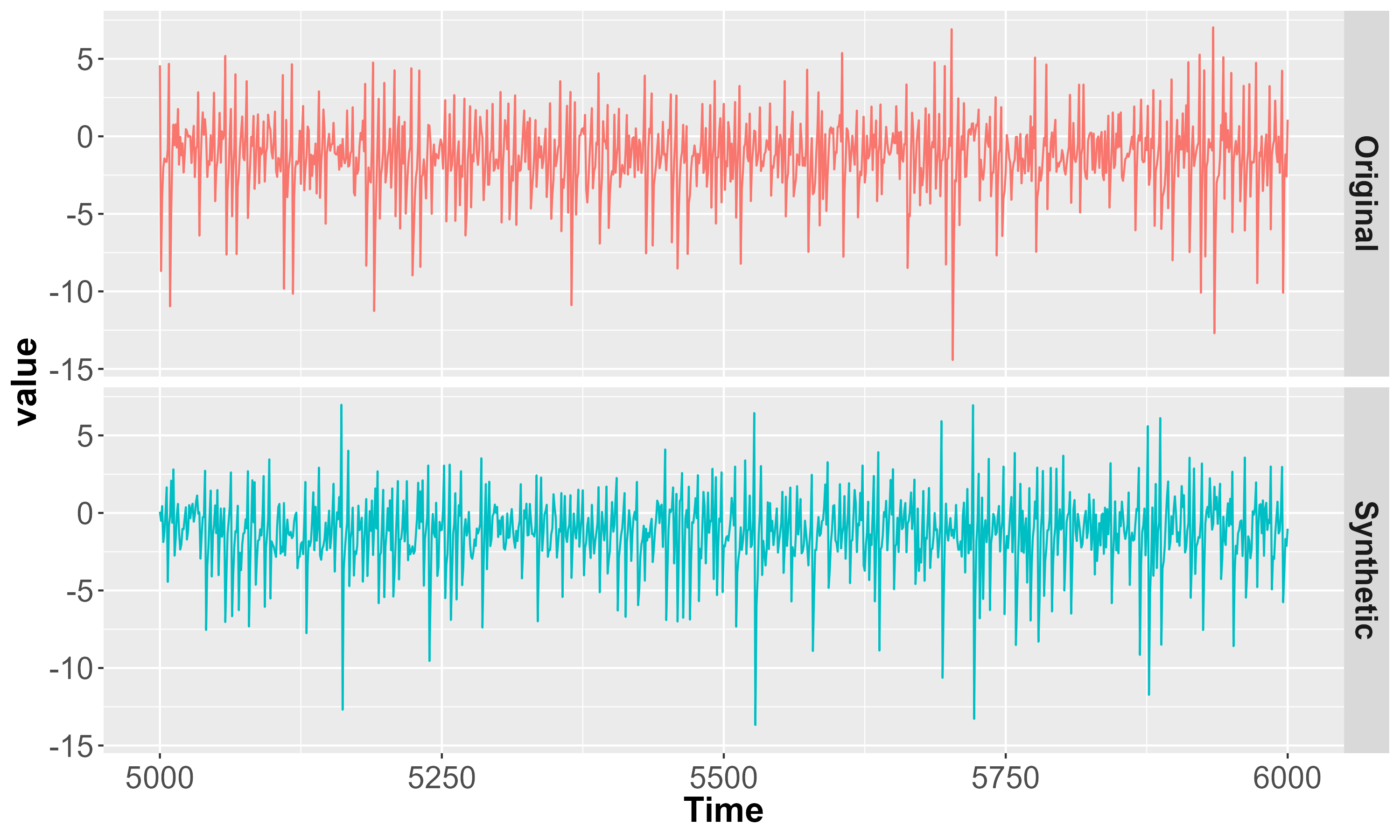}
        \caption{Time Series}
        \label{fig:ts_plot_setar}
    \end{subfigure}
    \caption{Comparison between an original and synthetic \textbf{SETAR} time series. (a) Autocorrelation function (ACF). (b) Time series (first 1000 observations).}
    \label{fig:acf_ts_plot_setar}
\end{figure}

\begin{figure}[h]
    \centering
    \begin{subfigure}{0.5\textwidth}
        \centering
        \includegraphics[width=\linewidth]{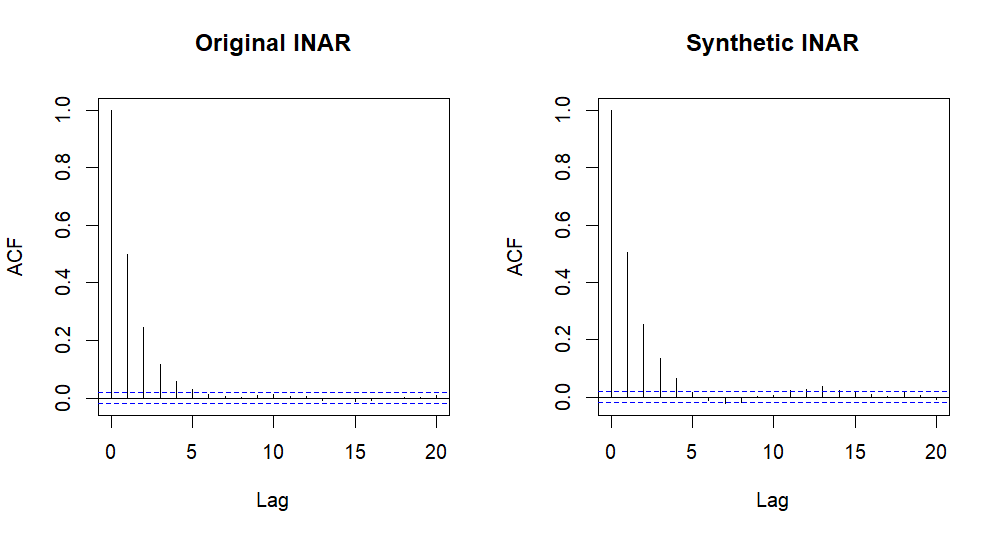}
        \caption{ACF Plots}
        \label{fig:acf_plot_inar}
    \end{subfigure}
    \hfill
    \begin{subfigure}{0.45\textwidth}
        \centering
        \includegraphics[width=\linewidth]{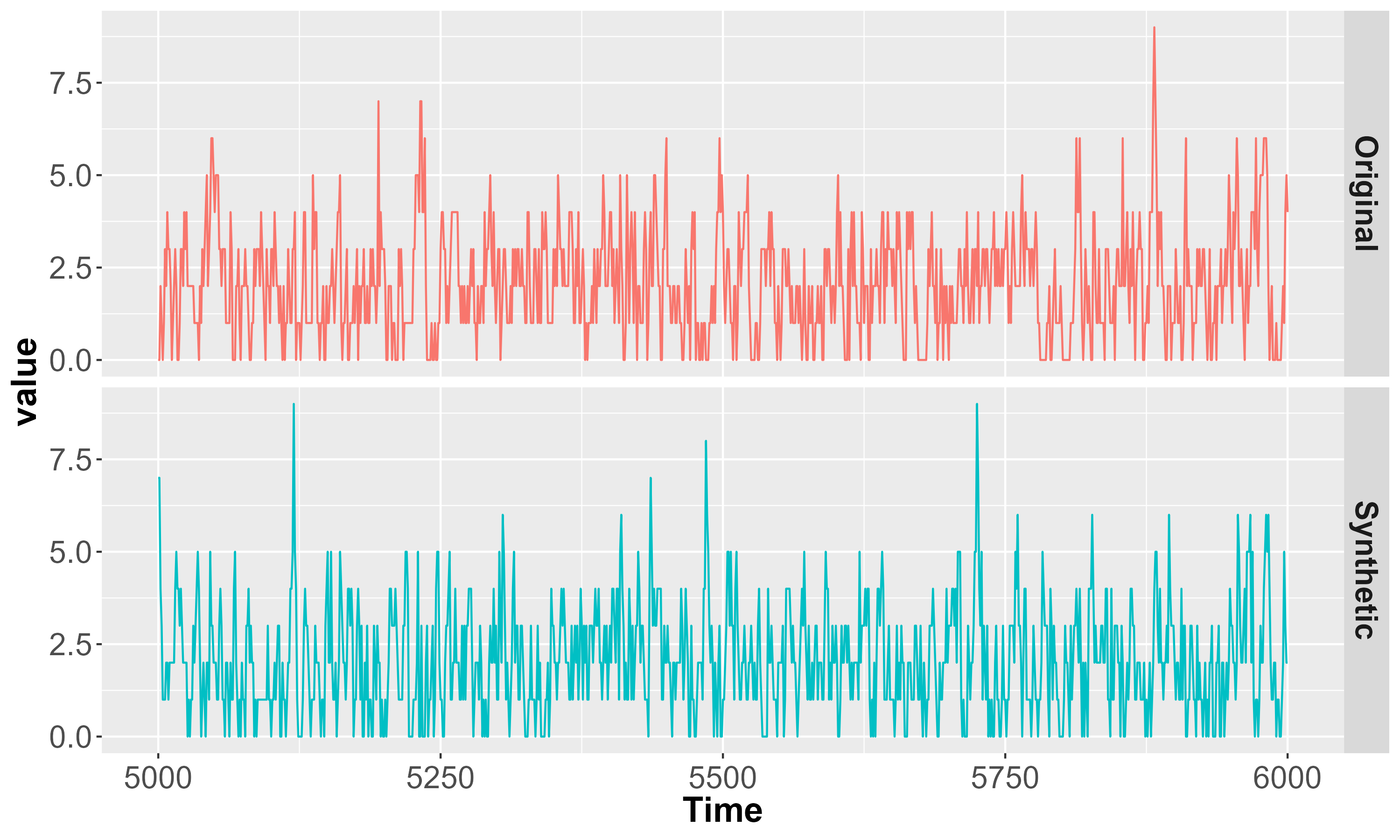}
        \caption{Time Series}
        \label{fig:ts_plot_inar}
    \end{subfigure}
    \caption{Comparison between an original and synthetic \textbf{INAR} time series. (a) Autocorrelation function (ACF). (b) Time series (first 1000 observations).}
    \label{fig:acf_ts_plot_inar}
\end{figure}

\section{Real Dataset: Comparing with Benchmark Methods}\label{app:t_sne_time_gan_vs_qg_real}

\begin{figure}[!h]
\centering
\captionsetup[subfigure]{justification=centering}

% First row
\hspace*{\fill}
\begin{subfigure}[t]{0.15\textwidth}
  \includegraphics[width=\linewidth]{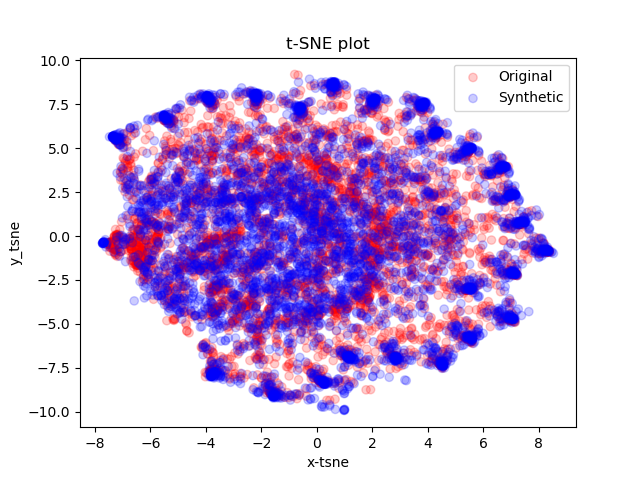}
\end{subfigure}
    \hfill
\hspace*{\fill}
\begin{subfigure}[t]{0.15\textwidth}
  \includegraphics[width=\linewidth]{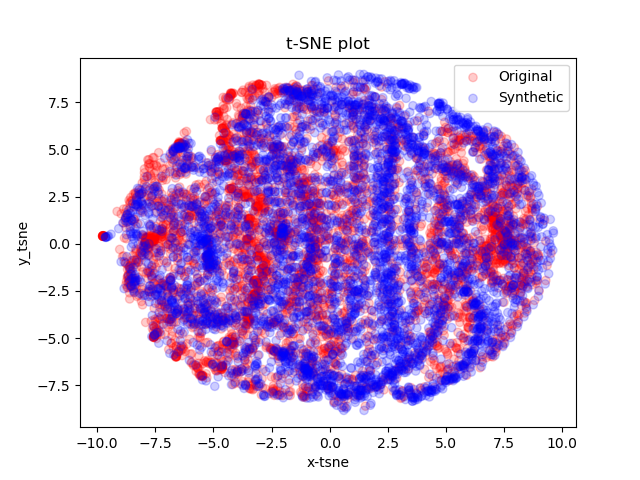}
\end{subfigure}
    \hfill
\begin{subfigure}[t]{0.15\textwidth}
  \includegraphics[width=\linewidth]{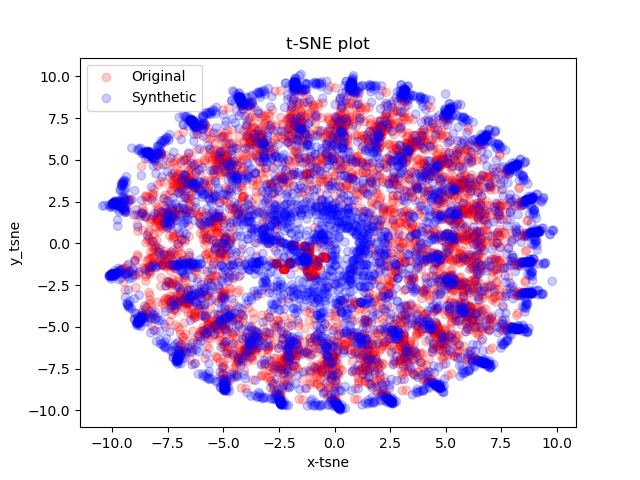}
\end{subfigure}
    \hfill
\begin{subfigure}[t]{0.15\textwidth}
  \includegraphics[width=\linewidth]{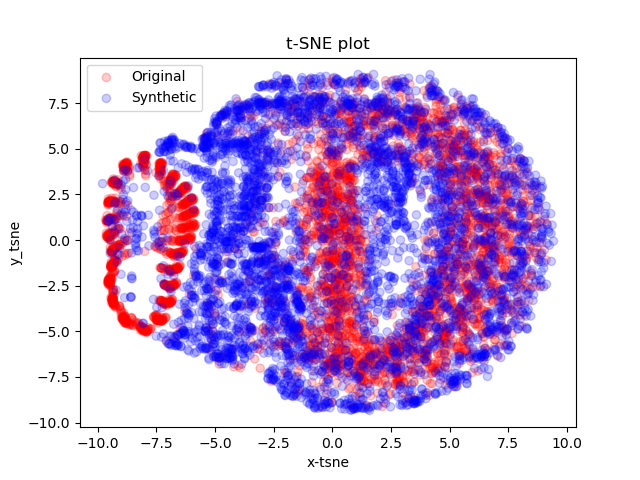}
\end{subfigure}
    \hfill
\begin{subfigure}[t]{0.15\textwidth}
  \includegraphics[width=\linewidth]{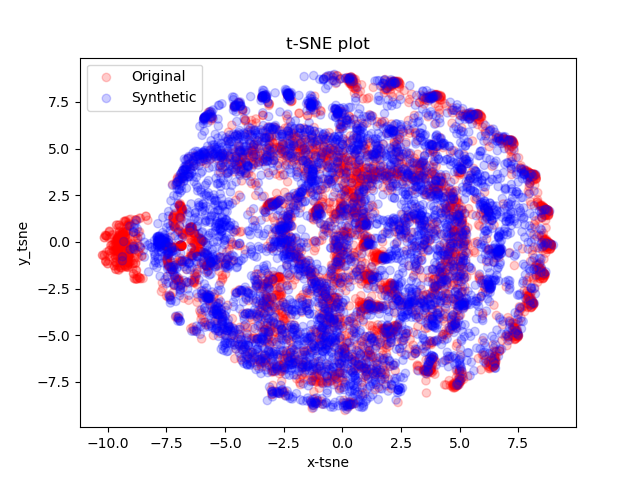}
\end{subfigure}
    \hfill
\begin{subfigure}[t]{0.15\textwidth}
  \includegraphics[width=\linewidth]{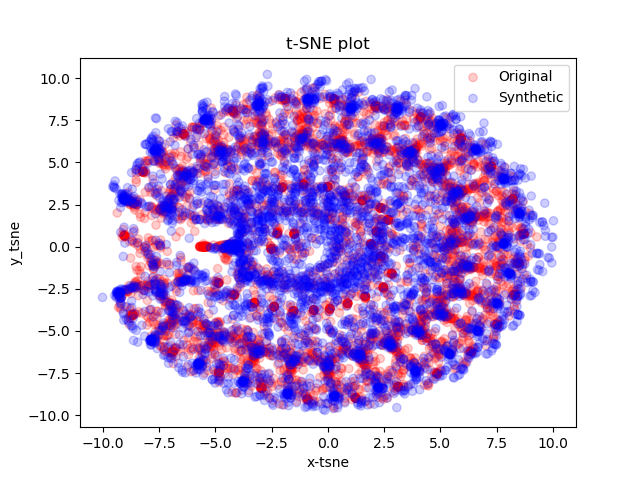}
\end{subfigure}
\hspace*{\fill}

% Second row
\hspace*{\fill}
\begin{subfigure}[t]{0.15\textwidth}
  \includegraphics[width=\linewidth]{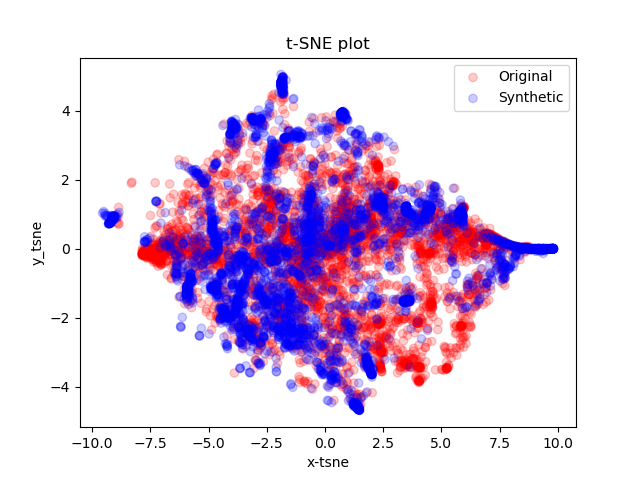}
\end{subfigure}
    \hfill
\begin{subfigure}[t]{0.15\textwidth}
  \includegraphics[width=\linewidth]{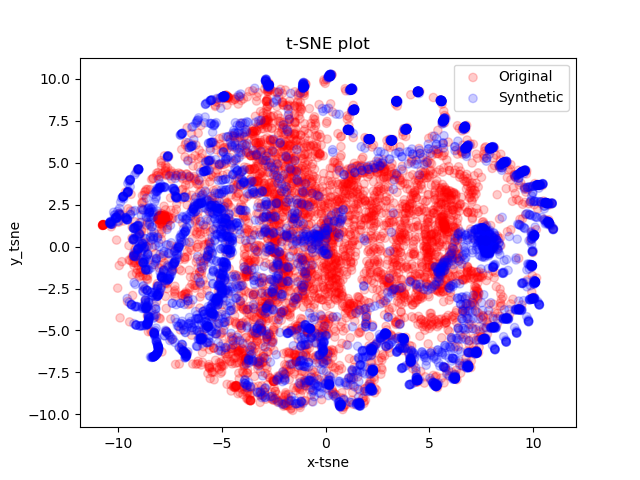}
\end{subfigure}
    \hfill
\begin{subfigure}[t]{0.15\textwidth}
  \includegraphics[width=\linewidth]{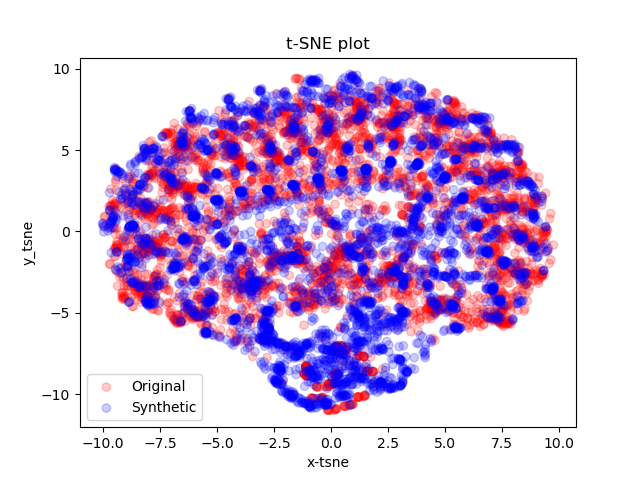}
\end{subfigure}
    \hfill
\begin{subfigure}[t]{0.15\textwidth}
  \includegraphics[width=\linewidth]{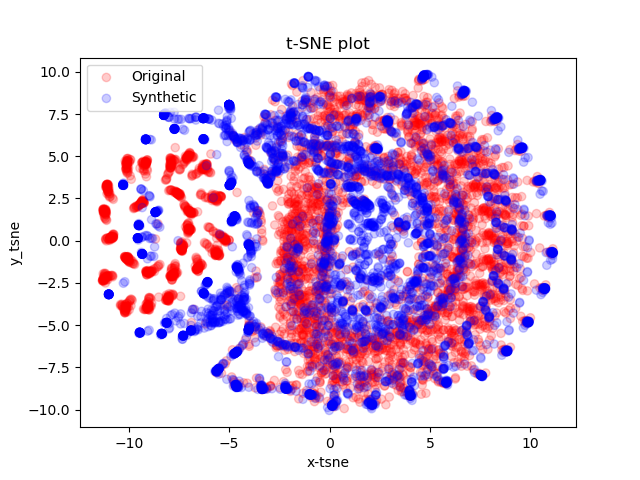}
\end{subfigure}
    \hfill
\begin{subfigure}[t]{0.15\textwidth}
  \includegraphics[width=\linewidth]{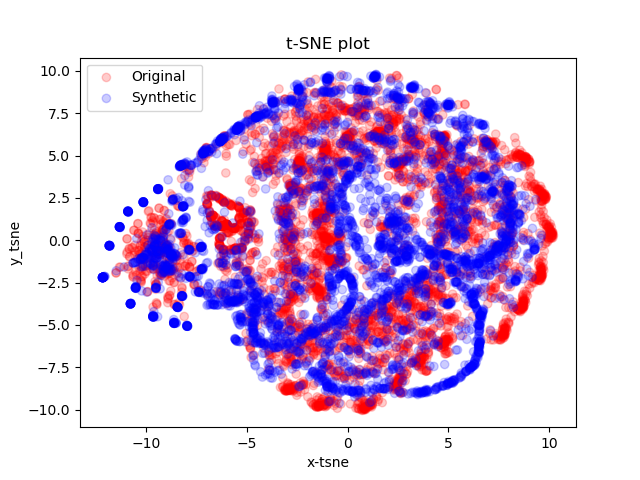}
\end{subfigure}
    \hfill
\begin{subfigure}[t]{0.15\textwidth}
  \includegraphics[width=\linewidth]{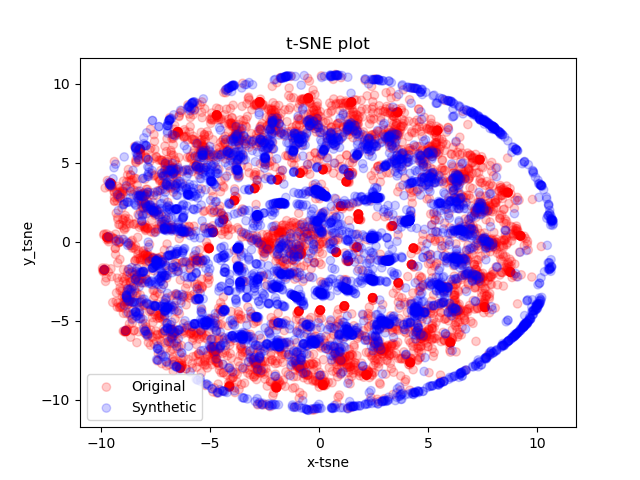}
\end{subfigure}
\hspace*{\fill}

\hspace*{\fill}
\caption[t‑SNE projections for synthetic data  (continued).]{t‑SNE projections for synthetic data generated by InvQG (top row)%, 
and TimeGAN (bottom row) for  households 1, 2, 3, 4, 9, and 10 \textit{(continued)}.}
\label{fig:t_sne_time_gan_vs_qg_real2}
\end{figure}

\begin{figure}[!h]
\centering
\captionsetup[subfigure]{justification=centering}

% First row
\hspace*{\fill}
\begin{subfigure}[t]{0.15\textwidth}
  \includegraphics[width=\linewidth]{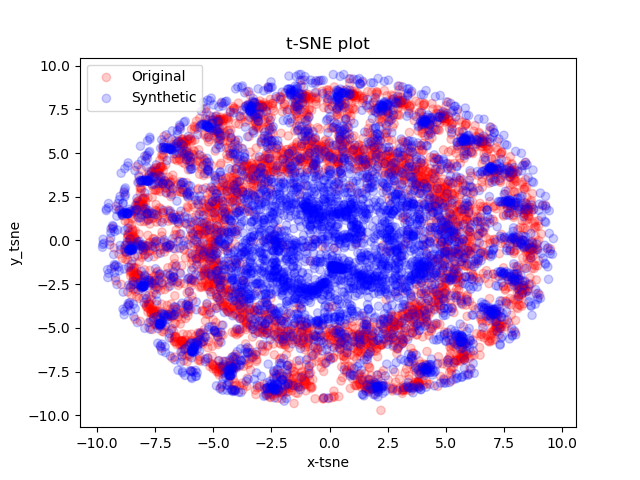}
\end{subfigure}
    \hfill
\hspace*{\fill}
\begin{subfigure}[t]{0.15\textwidth}
  \includegraphics[width=\linewidth]{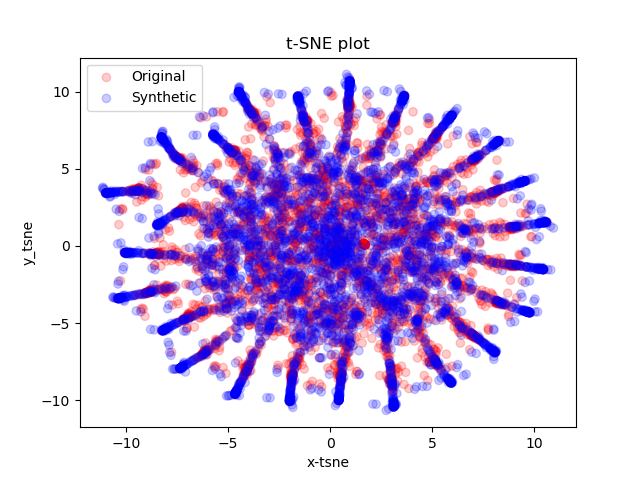}
\end{subfigure}
    \hfill
\begin{subfigure}[t]{0.15\textwidth}
  \includegraphics[width=\linewidth]{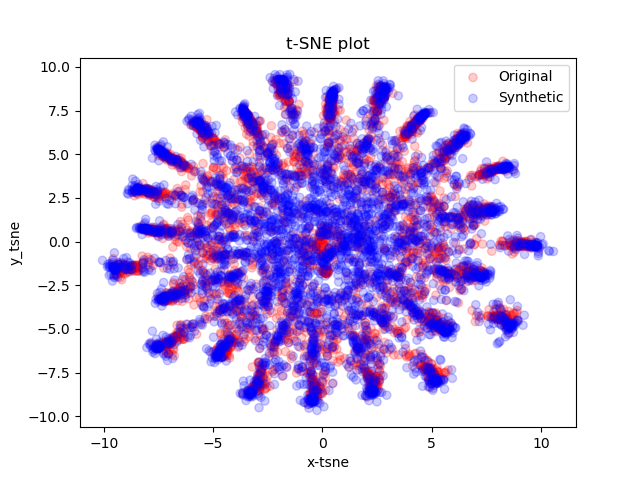}
\end{subfigure}
    \hfill
\begin{subfigure}[t]{0.15\textwidth}
  \includegraphics[width=\linewidth]{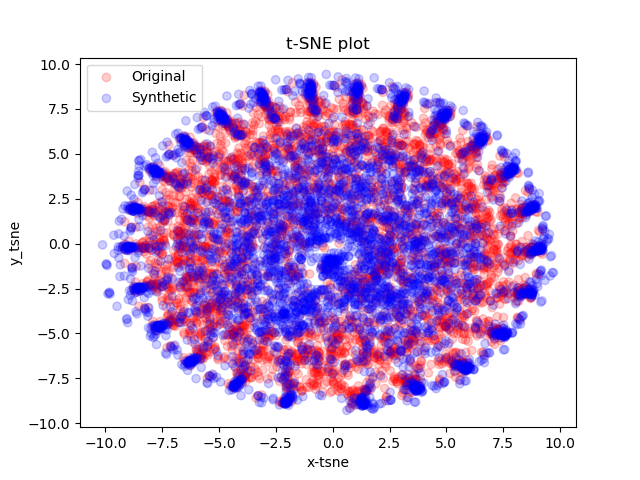}
\end{subfigure}
    \hfill
\begin{subfigure}[t]{0.15\textwidth}
  \includegraphics[width=\linewidth]{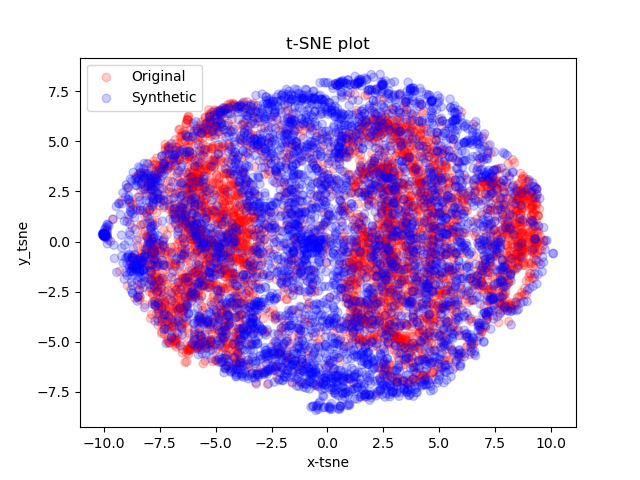}
\end{subfigure}
    \hfill
\begin{subfigure}[t]{0.15\textwidth}
  \includegraphics[width=\linewidth]{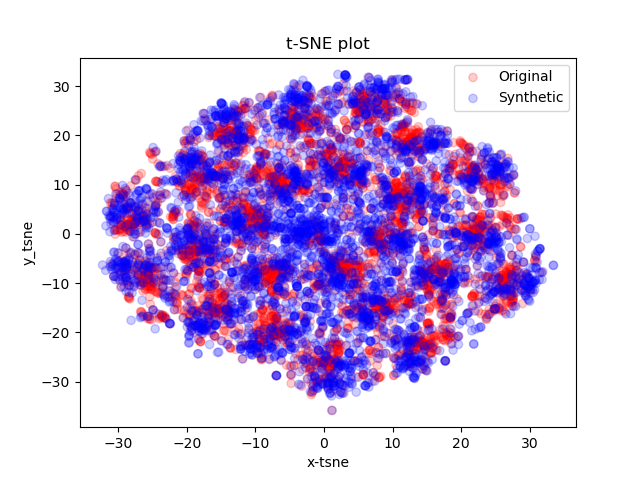}
\end{subfigure}
\hspace*{\fill}

% Second row
\hspace*{\fill}
\begin{subfigure}[t]{0.15\textwidth}
  \includegraphics[width=\linewidth]{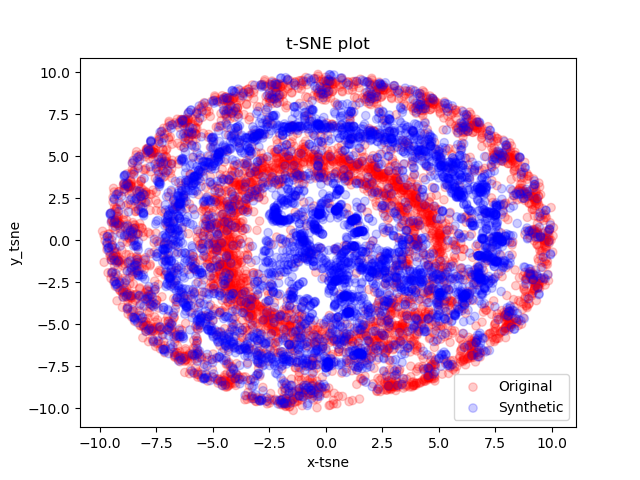}
\end{subfigure}
    \hfill
\begin{subfigure}[t]{0.15\textwidth}
  \includegraphics[width=\linewidth]{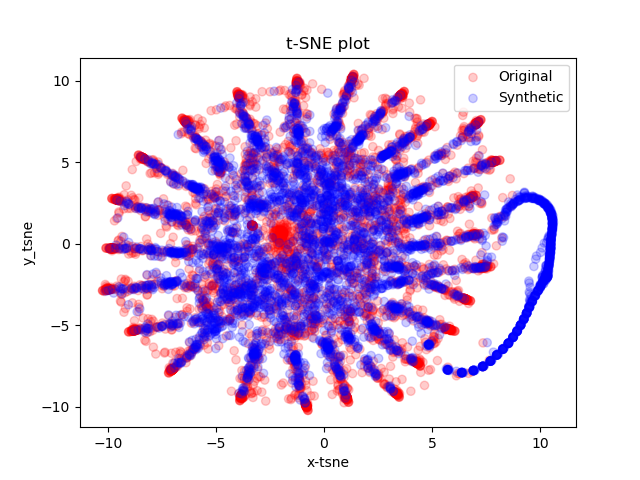}
\end{subfigure}
    \hfill
\begin{subfigure}[t]{0.15\textwidth}
  \includegraphics[width=\linewidth]{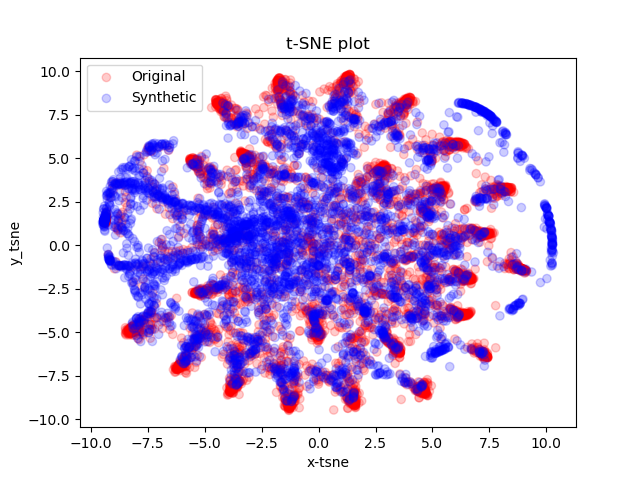}
\end{subfigure}
    \hfill
\begin{subfigure}[t]{0.15\textwidth}
  \includegraphics[width=\linewidth]{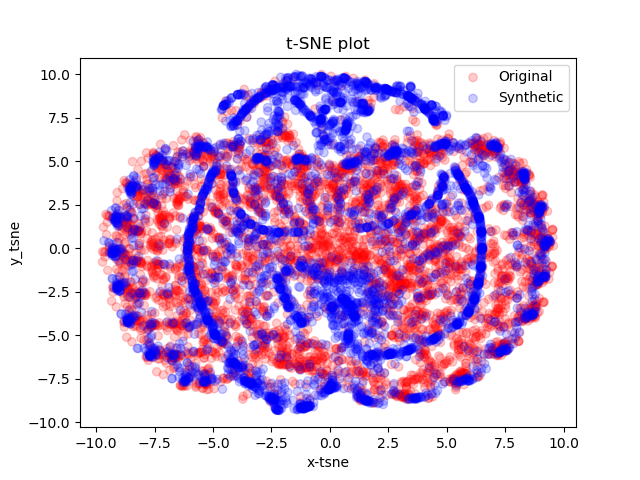}
\end{subfigure}
    \hfill
\begin{subfigure}[t]{0.15\textwidth}
  \includegraphics[width=\linewidth]{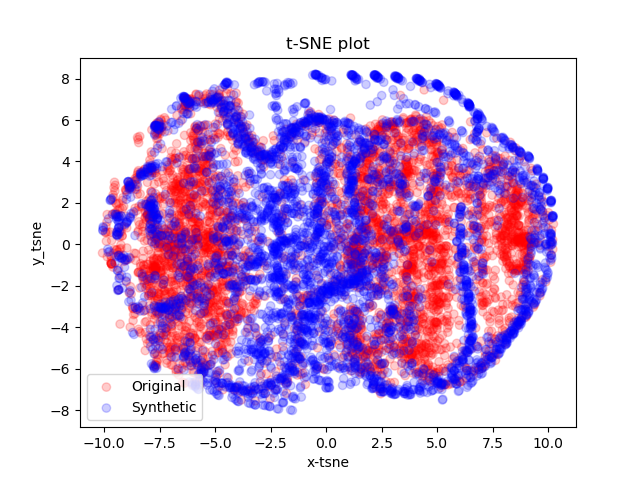}
\end{subfigure}
    \hfill
\begin{subfigure}[t]{0.15\textwidth}
  \includegraphics[width=\linewidth]{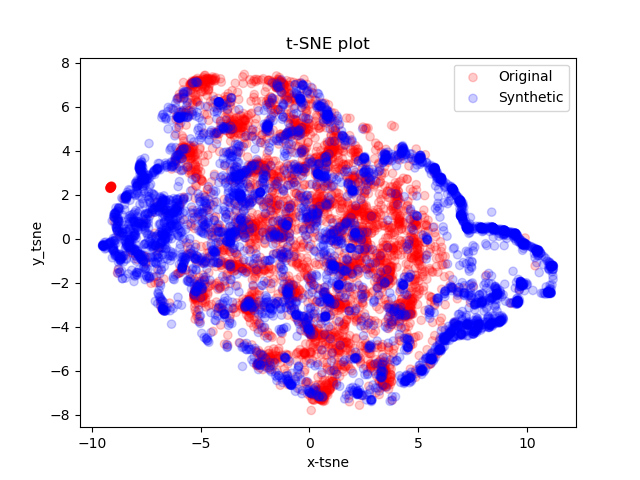}
\end{subfigure}
\hspace*{\fill}

\hspace*{\fill}
\caption[t‑SNE projections for synthetic data  (continued).]{t‑SNE projections for synthetic data generated by InvQG (top row)%, 
 and TimeGAN  (bottom row) for  households 11, 12, 13, 15, 16, and 18 \textit{(continued)}.}
\label{fig:t_sne_time_gan_vs_qg_real3}
\end{figure}

\begin{figure}[!h]
\centering
\captionsetup[subfigure]{justification=centering}

% First row
\hspace*{\fill}
\begin{subfigure}[t]{0.15\textwidth}
  \includegraphics[width=\linewidth]{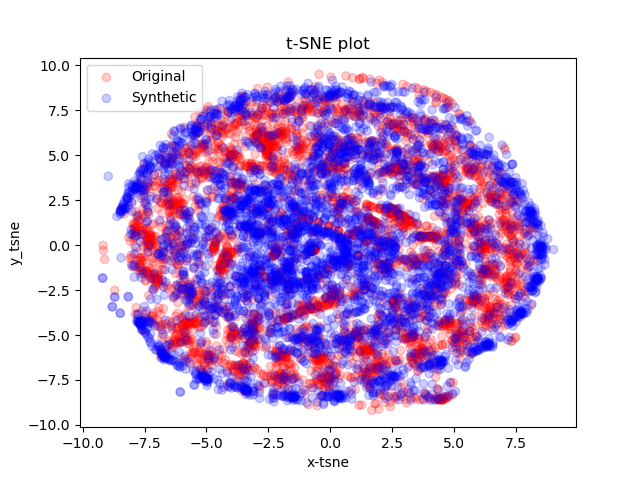}
\end{subfigure}
    \hfill
\begin{subfigure}[t]{0.15\textwidth}
  \includegraphics[width=\linewidth]{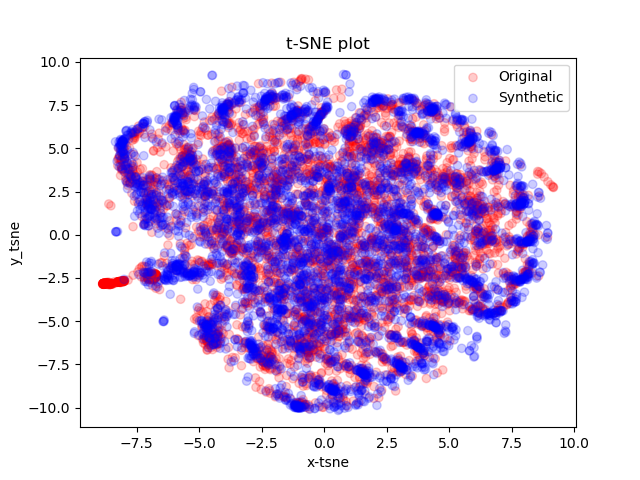}
\end{subfigure}
    \hfill
\begin{subfigure}[t]{0.15\textwidth}
  \includegraphics[width=\linewidth]{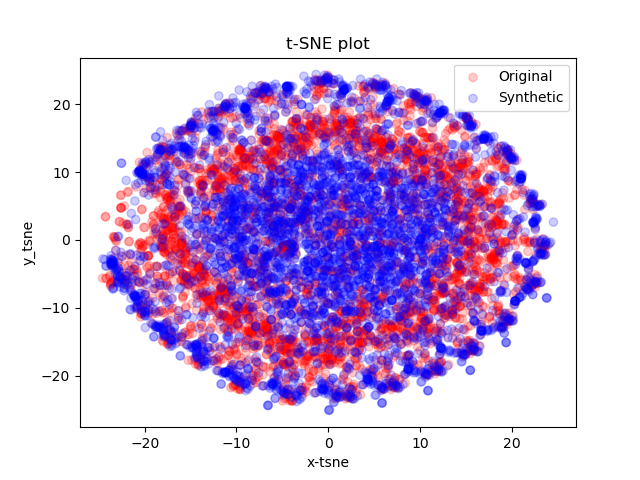}
\end{subfigure}
    \hfill
\begin{subfigure}[t]{0.15\textwidth}
  \includegraphics[width=\linewidth]{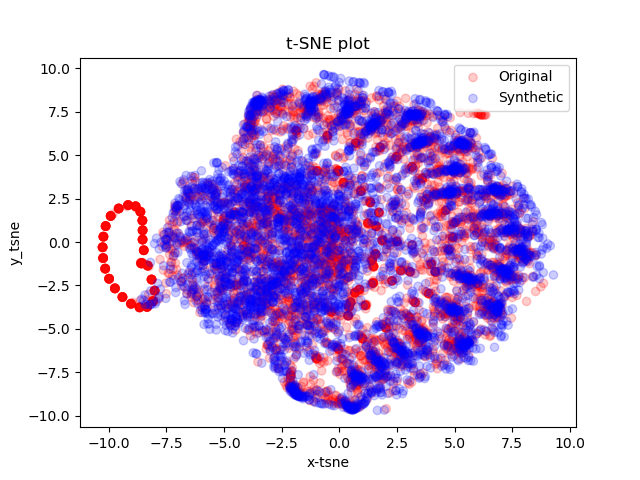}
\end{subfigure}
\hspace*{\fill}

% Second row
\hspace*{\fill}
\begin{subfigure}[t]{0.15\textwidth}
  \includegraphics[width=\linewidth]{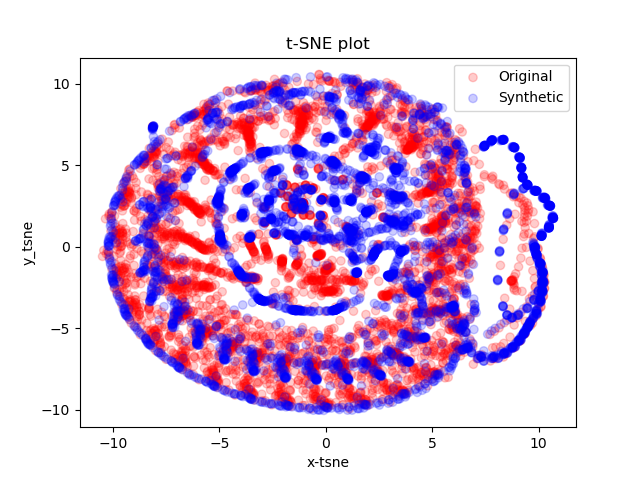}
\end{subfigure}
    \hfill
\begin{subfigure}[t]{0.15\textwidth}
  \includegraphics[width=\linewidth]{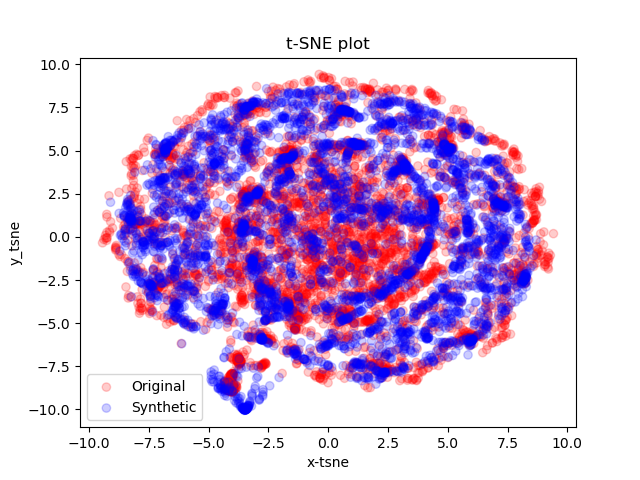}
\end{subfigure}
    \hfill
\begin{subfigure}[t]{0.15\textwidth}
  \includegraphics[width=\linewidth]{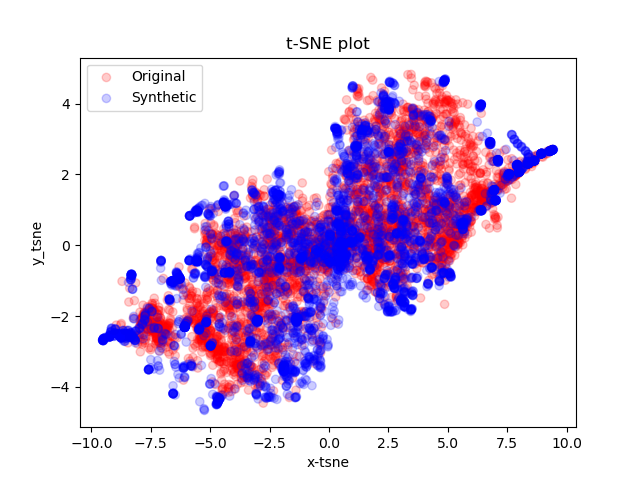}
\end{subfigure}
    \hfill
\begin{subfigure}[t]{0.15\textwidth}
  \includegraphics[width=\linewidth]{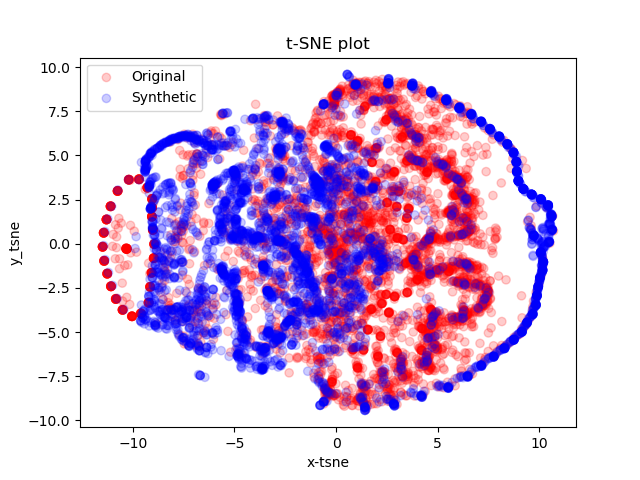}
\end{subfigure}
\hspace*{\fill}

\hspace*{\fill}
\caption[t‑SNE projections for synthetic data  (continued).]{t‑SNE projections for synthetic data generated by InvQG (top row)%, 
 and 
 TimeGAN  (bottom row) for  households 19, 20, 21,and 22 \textit{(continued)}.}
\label{fig:t_sne_time_gan_vs_qg_real4}
\end{figure}

\end{document}